\documentclass{article}

\PassOptionsToPackage{numbers}{natbib}
\usepackage[main, final]{neurips_2026}

\usepackage[utf8]{inputenc}
\usepackage[T1]{fontenc}
\usepackage{hyperref}
\usepackage{url}
\usepackage{booktabs}
\usepackage{amsfonts}
\usepackage{nicefrac}
\usepackage{microtype}
\usepackage{xcolor}

\usepackage{multirow}
\usepackage{caption}
\usepackage{subcaption}
\usepackage{colortbl}
\usepackage{array}
\usepackage{amssymb}
\usepackage{pifont}
\newcommand{\tablestyle}[2]{\setlength{\tabcolsep}{#1}\renewcommand{\arraystretch}{#2}\centering\footnotesize}

\usepackage{makecell}
\usepackage[most]{tcolorbox}
\usepackage{amsmath}
\usepackage{mathtools}
\usepackage{amsthm}
\usepackage{enumitem}
\usepackage{rotating}
\usepackage{graphicx}
\graphicspath{{figures/}}

\theoremstyle{plain}

\theoremstyle{definition}

\theoremstyle{remark}

\newtcolorbox{protocolbox}[2][]{
    enhanced,
    breakable,
    colback=gray!5!white,
    colframe=gray!80!black,
    title=#2,
    fonttitle=\bfseries,
    attach boxed title to top left={yshift=-2mm, xshift=5mm},
    boxed title style={colback=gray!80!black, sharp corners=south},
    arc=3mm,
    boxrule=0.8pt,
    top=4mm,
    #1
}

\title{ARK: A Dual-Axis Multimodal Retrieval Benchmark along Reasoning and Knowledge }

\author{%
  Yijie Lin\textsuperscript{1,4}\thanks{Equal contribution}\quad
  Guofeng Ding\textsuperscript{2}\footnotemark[1]\quad
  Haochen Zhou\textsuperscript{2}\footnotemark[1]\quad
  Haobin Li\textsuperscript{2}\quad
  Mouxing Yang\textsuperscript{2}\quad
  Xi Peng\textsuperscript{1,3}\thanks{Corresponding author}\\[3pt]
  \textsuperscript{1}School of Artificial Intelligence, Sichuan University, Chengdu, China\\
  \textsuperscript{2}College of Computer Science, Sichuan University, Chengdu, China\\
  \textsuperscript{3}National Key Laboratory of Fundamental Algorithms and Models
  for \\ Engineering Simulation, Sichuan University, Chengdu, China\\
  \textsuperscript{4}The Hong Kong University of Science and Technology, Hong Kong SAR, China\\[3pt]
  {\footnotesize\texttt{\{linyijie.gm,guofengding.gm,haochenzhou.xl,haobinli.gm,yangmouxing,pengx.gm\}@gmail.com}}\\[3pt]
  \url{https://huggingface.co/datasets/XLearning-SCU/ARK-Bench}
}

\begin{document}

\maketitle

\begin{abstract}
Existing multimodal retrieval benchmarks largely emphasize semantic matching on daily-life images and offer limited diagnostics of professional knowledge and complex reasoning. To address this gap, we introduce ARK, a benchmark designed to analyze multimodal retrieval from two complementary perspectives: (i) knowledge domains (five domains with 17 subtypes), which characterize the content and expertise retrieval relies on, and (ii) reasoning skills (six categories), which characterize the type of inference over multimodal evidence required to identify the correct candidate. Specifically, ARK evaluates retrieval with both unimodal and multimodal queries and candidates, covering 16 heterogeneous visual data types. To avoid shortcut matching during evaluation, most queries are paired with targeted hard negatives that require multi-step reasoning. We evaluate 25 representative text-based and multimodal retrievers and observe a pronounced gap between knowledge- and reasoning-intensive retrieval, with fine-grained visual and spatial reasoning as persistent bottlenecks. We further show that enhancements such as re-ranking, rewriting, and agentic retrieval yield consistent gains, but substantial headroom remains.
\end{abstract}

\section{Introduction}

\begin{figure}[t]
    \centering
    \includegraphics[width=0.95\linewidth]{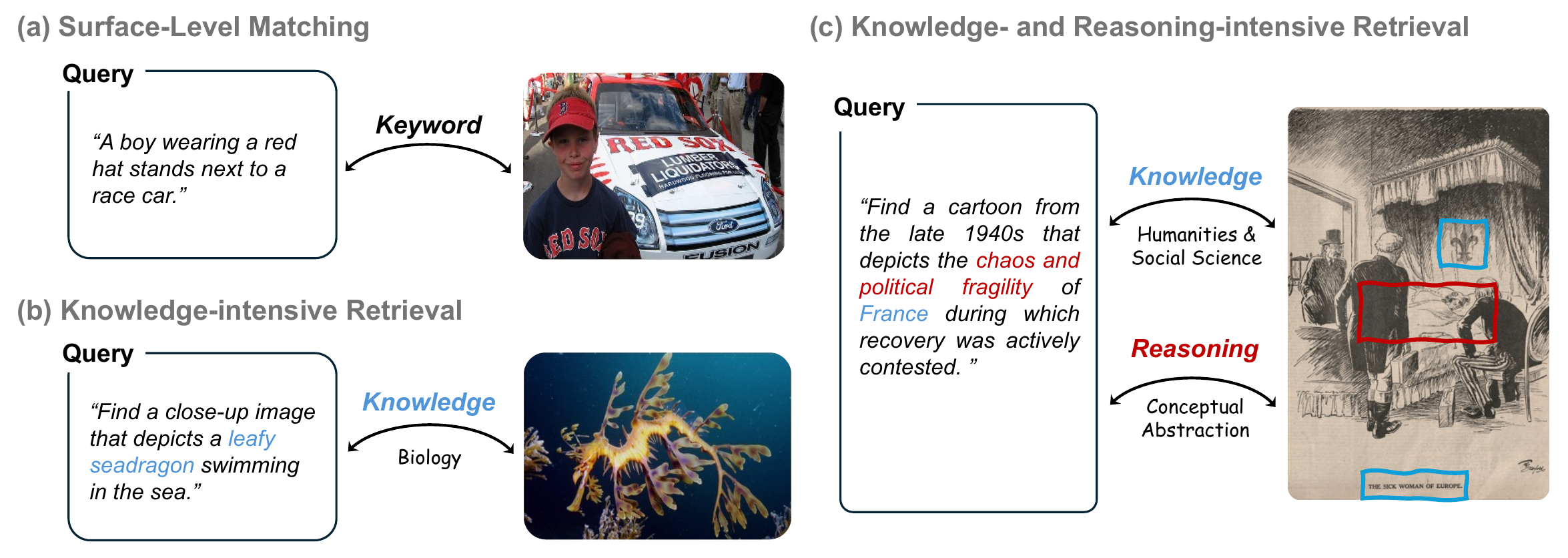}
  \caption{
  \textbf{Evolution of multimodal retrieval.}
(a) Traditional benchmarks emphasize object-level or surface-level semantic similarity matching on daily imagery.
(b) Recent benchmarks shift toward \emph{knowledge-intensive} retrieval, where relevance hinges on domain knowledge beyond surface matching.
(c) ARK explicitly separates \emph{knowledge} and \emph{reasoning} as two evaluation axes, enabling diagnosis of retrieval that requires both domain knowledge and structured reasoning over multimodal evidence.
    }
    \label{fig:fig1}
\end{figure}

Multimodal retrieval is a fundamental capability for modern AI systems, supporting applications such as recommender systems~\cite{li2025towards,zhou2026wemm}, retrieval-augmented generation (RAG)~\cite{yasunaga2023retrieval,hu2023reveal,xu2026multi}, and Deep Research~\cite{openai_deep_research_card_2025}. To assess progress, the community has developed a variety of benchmarks. In most settings, relevance between a query and candidate documents can be captured by relatively simple matching patterns, \textit{e.g.}, shared object categories or surface-level semantic similarity~\cite{COCO,flickr} as shown in Fig.~\ref{fig:fig1}(a).

However, in real-world scenarios, relevance is often established through nuanced and complex connections that go far beyond direct semantic overlap, requiring substantial domain knowledge and multi-step reasoning to retrieve the correct images or documents.
As illustrated in Fig.~\ref{fig:fig1}(c), a historian may search for comics that implicitly encode a specific historical event, where relevance hinges on interpreting visual evidence (\textit{e.g.}, characters and symbols) together with query-specific context (\textit{e.g.}, temporal cues).
In such cases, neither text-only nor vision-only cues suffice, and accurate retrieval depends on combining domain knowledge with reasoning over multimodal evidence.

To systematically evaluate such capabilities, we introduce a dual-Axis multimodal retrieval benchmark for Reasoning and Knowledge (ARK).
Specifically, ARK is designed to assess retrieval systems along two orthogonal and complementary dimensions: i) knowledge-intensive scenarios, where correct retrieval mainly depends on recognizing and recalling domain concepts, and ii) reasoning-intensive scenarios, where retrieval requires multi-step inference, relational understanding, or abstraction.
Compared with existing multimodal retrieval benchmarks, ARK provides several distinctive contributions:
\begin{itemize}
  [leftmargin=12pt,
  topsep=-1pt,itemsep=0pt]
  \item \textbf{The first multimodal retrieval benchmark that explicitly separates knowledge and reasoning as two evaluation axes}.
  Existing benchmarks~\cite{mrmr,mr2} conflate these two factors, obscuring whether failures stem from missing domain knowledge or insufficient reasoning.
  As illustrated in Fig.~\ref{fig:fig1}(b), retrieving images of ``leafy seadragon'' is largely knowledge-intensive but reasoning-light. Once the visual concept is known, the task can be solved at a glance, which may overestimate models that primarily rely on memorizing visual concepts.
  In contrast, ARK annotates each instance along both axes, enabling systematic analysis across knowledge domains and reasoning skills.

\item \textbf{Broad coverage of knowledge domains and data types}.
Most existing datasets focus on daily-life scenarios with natural images~\cite{VN,EDIS,fashioniq,CIRR}, providing limited evaluation of professional or technical knowledge.
Along the knowledge axis, ARK spans five domains: \textit{visual cognition, natural science, formal science, humanities and social science, and engineering and technology} (Fig.~\ref{fig:fig2}(a)), and includes 16 heterogeneous visual types, such as \textit{tables, line charts, chemical structures, artworks, comics, and cognitive maps} (Fig.~\ref{fig:fig2}(b)).
This diversity enables evaluation of retrieval systems that require specialized knowledge beyond commonsense visual understanding.

\item \textbf{Systematic evaluation of diverse reasoning skills}.
ARK evaluates six reasoning categories: \textit{knowledge reasoning, spatial reasoning, logical reasoning, symbolic reasoning, fine-grained visual reasoning, and conceptual abstraction} (Fig.~\ref{fig:fig2}(d)).
These categories cover a broader and more fine-grained spectrum of reasoning phenomena than prior benchmarks~\cite{mr2}.
To discourage shortcut matching, we carefully design queries and targeted hard negatives so that correct retrieval relies on genuine reasoning rather than superficial semantic cues.

\end{itemize}

Based on ARK, we evaluate 25 representative retriever configurations, including embedding-based retrievers, re-rankers, and agentic retrievers, spanning diverse model sizes and architectures, and obtain three key findings:
(i) While current systems handle many knowledge-intensive queries reasonably well, they fall sharply behind on reasoning-intensive tasks. Performance also varies substantially across disciplines, revealing limited generalization and pronounced domain dependence. Together, these results indicate that closing the gap in multimodal retrieval requires not only broader knowledge coverage but also stronger and more transferable reasoning capabilities.
(ii) Vision-centric reasoning is a key bottleneck on ARK.
Breaking down performance by reasoning skill, we observe a consistent trend across models: performance is relatively stronger on knowledge reasoning and conceptual abstraction, but drops markedly on vision-centric skills such as fine-grained visual reasoning and spatial reasoning.
These results suggest that current retrievers struggle to localize subtle evidence in high-resolution images and to reason over geometric or topological relations. In short, visual grounding and spatial inference remain key obstacles for reasoning-intensive multimodal retrieval.
(iii) While larger models generally achieve stronger retrieval performance, scaling alone does not eliminate the core reasoning bottlenecks on ARK.
In contrast, explicitly injecting reasoning at inference time through lightweight interventions such as query rewriting and re-ranking yields consistent gains, highlighting a practical path to improve reasoning-intensive multimodal retrieval.

We hope that ARK and our empirical findings will help illuminate current limitations and guide the development of more robust and reasoning-aware multimodal retrieval systems.

\section{Related Work}

\subsection{Benchmarks for Multimodal Retrieval}
Early benchmarks mainly focus on text--image~\cite{COCO,flickr} and text--video~\cite{msrvtt,msvd,lsmdc} matching based on semantic relevance, and have been further extended to composed image retrieval~\cite{zhang2024magiclens} and interactive retrieval~\cite{das2017visual,lullava}.
However, most existing benchmarks are dominated by everyday natural images and videos, offering limited coverage of professional knowledge and placing little emphasis on reasoning.

Very recently, MRMR~\cite{mrmr} and MR$^2$-Bench~\cite{mr2} introduce more challenging retrieval settings with reasoning requirements.
However, many of their queries are adapted from knowledge-heavy VQA sources, which tend to emphasize domain knowledge (\textit{e.g.}, nature, cooking, medicine) rather than providing systematic coverage and diagnostics of reasoning skills.
Moreover, the reasoning they probe is typically limited to relatively simple visual relations.
In contrast, ARK targets a broader spectrum of reasoning demands, including richer spatial reasoning (\textit{e.g.}, 3D projection/folding/rotation and cognitive-map understanding), fine-grained visual reasoning, and higher-level logical, symbolic, and abstract inference.
Overall, ARK complements these efforts by explicitly separating knowledge and reasoning as two evaluation axes and enabling finer-grained evaluation across diverse domains and reasoning types.
A detailed comparison with existing benchmarks is provided in Appendix~\ref{appendix:comparison_existing_benchmarks}.

\subsection{Multimodal Retrievers}
Progress in multimodal retrieval~\cite{zhou2024vista} has closely tracked advances in model architectures and pretraining paradigms. Early retrievers typically adopt dual-stream designs that encode images and texts separately and align them in a shared embedding space for efficient similarity search~\cite{clip,EVA-CLIP-18B,siglip,lin2024multi,li2025enhancing}.
With the rise of multimodal foundation models, an increasing number of methods move toward re-ranking architectures that model cross-modal interactions over concatenated inputs~\cite{albef,blip2,huang2024noise}. Recently, retrievers built on large language models (LLMs) have demonstrated strong retrieval capability~\cite{lamra,uniir,li2026qwen3,huang2026llm2clip}.

However, most of these approaches still frame retrieval primarily as a semantic similarity matching problem and therefore do not explicitly elicit multi-step reasoning during search.
Inspired by recent progress on reasoning-augmented text retrievers~\cite{diver,reasonir}, emerging work has started to inject reasoning signals into multimodal retrieval, \textit{e.g.}, via reasoning-enhanced fine-tuning~\cite{xu2025mm} or by generating richer textual contexts~\cite{cui2025think} to improve embeddings.
In this work, we systematically evaluate representative multimodal retrievers on ARK and analyze how model architecture, scale, and reasoning-oriented training affect both knowledge-intensive and reasoning-intensive retrieval performance.

\section{ARK Benchmark}

We introduce ARK, a knowledge- and reasoning-intensive benchmark for multimodal retrieval.
In this section, we first present the task formulation in Section~\ref{sec:task_formulation}, then describe the dual-axis taxonomy of knowledge and reasoning that structures ARK in Section~\ref{sec:knowledge_reasoning_axes}, and finally detail the data curation pipeline in Section~\ref{sec:data_curation_process}.

\subsection{Task Formulation}
\label{sec:task_formulation}

ARK evaluates retrieval with both unimodal and multimodal queries and candidates.
Given a query $q$ and the gallery $\mathcal{D} = \{d_i\}_{i=1}^{N}$, the goal is to rank all candidates such that the ground-truth relevant item $d^{+}$ is ranked as high as possible.

\begin{table*}[t]
\centering
\caption{Data statistics of queries and gallery from the knowledge perspective in ARK benchmark.}
\label{tab:ark_knowledge_stats}

\begin{minipage}[t]{0.49\textwidth}
\centering
\vspace{0pt}
{
\tablestyle{1.5pt}{1}{
\begin{tabular}{l l r r}
\toprule
\textbf{Knowledge Domain} & \textbf{Subtype} & \textbf{Query} & \textbf{Gallery} \\
\midrule

\multirow{3}{*}{\parbox{2.6cm}{\centering\textsc{Natural\\Science}}}
 & Biology        & 83  & 1834 \\
 & Physics        & 80  & 712  \\
 & Chemistry      & 105 & 1127 \\
\midrule
\addlinespace[5.0625pt]

\multirow{2}{*}{\parbox{2.6cm}{\centering\textsc{Formal\\Science}}}
 & Mathematics    & 152 & 889 \\
 & Code-Drawing   & 100 & 551 \\
\midrule
\addlinespace[5.0625pt]

\multirow{3}{*}{\parbox{2.6cm}{\centering\textsc{Engineering \&\\ Technology}}}
 & Computer-Science & 80  & 5231 \\
 & Mechanical       & 82  & 917  \\
 & Electronics      & 101 & 619  \\
\end{tabular}
}}
\end{minipage}
\hfill
\begin{minipage}[t]{0.49\textwidth}
\centering
\vspace{0pt}
{
\tablestyle{1.5pt}{1}{
\begin{tabular}{l l r r}
\toprule
\textbf{Knowledge Domain} & \textbf{Subtype} & \textbf{Query} & \textbf{Gallery} \\
\midrule

\multirow{4}{*}{\parbox{2.6cm}{\centering\textsc{Visual\\ Cognition}}}
 & Daily-Life     & 236 & 5248 \\
 & Fine-Grained   & 92  & 2101 \\
 & Spatial-Cogmap & 81  & 1270 \\
 & Spatial-3D     & 131 & 7500 \\
\midrule

\multirow{5}{*}{\parbox{2.6cm}{\centering\textsc{Humanity \&\\ Social\\Science}}}
 & Economics      & 80  & 2614 \\
 & Comic          & 87  & 3468 \\
 & Art            & 100 & 2101 \\
 & Calendar       & 80  & 378  \\
 & Metro          & 80  & 298  \\
\end{tabular}
}}
\end{minipage}

\par\vspace{0pt}
\begingroup
\footnotesize
\renewcommand{\arraystretch}{1.08}
\setlength{\tabcolsep}{0pt}
\makebox[\textwidth][l]{%
\begin{tabular}{@{}c@{}}
\midrule
\makebox[\dimexpr\textwidth-5.621pt\relax][c]{%
  \textbf{5 Domains}\hspace{2.5em}%
  \textbf{17 Subtypes}\hspace{2.5em}%
  \textbf{1,750 Queries}\hspace{2.5em}%
  \textbf{36,858 Gallery Candidates}%
} \\
\bottomrule
\end{tabular}%
}
\endgroup
\end{table*}

Each query in ARK is associated with exactly one positive candidate and, in most cases, a set of carefully designed hard negatives that are semantically related but incorrect.
These negatives are intentionally constructed to discourage shortcut matching where they may partially satisfy the query while violating the required relations or reasoning steps.
As a result, correct retrieval cannot be achieved by relying on partial cues and instead requires following the full query and performing the intended reasoning, together with the necessary domain knowledge.

\subsection{Knowledge and Reasoning Axes}
\label{sec:knowledge_reasoning_axes}

ARK is structured along two orthogonal axes, \textit{i.e.}, \emph{knowledge domains} and \emph{reasoning skills}.
The knowledge axis captures the types of domain knowledge that retrieval instances rely on, while the reasoning axis reflects the nature and amount of inference required to resolve a query during retrieval.
This dual-axis design facilitates fine-grained analysis of model strengths and weaknesses from complementary perspectives and helps avoid conflating domain knowledge with reasoning capability.

\begin{figure*}[!tp]
\centering
    \includegraphics[width=\linewidth]{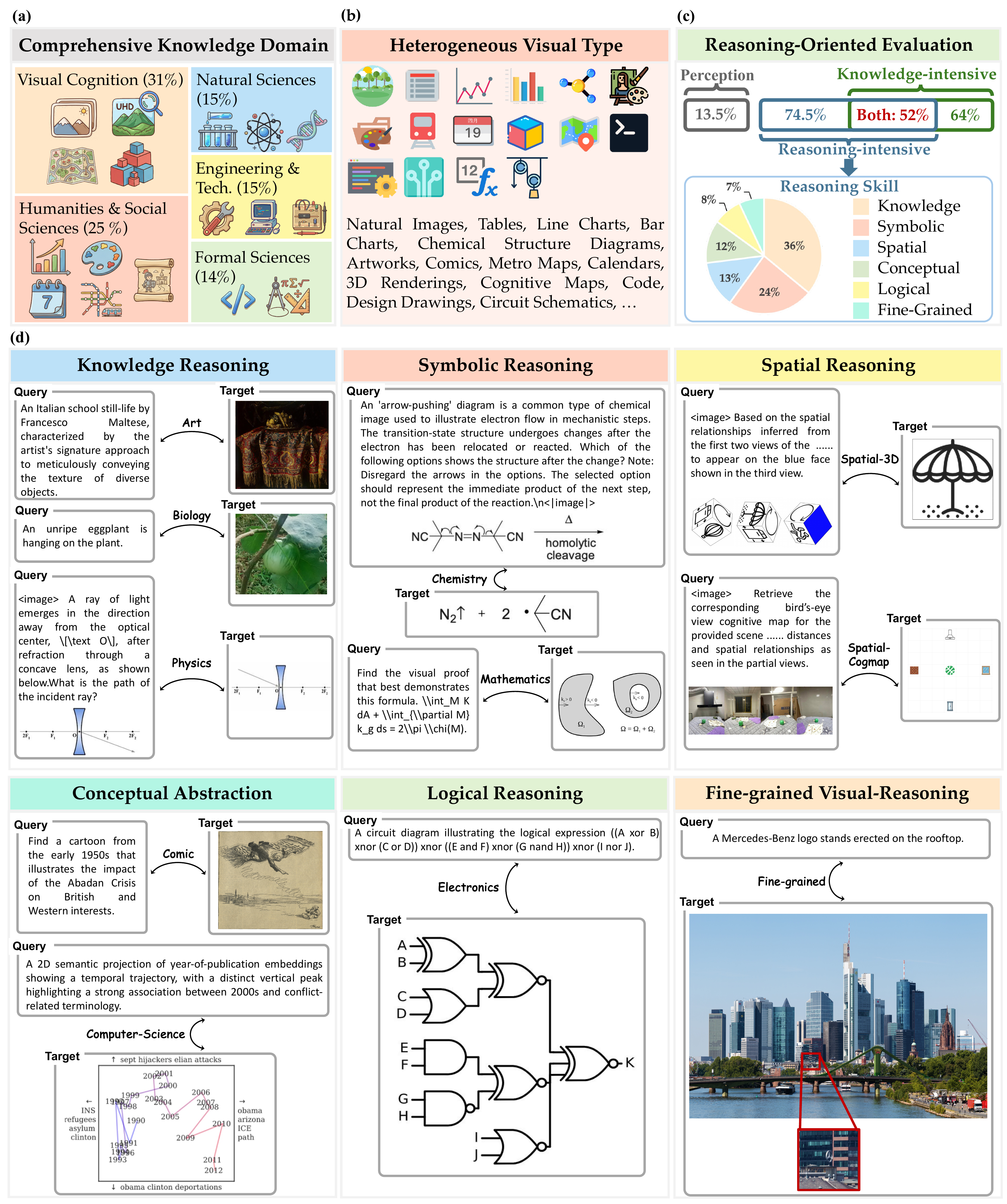}
  \caption{
ARK is a dual-axis benchmark for multimodal retrieval.
(a)–(c) summarize three key characteristics:
(a) \textbf{Comprehensive knowledge domains}: it covers five broad knowledge domains and 17 fine-grained subtypes,
(b) \textbf{Heterogeneous visual types}: queries and candidates span diverse visual formats beyond everyday imagery, and
{
(c) \textbf{Reasoning-oriented evaluation}: each instance in ARK is first labeled by its overall cognitive demand as \textit{perception-only}, \textit{knowledge-intensive}, \textit{reasoning-intensive}, or \textit{both knowledge- and reasoning-intensive}.
Reasoning-intensive instances are further annotated with six reasoning-skill categories, enabling fine-grained analysis of model behavior.
Note that a single instance may involve multiple reasoning skills.
}
(d) {Examples:} representative instances of each reasoning skill.
For reference, the primary reasoning skills and visual data types for each knowledge subtype are summarized in Table~\ref{tab:ark_data_source}.
}
  \label{fig:fig2}
\end{figure*}

\begin{figure*}[t]
    \centering
    \includegraphics[width=\linewidth]{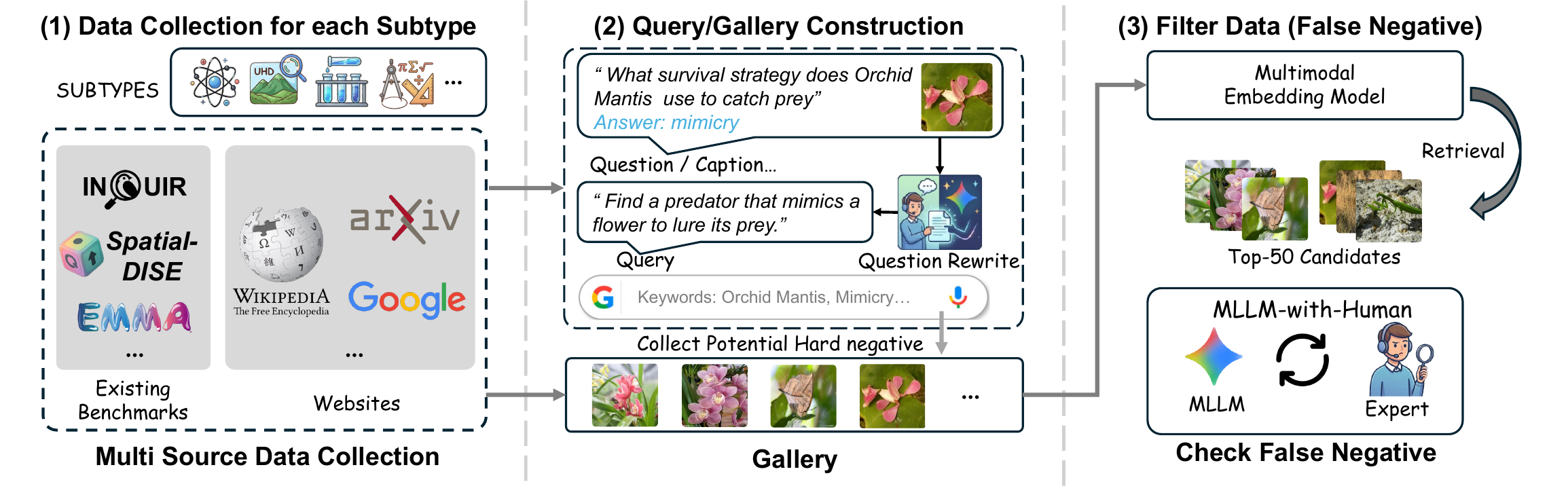}
  \caption{Illustration of the data curation process for ARK.}
    \label{fig:data_curation_process}
\end{figure*}

\paragraph{Knowledge axis.}
Table~\ref{tab:ark_knowledge_stats} summarizes ARK from the knowledge perspective, comprising 17 subtypes across five knowledge domains, with 1,750 queries and 36,858 gallery candidates in total.
Along this axis, ARK is organized into five meta-domains:
(i) \textit{Visual Cognition} serves as a perceptual baseline. It evaluates everyday recognition and fine-grained perception of subtle details in high-resolution images, aligning with recent interest in ``thinking with images''~\cite{openai2025thinkingwithimages,zheng2025deepeyes}. It also covers spatial perception over structured visual objects;
(ii) \textit{Natural Science}, which requires scientific knowledge of concepts and phenomena (\textit{e.g.}, biology, physics, and chemistry);
(iii) \textit{Formal Science}, which emphasizes symbolic and formal representations, including math problems and code-specified graphics;
(iv) \textit{Humanities \& Social Science}, which involves culturally and socially grounded content, including history-, art-, and society-related materials;
(v) \textit{Engineering \& Technology}, which covers technical domains such as computer science and engineering schematics, emphasizing understanding of technical diagrams and designs. For example, it includes extracting high-level conclusions from experimental results by identifying empirical trends such as \emph{scaling law}.
Across domains, ARK spans 16 heterogeneous visual types
including natural images, tables, line/bar charts, chemical structure diagrams, artworks, comics, metro maps, calendars, 3D renderings, cognitive maps, code, design drawings, circuit schematics, symbolic mathematical plots, and physics phenomenon illustrations, enabling evaluation beyond common-sense visual understanding (Fig.~\ref{fig:fig2}(b)).
Although modest compared with large-scale retrieval datasets, ARK is manually curated for reasoning-oriented evaluation: each instance requires query construction, positive-target verification, hard-negative validation, and reasoning-label annotation. Its scale is therefore comparable to recent reasoning-focused retrieval benchmarks~\cite{mrmr,weller2025followir,su2024bright,dong2025mmdocir}, where careful per-instance curation is more labor-intensive than collecting surface-level semantic matching pairs.
More details and examples for every subtype are provided in Appendix~\ref{appendix:dataset_construction_details}.

\paragraph{Reasoning axis.}

Fig.~\ref{fig:fig2}(c) reports the distribution of instances by required reasoning skills, and Fig.~\ref{fig:fig2}(d) provides representative examples for each category.
Along this axis, ARK organizes instances into six reasoning categories:
(i) \textit{Knowledge Reasoning} covers queries that require inference grounded in background knowledge, beyond mere recognition or direct factual retrieval.
(ii) \textit{Spatial Reasoning} requires understanding geometric and topological relations, such as spatial layouts in cognitive maps and metro networks, as well as 3D object projection, folding, and rotation.
(iii) \textit{Fine-grained Visual Reasoning} requires detail-sensitive visual inference in high-resolution images, where the target can only be identified by subtle, localized cues~\cite{yang2026endowing}. This capability closely aligns with the recent emphasis on ``thinking with images'' for careful visual inspection and verification~\cite{openai2025thinkingwithimages}.
(iv) \textit{Logical Reasoning} involves structured deduction over conditions and relations (\textit{e.g.}, conjunction/disjunction/negation, implication, and Boolean expressions such as AND/OR/NOT), where correct retrieval depends on satisfying explicit logical constraints rather than surface-level semantic cues.
(v) \textit{Symbolic Reasoning} focuses on interpreting structured symbolic representations and grounding them into the corresponding visual structures, such as mapping mathematical expressions to plots, code-like specifications to diagrams, or reasoning over chemical structure diagrams to infer the resulting molecular structure.
(vi) \textit{Conceptual Abstraction} evaluates higher-level inference over patterns, analogies, and inductive generalization, such as interpreting historical cartoons or summarizing high-level conclusions from experimental results in scientific papers.

Importantly, each retrieval instance in ARK is annotated along both knowledge and reasoning axes, enabling fine-grained analysis across knowledge, reasoning, and their interactions.
In addition, each instance is labeled by its overall cognitive demand as
\textit{perception-only}, \textit{knowledge-intensive}, \textit{reasoning-intensive}, or \textit{both knowledge- and reasoning-intensive}.
Fig.~\ref{fig:fig2}(c) summarizes the distribution and Table~\ref{tab:ark_data_source} reports the primary reasoning skills and visual data types associated with each knowledge subtype.

\subsection{Data Curation Process}
\label{sec:data_curation_process}

We curate ARK through a multi-stage pipeline that combines taxonomy-driven collection, automated refinement, and human verification, as illustrated in Fig.~\ref{fig:data_curation_process}.

\textbf{Taxonomy-driven collection.}
We first define a taxonomy spanning five knowledge domains and six reasoning dimensions.
Guided by this framework, we curate the gallery by selecting 17 subtypes across the five domains and collecting high-quality multimodal query--target and question--answer pairs from authoritative sources and expert-level benchmarks, forming the initial pool of positive pairs and gallery items.

\textbf{Reasoning-aligned refinement and quality control.}
{
After collecting the initial pairs, we use multimodal LLMs~\cite{gemini} to rewrite the original queries and questions, ensuring better alignment with the intended knowledge constraints and reasoning requirements.
For query--target pairs, we rewrite  queries to remove trivial cues (\textit{e.g.}, overly explicit keywords) while preserving the event background and the core reasoning intent.
For question--answer pairs,  we convert the original questions into retrieval-style queries, keeping  the underlying reasoning premises and constraint conditions required to identify the correct evidence.
To discourage shortcut matching during evaluation, we construct targeted hard negatives whenever feasible.
Specifically, for each rewritten query, we first mine visually or semantically similar candidates via web search, guided by keywords and topical cues suggested by the large model~\cite{gemini}, and then manually select distractors that share superficial cues with the positive target while violating the underlying conditions.
To further reduce false negatives~\cite{lin2026learning,li2026learning}, we embed each query with an up-to-date retriever~\cite{seed16embeddingapi}, retrieve its top-$50$ nearest candidates, and manually remove ambiguous items, spurious negatives, and near-duplicates.
}

{

\textbf{Human-in-the-loop curation.}
ARK construction involves two groups of human participants with distinct roles.
First, trained contributors and domain-aware reviewers, including PhD students and undergraduate annotators from biology, mathematics, computer science, and related disciplines, participated in data collection, query refinement, hard-negative construction, and consistency checking.
Second, reasoning-related labels, including cognitive demand and reasoning skills, were independently annotated by six PhD annotators under unified guidelines, with disagreements resolved through panel adjudication.
This separation supports domain-aware data construction while promoting consistent multi-label annotation~\cite{koumll,koubldl,kou2025nips} of reasoning skills.
Details of the annotation protocol and consistency analysis are provided in Appendix~\ref{appendix:annotation_protocol}.

}

\begin{table*}[t]
\caption{Domain-wise Recall@1 on ARK (knowledge axis). Columns correspond to ARK's five knowledge domains and corresponding subtypes. ``Avg.'' denotes the macro-average across all subtypes.
Models within each group are ordered by parameter size. The \textbf{bold} and \underline{underlined} numbers indicate the best and second-best results within each subtype, respectively.
}
\centering
\renewcommand{\arraystretch}{1.1}
\resizebox{\textwidth}{!}
{
{
\begin{tabular}{l|cccc|ccc|cc|ccccc|ccc|c}
\toprule
\multirow{2}{*}{\textbf{Model}} & \multicolumn{4}{c|}{\textbf{Visual Cognition}}
& \multicolumn{3}{c|}{\textbf{Natural Science}}
& \multicolumn{2}{c|}{\textbf{Formal Science}}
& \multicolumn{5}{c|}{\textbf{Humanities \& Social Science}}
& \multicolumn{3}{c|}{\textbf{Engineering \& Technology}}
& \multirow{2}{*}{\textbf{Avg.}} \\
 & Daily. & Fine. & SCog. & S3D. & Bio. & Phy. & Che. & Mat. & Cod. & Eco. & Comic. & Art & Cal. & Met. & Compu. & Mecha. & Ele. &  \\
\midrule
\multicolumn{19}{c}{\textit{\textbf{Text Embedding Models}}} \\
\midrule
BGE-M3-0.6B & 37.29 & 2.17 & 0.00 & 0.00 & 32.53 & 15.00 & 0.95 & 3.29 & 2.00 & 36.25 & 1.15 & 2.00 & 6.25 & 7.50 & 25.00 & 9.76 & 0.99 & 10.71 \\
Diver-Embed-1.7B & 35.59 & 2.17 & 0.00 & 0.00 & 26.51 & \underline{25.00} & 2.86 & 11.84 & 2.00 & 23.75 & 1.15 & 4.00 & \underline{8.75} & 6.25 & 38.75 & 4.88 & 0.00 & 11.38 \\
Diver-Embed-4B & 46.19 & 1.09 & 0.00 & 0.76 & 39.76 & 22.50 & 2.86 & \textbf{14.47} & 2.00 & \textbf{55.00} & 2.30 & 5.00 & 6.25 & 10.00 & 46.25 & 10.98 & 0.00 & 15.61 \\
Qwen3-Embed-4B & 42.37 & 2.17 & \textbf{3.70} & 0.76 & 27.71 & 21.25 & 1.90 & 10.53 & 3.00 & 36.25 & 1.15 & 4.00 & 5.00 & 3.75 & 33.75 & 10.98 & 2.97 & 12.43 \\
\midrule
Qwen3-Embed-8B & 46.19 & 1.09 & 0.00 & 0.76 & 40.96 & 22.50 & 4.76 & \underline{12.50} & 4.00 & 47.50 & 2.30 & 4.00 & 7.50 & 2.50 & 46.25 & 8.54 & 1.98 & 14.90 \\
ReasonIR-8B & 40.25 & 1.09 & 0.00 & 0.00 & 28.92 & 20.00 & 1.90 & 10.53 & 1.00 & 42.50 & 4.60 & 1.00 & 3.75 & 13.75 & 36.25 & 9.76 & 0.99 & 12.72 \\
\midrule
\multicolumn{19}{c}{\textit{\textbf{Multimodal Embedding Models}}} \\
\midrule
CLIP-0.4B & 37.29 & 0.00 & \underline{1.23} & 3.05 & 26.51 & 7.50 & 3.81 & 1.32 & 1.00 & 7.50 & 3.45 & 10.00 & 5.00 & 3.75 & 7.50 & 7.32 & 3.96 & 7.66 \\
BGE-VL-0.4B & 52.97 & 0.00 & 0.00 & 0.00 & 32.53 & 0.00 & 0.00 & 1.32 & 0.00 & 6.25 & 0.00 & 5.00 & 1.25 & 2.50 & 10.00 & 7.32 & 1.98 & 7.12 \\
SigLIP2-0.9B & 68.22 & 2.17 & 0.00 & 1.53 & 33.73 & 3.75 & 1.90 & 4.61 & 0.00 & 13.75 & 5.75 & 3.00 & 6.25 & 1.25 & 20.00 & 14.63 & 0.99 & 10.68 \\
\midrule
MetaCLIP2-2B & 60.17 & 3.26 & 0.00 & 3.05 & 33.73 & 20.00 & 1.90 & 1.32 & 3.00 & 21.25 & 3.45 & \textbf{20.00} & 1.25 & 3.75 & 12.50 & 13.41 & 2.97 & 12.06 \\
VLM2VecV2-2B & 43.22 & 1.09 & 0.00 & 0.76 & 24.10 & 6.25 & 2.86 & 7.89 & 4.00 & 32.50 & 2.30 & 4.00 & 2.50 & 10.00 & 22.50 & 7.32 & 0.00 & 10.08 \\
Ops-MM-Embed-v1-2B & 63.14 & 1.09 & \underline{1.23} & 2.29 & 38.55 & 10.00 & 3.81 & 6.58 & 3.00 & 21.25 & 8.05 & 7.00 & 3.75 & 13.75 & 23.75 & 9.76 & 2.97 & 12.94 \\
Qwen3-VL-Embed-2B & 65.25 & 1.09 & 0.00 & \underline{4.58} & 27.71 & 21.25 & 0.95 & 7.24 & 4.00 & 38.75 & 4.60 & 3.00 & 2.50 & 6.25 & 26.25 & 8.54 & 3.96 & 13.29 \\
\midrule
LamRA-Ret-7B & 60.17 & \textbf{5.43} & 0.00 & 0.00 & 38.55 & 1.25 & 2.86 & 9.87 & 6.00 & 35.00 & 3.45 & 8.00 & 2.50 & 5.00 & 5.00 & 7.32 & 0.99 & 11.26 \\
RzenEmbed-7B & \textbf{69.92} & 2.17 & \underline{1.23} & 1.53 & 32.53 & 10.00 & \underline{5.71} & 11.84 & 7.00 & 43.75 & 2.30 & 4.00 & 6.25 & 7.50 & \underline{48.75} & \underline{15.85} & 4.95 & 16.19 \\
Ops-MM-Embed-v1-7B & 62.29 & 1.09 & \underline{1.23} & \textbf{5.34} & \underline{42.17} & 15.00 & 3.81 & 9.21 & 6.00 & 41.25 & 6.90 & 13.00 & 5.00 & \textbf{17.50} & 42.50 & 12.20 & \underline{5.94} & 17.08 \\
Qwen3-VL-Embed-8B & \underline{69.07} & 0.00 & 0.00 & 3.82 & 39.76 & \textbf{26.25} & 3.81 & 9.87 & \underline{14.00} & \underline{51.25} & \underline{11.49} & 4.00 & 3.75 & 7.50 & 46.25 & 13.41 & \textbf{6.93} & \underline{18.30} \\
\midrule
EVA-CLIP-18B & 63.98 & 2.17 & 0.00 & 2.29 & \textbf{43.37} & 13.75 & \textbf{6.67} & 0.66 & 3.00 & 18.75 & 0.00 & \underline{18.00} & 3.75 & 3.75 & 16.25 & \textbf{21.95} & 3.96 & 13.08 \\
Seed1.6-embedding & 65.25 & \underline{4.35} & 0.00 & 3.82 & 31.33 & 18.75 & 3.81 & 11.84 & \textbf{18.00} & 45.00 & \textbf{12.64} & 10.00 & \textbf{13.75} & \underline{16.25} & \textbf{57.50} & 12.20 & 3.96 & \textbf{19.32} \\

\bottomrule
\end{tabular}
}
}
\label{table:knowledge_score}
\end{table*}

\section{Experiments}

We first evaluate embedding-based retrievers on ARK, covering both text-only and multimodal models. We then examine two inference-time enhancement strategies, query rewriting and re-ranking, and further evaluate agentic retrievers in a multi-round retrieval setting in Appendix~\ref{appendix:agent_multiround}. Details of evaluated models are provided in Appendix~\ref{appendix:evaluated_models}.
We use Recall@1 as the primary metric, as it directly measures whether the correct target is ranked first. To provide a more complete view of retrieval quality, we also report Recall@5, Recall@10, nDCG@5, nDCG@10, and nDCG@20 in Appendix~\ref{appendix:results_metrics}.

\subsection{Evaluation across Knowledge Domains}

We summarize the overall evaluation results of all investigated retrieval baselines on ARK in Table~\ref{table:knowledge_score}.
All experiments are conducted within each individual subtype using its own dedicated retrieval corpus, ensuring that retrieval difficulty and domain coverage are controlled per subtype.
For each subtype, we report Recall@1, together with the macro-average across all 17 subtypes.
For text retrievers, we first convert visual inputs into textual descriptions using Qwen3-VL-8B-Instruct~\cite{Qwen3-VL} (with the prompt shown in Table~\ref{prompt:caption_generation}), and then perform retrieval in the text embedding space based on the generated captions.
Based on these results, we draw several key observations:

\textbf{Observation 1: ARK is substantially more challenging and diagnostic than prior retrieval benchmarks.}
Performance on ARK remains low even for the strongest retrievers.
The best embedding model, Seed1.6~\cite{seed16embeddingapi}, achieves only 19.32 R@1 in macro-average.
This is in sharp contrast to its reported 75.59 R@1 on the embedding-focused MMEB-v2 leaderboard~\cite{mmeb_leaderboard}, where Qwen3-VL-Embed~\cite{li2026qwen3} also reaches 77.82 R@1.
A similar gap appears for text retrieval: Qwen3-Embed-8B~\cite{qwen3embedding} obtains 54.1 nDCG@10 on the reasoning-oriented MRMR benchmark~\cite{mrmr}, but only 24.27 nDCG@10 on ARK (Table~\ref{table:knowledge_score_ndcg_10}).
These results suggest that ARK is not well captured by surface-level semantic matching and exposes failure modes that are less visible in existing benchmarks.

\textbf{Observation 2: Performance varies sharply across knowledge domains, exposing clear bottlenecks.}
Across models, retrieval is relatively strong on daily-life natural images but drops markedly on domains requiring fine-grained perception or spatial grounding. The Fine-Grained (high-resolution image), Spatial-Cogmap, and Spatial-3D subsets are among the hardest, indicating model brittleness under detail-sensitive visual inspection and non-trivial spatial reasoning.
Beyond visual cognition, performance is highly subtype-dependent: models perform reasonably well on some knowledge-heavy subtypes such as Biology, Economics, and Computer Science, but remain weak on Chemistry, Mathematics, and Code-Drawing.
This suggests that ARK's bottlenecks are not determined by broad knowledge domains alone, but by the interaction between domain knowledge, visual format, and reasoning demands.

\textbf{Observation 3: Model rankings are not stable across domains, reflecting differences in pretraining bias and representational strengths.}
While several multimodal embedding models rank highly on the overall average, their advantages are not consistent across subtypes. For instance, Qwen3-VL-Embed performs better than Seed1.6 on knowledge-heavy categories such as biology and physics, but is weaker on high-resolution subsets such as Metro and Fine-Grained.
This variability indicates that ``best overall'' does not translate to ``best everywhere'', and ARK’s domain-level breakdown provides more actionable diagnosis than single-number evaluations.

\textbf{Observation 4: Transforming multimodal retrieval to text-space via captioning yields a strong baseline, but is limited by caption fidelity.}
Reducing multimodal retrieval to text by captioning images and retrieving in a text embedding space~\cite{visa} can be competitive on some subtypes, as shown by the results in the top part of Table~\ref{table:knowledge_score}.
However, this pipeline is computationally costly and frequently discards critical visual evidence (\textit{e.g.}, fine-grained details, spatial relations, and diagram structure).
This limitation helps explain why reasoning-oriented text retrievers (ReasonIR~\cite{reasonir}, Diver~\cite{diver}) do not consistently dominate on ARK: when the caption omits key cues, downstream reasoning has insufficient evidence.

\begin{table*}[t]
\caption{Recall@1 on ARK by cognitive demand and reasoning skill (reasoning axis). ``Both'' denotes instances that are both knowledge-intensive and reasoning-intensive.
}

\renewcommand{\arraystretch}{1.1}
\centering
\resizebox{\textwidth}{!}{
\tablestyle{2.5pt}{1.1}{
\begin{tabular}{l|cccc|ccccccc}
\toprule

\multirow{2}{*}{\textbf{Model}}
& \multicolumn{4}{c|}{\textbf{Cognitive Demand}}
& \multicolumn{7}{c}{\textbf{Reasoning Skill}} \\
& \makecell{\textbf{Perception-}\\\textbf{Only}}
& \makecell{\textbf{Knowledge-}\\\textbf{Intensive}}
& \makecell{\textbf{Reasoning-}\\\textbf{Intensive}}
& \textbf{Both}
& \makecell{\textbf{Knowledge}\\\textbf{Reasoning}}
& \makecell{\textbf{Conceptual}\\\textbf{Abstraction}}
& \makecell{\textbf{Fine-Grained}\\\textbf{Visual Reasoning}}
& \makecell{\textbf{Logical}\\\textbf{Reasoning}}
& \makecell{\textbf{Spatial}\\\textbf{Reasoning}}
& \makecell{\textbf{Symbolic}\\\textbf{Reasoning}}
& \textbf{Average}
\\

\midrule
\multicolumn{12}{c}{\textit{\textbf{Text Embedding Models}}} \\
\midrule

BGE-M3-0.6B & 37.29 & 12.09 & 7.89 & 10.12 & 11.70 & 18.34 & 3.08 & 4.29 & 2.31 & 3.90 & 7.27 \\
Diver-Embed-1.7B & 36.02 & 14.86 & 9.88 & 13.20 & 14.84 & 21.83 & 2.31 & 5.52 & 2.69 & 7.79 & 9.16 \\
Diver-Embed-4B & 46.19 & \underline{17.73} & 11.94 & \underline{15.95} & \underline{18.12} & 27.07 & 3.85 & 6.13 & 1.54 & \underline{9.31} & 11.00 \\
Qwen3-Embed-4B & 41.53 & 12.09 & 7.73 & 9.90 & 11.98 & 17.90 & 1.54 & 4.29 & 2.69 & 5.19 & 7.26 \\
\midrule
Qwen3-Embed-8B & 45.76 & 15.67 & 9.57 & 13.09 & 15.98 & 23.58 & 0.77 & 4.91 & 1.54 & 6.71 & 8.92 \\
ReasonIR-8B & 40.25 & 15.31 & 11.18 & 14.41 & 16.41 & 28.38 & 3.85 & 3.07 & 4.23 & 7.58 & 10.59 \\
\midrule
 \multicolumn{12}{c}{\textit{\textbf{Multimodal Embedding Models}}} \\
\midrule
CLIP-0.4B & 37.29 & 6.54 & 4.13 & 4.84 & 5.56 & 5.68 & 0.77 & 3.68 & 3.08 & 2.38 & 3.52 \\
BGE-VL-0.4B & 53.39 & 5.10 & 2.60 & 3.52 & 4.14 & 3.93 & 1.54 & 1.84 & 0.38 & 1.08 & 2.15 \\
SigLIP2-0.9B & 68.22 & 8.24 & 5.28 & 6.93 & 8.70 & 8.73 & 2.31 & 3.07 & 0.77 & 2.38 & 4.33 \\
\midrule
MetaCLIP2-2B & 60.59 & 10.38 & 6.58 & 8.14 & 9.56 & 11.79 & 3.85 & 2.45 & 3.08 & 2.16 & 5.48 \\
VLM2VecV2-2B & 44.07 & 8.86 & 6.66 & 8.25 & 10.41 & 16.16 & 3.85 & 1.23 & 2.31 & 4.11 & 6.34 \\
Ops-MM-Embed-v1-2B & 63.98 & 10.92 & 7.73 & 9.24 & 11.13 & 13.97 & \underline{4.62} & 3.68 & 4.23 & 4.55 & 7.03 \\
Qwen3-VL-Embed-2B & 65.68 & 12.00 & 8.58 & 11.11 & 12.98 & 23.58 & 3.08 & 4.29 & 3.08 & 5.19 & 8.70 \\
\midrule
LamRA-Ret-7B & 59.32 & 9.67 & 6.81 & 8.69 & 10.27 & 14.41 & 4.62 & 2.45 & 1.15 & 5.41 & 6.39 \\
RzenEmbed-7B & \textbf{70.34} & 15.04 & 10.57 & 13.75 & 16.12 & 22.71 & 4.62 & 6.13 & 2.31 & 8.01 & 9.98 \\
Ops-MM-Embed-v1-7B & 63.56 & 16.11 & \underline{12.02} & 14.63 & 17.12 & 23.58 & \textbf{6.92} & 6.13 & \underline{5.77} & 6.93 & 11.08 \\
Qwen3-VL-Embed-8B & \underline{68.64} & 17.73 & 11.94 & 15.95 & 17.55 & \underline{30.13} & 0.77 & \underline{6.75} & 3.85 & 8.87 & \underline{11.32} \\
\midrule
EVA-CLIP-18B & 63.98 & 11.46 & 6.43 & 8.14 & 9.42 & 9.61 & 2.31 & 3.68 & 2.31 & 3.46 & 5.13 \\
Seed1.6-embedding & 62.29 & \textbf{22.33} & \textbf{16.85} & \textbf{21.70} & \textbf{22.08} & \textbf{33.33} & 4.35 & \textbf{10.71} & \textbf{6.19} & \textbf{15.16} & \textbf{15.30} \\
\bottomrule
\end{tabular}
}
}
\label{table:reasoning_score}
\end{table*}

\subsection{Evaluation along Reasoning Skill}

To better understand where current retrievers succeed or fail in reasoning-intensive retrieval, we analyze ARK performance from a reasoning-centric perspective.
Specifically, we classify instances along  {cognitive demand} and {reasoning skill}.
{Cognitive demand} categorizes instances as \textit{perception-only}, \textit{knowledge-intensive}, \textit{reasoning-intensive}, or \textit{both knowledge- and reasoning-intensive}, reflecting whether a query is mainly solvable by knowledge recall, multi-step inference, or requires both.
{Reasoning skill} specifies the cognitive operation involved, covering \textit{knowledge reasoning}, \textit{spatial reasoning}, \textit{logical reasoning}, \textit{symbolic reasoning}, \textit{fine-grained visual reasoning}, and \textit{conceptual abstraction}.
Based on the results in Table~\ref{table:reasoning_score}, we have the following observations:

\textbf{Observation 5: Reasoning-intensive retrieval is substantially harder than knowledge-intensive retrieval.}
From the cognitive-demand breakdown in Table~\ref{table:reasoning_score}, \textit{perception-only} instances are consistently the easiest: most models perform strongly when retrieval reduces to basic visual recognition and cross-modal alignment.
However, performance degrades noticeably on \textit{knowledge-intensive} instances and drops further on \textit{reasoning-intensive} instances, suggesting that multi-step inference remains a critical bottleneck beyond domain-knowledge matching.
Instances labeled as \textit{both knowledge- and reasoning-intensive} also remain challenging overall, underscoring the compounded difficulty when retrieval requires knowledge and reasoning jointly.

\textbf{Observation 6: Models exhibit a consistent performance pattern across reasoning skills.}
As shown in Table~\ref{table:reasoning_score}, most retrievers follow a similar trend across skill categories: they perform relatively better on {knowledge reasoning} and {conceptual abstraction}, but struggle markedly with {fine-grained visual reasoning} and {spatial reasoning}. Performance on {logical} and {symbolic} reasoning lies in between, suggesting substantial room for improvement beyond scaling alone.
These results suggest that future work may benefit from integrating vision-centric reasoning mechanisms such as ``thinking with images''~\cite{openai2025thinkingwithimages,li2026reliable} approaches into embedding-based retrievers, so that fine-grained and spatial visual evidence can be more effectively preserved and exploited during retrieval.

\textbf{Observation 7: Model scale correlates with stronger retrieval, but does not eliminate reasoning bottlenecks.}
Tables~\ref{table:knowledge_score} and \ref{table:reasoning_score} show a clear trend that larger models tend to achieve higher Recall@1 across both the knowledge and reasoning axes, suggesting that increased capacity improves alignment.
However, the benefits are uneven: gains are most visible on knowledge-intensive queries and conceptual abstraction, while fine-grained visual, spatial, logical, and symbolic reasoning remain difficult even for the largest models.
This suggests that scaling alone is unlikely to close these gaps, pointing to the need for additional architectural or training improvements beyond model size.

\subsection{Query Rewriting and Re-ranking}

We study two inference-time interventions for improving retrieval: \emph{query rewriting} and \emph{re-ranking}.
For {query rewriting}, we use GPT-5.2~\cite{openai_gpt52} to transform the original query into a semantically equivalent but more explicit form that highlights reasoning cues (prompt provided in Table~\ref{prompt:query_rewrite}).
For {re-ranking}, we first retrieve the top-50 candidates using an embedding retriever, and then apply the reranker for each query--candidate pair to produce a refined ranking.
Based
on results in Tables~\ref{table:Reranker and Query Rewrite} and~\ref{table:Reranker and Query Rewrite for reasoning}, we draw the following observations:

\textbf{Observation 8: Query rewriting and re-ranking generally improve retrieval, and they are complementary.}
As shown in Table~\ref{table:Reranker and Query Rewrite}, query rewriting generally improves the Seed1.6 embedding baseline, increasing R@1 from 19.32 to 23.42 on the knowledge axis and from 15.30 to 17.49 on the reasoning axis.
Re-ranking further improves retrieval when the reranker has sufficient capacity: Jina-Reranker and Qwen3-VL-Reranker outperform the embedding-only baseline, while the 600M BGE-Reranker underperforms it.
Combining query rewriting with Qwen3-VL-Reranker achieves the best overall results on most metrics.
These results suggest complementary effects: rewriting strengthens query-side reasoning, while capable rerankers enable stronger pairwise inference over the top retrieved candidates.

\textbf{Observation 9: Query rewriting helps most when queries under-specify instructions.}
As shown in Table~\ref{table:Reranker and Query Rewrite for reasoning}, rewriting yields the largest gains for {logical} and {knowledge reasoning}, where queries often contain implicit rules or abstract conditions that can be made more explicit through rewriting.
In contrast, improvements are limited for fine-grained visual and spatial reasoning, where the main bottleneck lies in extracting subtle visual details or precise geometric relations from candidates rather than clarifying the query.
This suggests that beyond query-side rewriting, enriching gallery items with {query-guided, reasoning-aware} descriptions~\cite{visa} before re-ranking could be a promising direction.

\section{Conclusion}
We introduce ARK, a knowledge- and reasoning-intensive benchmark for multimodal retrieval that enables systematic diagnosis along both the knowledge and reasoning axes.
ARK spans 17 knowledge subtypes and six reasoning categories, and pairs most queries with targeted hard negatives to reduce shortcut matching, thereby revealing failure modes that are largely masked by conventional semantic-matching benchmarks.
Evaluations on 25 representative retrievers show a pronounced gap between knowledge-intensive and reasoning-intensive retrieval, with fine-grained visual and spatial reasoning remaining persistent bottlenecks.
Although interventions such as query rewriting, re-ranking, and agentic retrieval provide consistent improvements, substantial headroom remains for advancing reasoning-aware multimodal retrieval.
We hope ARK will support more diagnostic evaluation and foster progress toward retrieval models that better integrate domain knowledge with visual reasoning.

\bibliographystyle{unsrt}
\bibliography{example_paper}

\newpage
\appendix

\noindent{\Large\bfseries\textsc{Appendix}}

{

\vspace{0.8em}
\noindent In this appendix, we present:

\begin{enumerate}[leftmargin=1.6em,itemsep=0.35em,topsep=0.6em]
\item Comparison with Existing Benchmarks (Section~\ref{appendix:comparison_existing_benchmarks})
\item Dataset Details (Section~\ref{appendix:dataset_construction_details})
\item Details of Evaluated Models (Section~\ref{appendix:evaluated_models})
\item Evaluation with Query Rewriting and Re-ranking (Section~\ref{appendix:rewrite_rerank})
\item Results on ARK under Different Evaluation Metrics (Section~\ref{appendix:results_metrics})
\item Bootstrap Uncertainty Analysis of Main Results (Section~\ref{appendix:bootstrap_ci})
\item Multi-round Retrieval Results for Agent (Section~\ref{appendix:agent_multiround})
\item Captioning Robustness and Prompt Details (Section~\ref{appendix:prompts})
\item Impact Statement (Section~\ref{appendix:impact_statement})
\item Annotation Protocol and Consistency Analysis for Reasoning-related Labels (Section~\ref{appendix:annotation_protocol})
\item {Case Study} (Section~\ref{appendix:case_study})
\end{enumerate}

}

\section{Comparison with Existing Benchmarks}
\label{appendix:comparison_existing_benchmarks}

\begin{table*}[h]
\centering
\caption{Comparison of multimodal retrieval benchmarks.
Benchmarks are grouped into three parts: (i) \emph{Top}: traditional multimodal retrieval benchmarks dominated by everyday natural images and semantic matching; (ii) \emph{Middle}: knowledge-oriented benchmarks that incorporate web-scale or expert content; (iii) \emph{Bottom}: recent benchmarks that explicitly probe reasoning-intensive retrieval.
For benchmarks containing mixed task formats, we report only the number of retrieval-related tasks.
}

\resizebox{\textwidth}{!}{
\tablestyle{3pt}{1.1}{
\newcommand{\cmark}{\ding{51}}
\newcommand{\xmark}{\ding{55}}
\begin{tabular}{l r r c c c c c c}
\toprule
\textbf{Benchmarks} & \textbf{\# Queries} & \textbf{\# Tasks} &
\makecell{\textbf{Knowledge-}\\\textbf{Intensive}} &
\makecell{\textbf{Reasoning-}\\\textbf{Intensive}} &
\makecell{\textbf{Multi-}\\\textbf{Domain}} &
\makecell{\textbf{Diverse  }\\\textbf{Reasoning Type}} &
\makecell{\textbf{Heterogeneous}\\\textbf{Visual Type}} &
\makecell{\textbf{Fine-grained}\\\textbf{Visual Reasoning}} \\
\midrule

COCO~\cite{COCO}& 25K & 1 & \\
Flickr30K~\cite{flickr}& 5, 000 & 1 &  \\
CIRR~\cite{CIRR} & 4,148 & 1 & \\
FashionIQ~\cite{fashioniq} & 60K & 1 \\
CIRCO~\cite{circo}& 1,020 & 1\\

\midrule
WebQA~\cite{webqa} & 7{,}540 & 1 & \cmark &  \\

EDIS~\cite{EDIS} &3,241  &1 & \cmark\\

InfoSeek~\cite{infoseek}& 1.35M& 2&\cmark \\

SciMMIR~\cite{scimmir}& 530K & 11
& \cmark&   &\cmark & &  \\
M-BEIR~\cite{uniir} & 190K & 10 & \cmark &   &\cmark &  \\
ViDoRe~\cite{vidore} & 3{,}810 & 2  &\cmark \\
MMEB~\cite{mmeb} & 12K & 12 & \cmark  & & \cmark & \\

\midrule

MRMR~\cite{mrmr} &1,435 & 23
& \cmark &\cmark & \cmark  &-- & --
\\
MR$^2$-Bench~\cite{mr2} &1,309 & 12
& \cmark &\cmark &\cmark  & -- & --
 &
\\
\textbf{ARK (Ours)} & {1,750} & {17}  & \cmark & \cmark  & \cmark & 6 & 16  & \cmark\\
\bottomrule
\end{tabular}
\label{tab:benchmark_comparison}
}}
\end{table*}

As shown in Table~\ref{tab:benchmark_comparison}, multimodal retrieval benchmarks have evolved from \emph{surface-level semantic matching} to \emph{knowledge-intensive retrieval}, and more recently to \emph{reasoning-intensive retrieval}.
Early multimodal retrieval benchmarks (top block) are largely built on natural images from daily-life scenarios, where relevance is often well captured by object-level semantics or short textual descriptions.
Subsequent benchmarks (middle block) expand the scope to web-scale or expert content, placing stronger emphasis on knowledge-intensive queries and multi-domain coverage, but still provide limited diagnostics of the \emph{reasoning process} required to resolve a query.
The most recent benchmarks (bottom block) begin to explicitly incorporate reasoning requirements, yet they typically inherit query styles from knowledge-heavy sources and cover only a narrow slice of reasoning phenomena.

ARK differs from prior benchmarks along several axes reflected in Table~\ref{tab:benchmark_comparison}.
First, ARK is the first multimodal retrieval benchmark that explicitly separates {knowledge} and {reasoning} as two evaluation axes, enabling clearer diagnosis of model strengths and failures from both perspectives.
In contrast, prior benchmarks lack an explicit taxonomy of reasoning skills, making it difficult to attribute errors to missing knowledge versus insufficient reasoning.
Second, ARK goes beyond natural images by covering 16 heterogeneous visual formats spanning both natural and structured visuals (\textit{e.g.}, line/bar charts, chemical structure diagrams, artworks, 3D renderings, code, and design drawings)
as listed in Table~\ref{tab:ark_data_source}.
This diversity requires models to ground queries in visual evidence rather than relying on surface-level semantics.
Third, ARK provides fine-grained reasoning annotations spanning a diverse set of reasoning types (\textit{i.e.}, knowledge, spatial, logical, symbolic, fine-grained visual reasoning, and conceptual abstraction).
In particular, ARK emphasizes richer vision-centric demands, including complex spatial inference and detail-sensitive visual inspection (e.g., identifying subtle objects in high-resolution images), which closely aligns with the recent interest in ``thinking with images''~\cite{openai2025thinkingwithimages,zheng2025deepeyes}.
Together, these design choices enable more targeted analysis than prior benchmarks that primarily probe limited forms of visual relations.

\begin{table*}[t]
\centering
\caption{Data sources of ARK benchmark.
ARK is structured by five knowledge domains and six reasoning skill categories, and covers 16 heterogeneous visual data types.
For instances adapted from existing retrieval or QA datasets, we carefully rewrite queries to better elicit the intended reasoning skills, and we curate targeted hard negatives whenever feasible to reduce shortcut matching.
}
\label{tab:ark_data_source}
\resizebox{\textwidth}{!}{
\tablestyle{4pt}{1.2}{
\begin{tabular}{c l l l l}
\toprule
\textbf{Meta-Task} & \textbf{Subtype} & \textbf{Visual Type} & \textbf{Reasoning Skill} & \textbf{Adapted From} \\
\midrule

\multirow{4}{*}{\parbox{2cm}{\centering\textsc{Visual\\Cognition}}}
& Daily-Life        & natural images                   & --                   & COCO~\cite{COCO} and Fashion200K~\cite{F200K} \\
& Fine-Grained      & natural images                   & fine-grained visual reasoning                   & DIV8K~\cite{div8k} \\
& Spatial-Cogmap    & cognitive maps                   & spatial reasoning                               & MindCube~\cite{yin2025spatial} \\
& Spatial-3D        & 3D images                        & spatial reasoning                               & Spatial-DISE-12K~\cite{huang2025spatial} \\
\midrule

\multirow{3}{*}{\parbox{2cm}{\centering\textsc{Natural\\Science}}}
& Biology           & natural images                   & -- & INQUIRE~\cite{inquire} \\
& Physics           & physics phenomenon illustrations & knowledge reasoning           & Manually collected \\
& Chemistry         & chemical structure diagrams       & knowledge reasoning, symbolic reasoning          & EMMA~\cite{emma} \\
\midrule

\multirow{2}{*}{\parbox{2cm}{\centering\textsc{Formal\\Science}}}
& Mathematics       & symbolic mathematical plots       & knowledge reasoning, symbolic reasoning          & MR$^2$-Bench~\cite{mr2} \\
& Code-Drawing      & code snippets, design drawings    & symbolic reasoning                               & ReMI~\cite{remi} \\
\midrule

\multirow{5}{*}{\parbox{2cm}{\centering\textsc{Humanity \& \\ Social\\Science}}}
& Economics         & tables, line and bar charts       & knowledge reasoning, conceptual abstraction      & MR$^2$-Bench~\cite{mr2} \\
& Comic             & comics                            & knowledge reasoning, conceptual abstraction      & Bulletin Editorial Cartoons~\cite{sherratt_trove_journals_2024} \\
& Art               & artworks                          & knowledge reasoning      & SemArt~\cite{semart} \\
& Calendar          & calendars                         & knowledge reasoning, logical reasoning           & Manually collected \\
& Metro             & metro maps                        & spatial reasoning, fine-grained visual reasoning             & Manually collected \\
\midrule

\multirow{3}{*}{\parbox{2cm}{\centering\textsc{Engineering \& \\ Technology}}}
& Computer-Science  & tables, line and bar charts, design drawings & knowledge reasoning, conceptual abstraction & arXiv~\cite{arxiv_website} \\
& Mechanical        & design drawings                   & knowledge reasoning & Mechanical mechanism~\cite{mechanisms} \\
& Electronics       & circuit schematics, tables         & logical reasoning, symbolic reasoning            & ElectroVizQA~\cite{ElectroVizQA} \\
\bottomrule
\end{tabular}
}}
\end{table*}

\section{Dataset Details}
\label{appendix:dataset_construction_details}

The overall curation pipeline of ARK has been described in Section~\ref{sec:data_curation_process}.
In this section, we provide construction details for each knowledge subtype, including (i) data sources, (ii) the visual data types of queries and gallery items, (iii) the primary reasoning skills targeted, and (iv) hard-negative construction when applicable. Overall statistics are summarized in Table~\ref{tab:ark_data_source}.

\subsection{Visual Cognition}

\textbf{Daily-Life.}
This topic focuses on text-to-image retrieval over everyday visual content and primarily evaluates basic perceptual recognition and visual-text alignment.
It serves as a baseline for multimodal retrieval without requiring domain knowledge or multi-step reasoning.
Data are constructed from images in COCO~\cite{COCO} and Fashion200K~\cite{F200K}, covering daily scenes, object interactions, and apparel combinations.
As shown in Fig.~\ref{fig: case_Daily_Life}, queries describe observable attributes such as object categories, colors, spatial arrangements, and clothing details (\textit{e.g.}, style, color, and texture), and retrieval can be resolved through accurate attribute matching.
Hard negatives are selected using embedding-based similarity retrieval followed by manual verification, resulting in visually similar but non-matching candidates that require precise perception rather than higher-level reasoning to disambiguate.

\textbf{Fine-Grained.}
This topic focuses on retrieval over ultra-high-resolution images and evaluates fine-grained visual reasoning over subtle and localized visual evidence, rather than coarse semantic recognition.
We construct this subset based on the DIV8K dataset~\cite{div8k}, which provides native ultra-high-resolution images with rich fine-grained details.
As shown in Fig.~\ref{fig: case_Fine_Grained}, the selected images have an average resolution of $5960\times3900$, in contrast to the daily-life subset, where images are typically around $365\times461$, substantially increasing perceptual complexity and the need for precise inspection of local regions.
Queries are natural language descriptions of small or easily overlooked objects and fine-grained visual attributes, often coupled with spatial constraints, requiring models to localize relevant regions and verify detailed conditions rather than relying on global appearance or object-level semantics.
This setting closely relates to recent interest in high-resolution visual reasoning and ``thinking with images,'' where correct retrieval depends on deliberate inspection and reasoning over visual details.
Hard negatives are manually curated by selecting images that partially satisfy the query (\textit{e.g.}, similar scene layout or object categories) but violate critical fine-grained constraints such as object presence, size, or precise location, ensuring that correct retrieval requires reasoning over local evidence instead of matching salient global features.

\textbf{Spatial-Cogmap.}
Inspired by MindCube~\cite{yin2025spatial}, this subtype targets the spatial reasoning ability of retrieval models, testing whether retrievers can infer global layout, topological relationships, and relative object positions from limited views, rather than relying on object recognition alone.
In this subtype, the retrieval task is defined as follows: given a small set of perspective-view images captured from limited viewpoints (e.g., four views sampled along a 360$^\circ$ camera rotation, as shown in Fig.~\ref{fig: case_Spatial_Cogmap}), the model must retrieve the corresponding bird’s-eye cognitive map from a large gallery.
Each cognitive map is rendered on a standardized $10\times10$ coordinate grid (top-left $[0,0]$, bottom-right $[9,9]$), where object locations are marked using a unified set of icons.
To prevent models from relying on shallow matching based on object co-occurrence, we construct hard negative samples by randomly swapping the coordinates of adjacent objects while keeping object categories and counts unchanged, thereby perturbing the spatial structure.
This design forces models to perform genuine spatial transformation and geometric reasoning from perspective views to top-down maps, effectively evaluating their ability to build internal spatial representations.

\textbf{Spatial-3D.}
Beyond spatial reasoning in natural images, we include a synthetic retrieval topic that targets abstract 3D geometric reasoning.
We build this subset on Spatial-DISE-12K~\cite{huang2025spatial}, a Blender-generated dataset with geometric tasks such as 3D rotation, folding, and composition.
To adapt it to retrieval, we convert the original multiple-choice questions into retrieval queries.
The correct option is treated as the positive target, while the original distractor options are directly reused as hard negatives, preserving their carefully designed geometric confusability.
As shown in Fig.~\ref{fig: case_Spatial_3D}, in a 3D projection task where the query requires finding the correct planar projection from a specified viewpoint, the model must distinguish it from distractors generated by alternative viewing directions, occlusion patterns, or depth-order permutations.
Because these negative samples may appear visually similar while differing in subtle geometric relationships (\textit{e.g.}, relative depth, overlap, and foreshortening), this task compels models to move beyond surface-level appearance matching and perform precise 3D-to-2D spatial transformations and viewpoint-aware geometric reasoning.

\subsection{Natural Science}
\textbf{Biology.}
Fine-grained biological identification is a long-standing challenge in multimodal retrieval, as many species exhibit highly similar visual appearances and differ only in subtle morphological traits.
This subtype targets knowledge-intensive retrieval: queries are natural-language descriptions or common names referring to specific species, and candidates are natural images of organisms, where correct retrieval primarily relies on recognizing subtle morphological cues and recalling taxonomic knowledge rather than multi-step reasoning.
We sample instances from the INQUIRE dataset~\cite{inquire}, which contains images of over 10{,}000 species annotated by citizen scientists and experts, providing rich visual diversity and expert-level taxonomic labels.
Hard negatives are constructed by first using LLMs to propose visually similar species as candidate distractors, and then manually selecting images that closely resemble the positive target in overall appearance or ecological traits but belong to different taxa.
As shown in Fig.~\ref{fig: case_Biology}, for the given query ``A resting Common Raven on the ground.'', common raven and corvus macrorhynchos exhibit nearly identical color distributions and body shapes, yet differ in subtle morphological details.
Such distractors ensure that successful retrieval depends on precise biological knowledge.

\textbf{Physics.}
This topic targets both {knowledge-intensive} and {reasoning-intensive} retrieval in physics, evaluating whether models can retrieve the correct visual evidence by combining domain knowledge with diagram-based understanding.
We curate undergraduate-level and advanced physics concepts and phenomenon illustrations spanning classical mechanics, electromagnetism, thermodynamics, and modern physics.
Queries are natural-language descriptions that often reference mechanisms, conditions, or observable outcomes, requiring models to interpret schematic diagrams or phenomenon figures and reason about the underlying physical process rather than relying on surface-level semantics.
Hard negatives are constructed whenever feasible by first using an LLM to propose closely related physical concepts that are easily confusable with the target, and then collecting corresponding visual illustrations from Wikimedia Commons~\cite{wikimediacommons}.
We restrict our collection to Commons assets with explicit reuse permissions and preserve attribution.
For the example in Fig.~\ref{fig: case_Physics}, when the ground-truth corresponds to the effective potential contours and Lagrange points, hard negatives include topologically similar geometries.
These candidates share similar nuclear contexts and visually similar diagrammatic patterns, yet differ in the underlying physical mechanism, ensuring that correct retrieval depends on physics-grounded reasoning and precise concept discrimination rather than superficial similarity.

\textbf{Chemistry.}
This topic targets {knowledge reasoning} and {symbolic reasoning} in chemistry, where correct retrieval requires understanding structured molecular representations and chemistry-specific transformation principles.
We source reaction simulation problems from EMMA~\cite{emma}, which are curated by domain experts who annotate reaction products and design questions with professional rigor, covering diverse reaction families and mechanistic patterns.
Queries describe a reaction setup using chemical structure diagrams and symbolic descriptions, and candidates contain corresponding molecular structures or reaction outcomes.
Hard negatives are constructed whenever feasible by manually selecting alternative reaction outcomes that are chemically plausible and structurally similar to the ground-truth products, but do not strictly satisfy the specific requirements implied by the query, as shown in Fig.~\ref{fig: case_chemistry}.
These distractors are designed to force models to rely on chemistry-specific reasoning over molecular structures instead of shortcut matching.

\subsection{Formal Science}

\textbf{Mathematics.}
Mathematics is a core topic in formal science. In ARK, this subtype is designed to evaluate knowledge reasoning and symbolic reasoning.
Following MR$^2$-Bench~\cite{mr2,math}, theorems and relations are expressed via geometric constructions and symbolic diagrams.
As shown in Fig.~\ref{fig: case_Mathematics}, answering these queries requires models to combine domain knowledge (\textit{e.g.}, mathematical definitions, theorems, and constraints) with symbolic manipulation, and to ground the reasoning on visual evidence such as geometric figures.
To increase retrieval difficulty, we construct hard negatives by manually selecting diagrams or symbolic expressions that are highly similar to the query in appearance or surface form, but conflict with key mathematical constraints.

\textbf{Code-Drawing.}
This subtype targets {symbolic reasoning} over code-conditioned visual generation, requiring models to ground structured program symbols into the corresponding visual structures.
We use the TikZ code--image pairs from ReMI~\cite{remi}, where queries are raw TikZ snippets and gallery items are the rendered vector graphics produced by compiling the code as shown in Fig.~\ref{fig: case_code_drawing}.
Correct retrieval therefore depends on understanding how syntactic primitives and parameters (\textit{e.g.}, coordinates, commands, and constraints) deterministically specify geometric layouts.
Hard negatives are constructed via a single-line perturbation strategy. Specifically, we minimally edit a critical line in the original TikZ code and compile the modified program to obtain a distractor image that remains visually similar but changes key geometric relations (\textit{e.g.}, relative positions, connectivity, or angles).
Such negatives enforce precise code-to-image grounding and discourage shortcut matching based on global appearance.

\subsection{Humanities and Social Science}

\textbf{Economics.}
This subtype targets knowledge reasoning and conceptual abstraction over economic data visualizations, evaluating whether models can retrieve the correct chart by understanding both economic concepts and the abstract information conveyed by graphical trends.
This subset follows the data sourcing of MR$^2$-Bench~\cite{mr2}: we use the curated economics reports and charts derived from World Bank Open Data~\cite{worldbank}, covering economic phenomena across countries and time periods.
Queries are natural-language descriptions that reference economic concepts, comparative relations, and temporal or predictive conditions, requiring models to interpret chart semantics rather than match surface text or visual patterns.
Hard negatives are constructed by selecting charts that share similar economic keywords or themes, but convey different economic mechanisms or implications than those required by the query.
For example, as shown in Fig.~\ref{fig: case_Economics}, for the query ``Assuming a host has a duty to admit people escaping danger, but discretion over opportunity-driven entrants, which diagram best supports concluding within a generic two-by-two that crosses push vs.
pull motives with the strength of the skills-demand alignment that the near-term net payoff is governed chiefly by the match, with an explicit side-by-side contrast of net gains versus net losses along the match gradient?'', hard negatives may include diagrams that emphasize push–pull forces without an explicit skill–demand match dimension or charts that depict aggregate trends rather than a match-based contrast of gains versus losses.

\textbf{Comic.}
This subtype targets {knowledge reasoning} and {conceptual abstraction} on historical cartoons, where meaning is conveyed implicitly through symbolism and metaphor rather than explicit visual descriptions.
Accurate retrieval requires recognizing historically grounded cues and mapping concrete visual elements (\textit{e.g.}, characters, icons, and allegories) to higher-level concepts and viewpoints, as illustrated in Fig.~\ref{fig: case_Comic}.
We construct this topic based on the Bulletin Editorial Cartoons dataset~\cite{sherratt_trove_journals_2024}, which covers cartoons published on the front pages of The Bulletin\footnote{\href{https://en.wikipedia.org/wiki/The_Bulletin_(Australian_periodical)}{The Bulletin (Australian periodical)}} between 1886 and 1952.
We select 3,468 cartoons to form the gallery and manually annotate 87 instances to create the query set.
Given the difficulty of constructing hard negative samples for this topic, we instead avoid shortcut learning by constraining the query format. Each query consists of only two components: a statement of the historical event and a revelation of the underlying viewpoint. This design deliberately avoids direct visual descriptions of the images, requiring models to genuinely understand the metaphors embedded in the cartoons rather than relying on superficial visual feature matching.

\textbf{Art.}
This subtype targets {knowledge reasoning} in art, where queries provide indirect art-historical cues rather than explicit entity names, and correct retrieval requires inferring the most likely artist, style, or subject matter from visual and textual evidence.
We use SemArt~\cite{semart}, which pairs paintings with expert-written descriptions and structured metadata (\textit{e.g.}, artist and genre), enabling queries grounded in art history.
As shown in Fig.~\ref{fig: case_Art}, queries describe stylistic signatures, creative techniques, iconography, or thematic conventions. Successful retrieval therefore requires reasoning with art knowledge to connect these cues to the appropriate artwork, rather than relying on direct name matching or generic aesthetic similarity.
Hard negatives are constructed by first retrieving style-neighbor candidates in the embedding space and then manually selecting artworks that are visually close in overall style but violate key knowledge constraints implied by the query (\textit{e.g.}, period, technique, or iconographic elements).
These distractors force models to perform knowledge-grounded inference over artworks rather than exploiting superficial stylistic resemblance.

\textbf{Calendar.}
This subtype targets {knowledge reasoning} and {logical reasoning} over artifacts from the traditional Chinese calendar system.
It evaluates cross-modal understanding and reasoning over traditional Chinese calendar artifacts, including the lunar calendar system and the 24 solar terms.
We curate this subset by manually selecting representative calendar images from publicly available online resources with reuse permission, focusing on solar-term annotations, lunar--Gregorian correspondences, and culturally grounded visual elements.
Queries require the model to analyze the text and visual information in calendar images (\textit{e.g.}, solar term names, lunar/Gregorian date positions, zodiac animal information, etc.) and incorporate reasoning based on knowledge, such as lunar calendar rules and the sequence of solar terms. As shown in Fig.~\ref{fig: case_Calendar}, a query like ``A calendar month page from which one can determine that people born in the lunar second month can celebrate their lunar birthday twice.'' requires the model to understand that having two birthdays is possible only when there is a leap month, specifically a leap second lunar month, and thus infer that the calendar image in question corresponds to a year with a leap second lunar month.

\textbf{Metro.}
This subtype targets {spatial reasoning} and fine-grained visual reasoning in metro diagrams. Unlike knowledge-intensive retrieval, the goal is to test whether models can find the correct map or station context by reasoning about connectivity and topology, rather than relying on domain knowledge or surface-level keyword matching.
We manually curate high-resolution subway route maps from publicly available online resources with reuse permission. The maps have an average resolution of $9858\times9057$, substantially increasing the demand for precise visual inspection and spatial reasoning.
As shown in Fig.~\ref{fig: case_Metro}, queries are natural-language descriptions that refer to route connectivity, transfer relations, and relative station positions, and candidates are the corresponding metro maps.

\subsection{Engineering and Technology}

\textbf{Computer-Science.}
This subtype targets {knowledge reasoning} and {conceptual abstraction} over figures from computer science literature, testing whether retrievers can interpret technical visuals and recover the intended high-level takeaway instead of relying on superficial keyword overlap.
We curate this subset from publicly available arXiv papers and complementary online resources with reuse permission, covering common CS figure types such as algorithm schematics, data-structure diagrams, and experimental plots (\textit{e.g.}, scaling trends and complexity-related curves).
As shown in Fig.~\ref{fig: case_Computer_Science},
queries are natural-language statements describing methodological claims or figure-implied conclusions, and candidates are the corresponding paper figures.
Correct retrieval requires grounding the query in structured visual evidence (\textit{e.g.}, axes, legends, symbols, and trend patterns) and using CS knowledge to infer the conclusion supported by the figure.

\textbf{Mechanical.}
This subtype is primarily {knowledge-intensive}, with a small portion requiring {knowledge reasoning} to disambiguate visually similar mechanisms. It evaluates cross-modal grounding between mechanical engineering terminology and technical imagery as shown in Fig.~\ref{fig: case_Mechanical}, where models must retrieve the correct part or mechanism given its name or functional description.
We construct this subset from a mechanical mechanism dataset~\cite{mechanisms} containing $\sim$8K images with structured descriptions, covering common components and mechanisms (\textit{e.g.}, gears, bearings, linkages, and fasteners).
Hard negatives are obtained by retrieving the most similar candidates in the embedding space and then manually filtering to retain distractors that are category-consistent yet differ in key geometric or functional attributes.

\textbf{Electronics.}
This subtype targets {logical reasoning} and {symbolic reasoning} in digital circuit understanding, evaluating retrieval over structured artifacts such as circuit diagrams, truth tables, and Karnaugh maps.
We build this subset on ElectroVizQA~\cite{ElectroVizQA}, which covers both combinational and sequential logic with paired visuals and Boolean-form descriptions.
As shown in Fig.~\ref{fig: case_Electronics}, queries require models to interpret gate symbols and circuit structure, or to infer Boolean relations from truth tables, and then retrieve the matching visual evidence.
Hard negatives are constructed by preserving highly similar visual layouts while perturbing the underlying logic (\textit{e.g.}, swapping gate types, modifying truth-table entries, or changing Boolean operators), yielding distractors that look plausible but violate the intended logical function.

\clearpage
\section{Details of Evaluated Models}
\label{appendix:evaluated_models}

{
We summarize the evaluated embedding-based retrievers, re-ranking models, and agentic retrievers in Tables~\ref{tab:embedding_models}, \ref{tab:rerank_models}, and \ref{tab:agentic_models}, respectively, covering 25 representative retriever configurations in total.
These configurations span text-only and multimodal retrieval paradigms, a wide range of parameter scales, and different training strategies, including models with and without explicit reasoning-oriented objectives.
This breadth enables a comprehensive assessment of how modality, architecture, model capacity, and retrieval paradigm affect knowledge-intensive and reasoning-intensive multimodal retrieval.
}

All locally executed inference experiments were conducted on NVIDIA RTX 4090 GPUs. For each evaluated model, we use the officially released checkpoint and default inference configuration.
For API-based models, computation is performed through the corresponding hosted service provider.

\begin{table}[!htb]
\caption{{Overview of Evaluated Embedding Models.}
}
\newcommand{\cmark}{\ding{51}}
\newcommand{\xmark}{\ding{55}}
\label{tab:embedding_models}
\centering
\resizebox{\textwidth}{!}{
\tablestyle{2pt}{1.1}{
\begin{tabular}{c l c c l}
\toprule
\textbf{\textsc{Modality}} & \textbf{\textsc{Model Name}} & \textbf{\textsc{Param.}} & \textbf{\textsc{Reasoning}} &\textbf{Model Link}\\
\midrule

\multirow{6}{*}{\textsc{Text}}
 & BGE-M3~\cite{bgem3} & 600M & & \href{https://huggingface.co/BAAI/bge-m3}{HF: BAAI/bge-m3}\\
  & Qwen3-Embedding~\cite{qwen3embedding} & 4B & & \href{https://huggingface.co/Qwen/Qwen3-Embedding-4B}{HF: Qwen/Qwen3-Embedding-4B}\\
 & Qwen3-Embedding~\cite{qwen3embedding} & 8B & & \href{https://huggingface.co/Qwen/Qwen3-Embedding-8B}{HF: Qwen/Qwen3-Embedding-8B}\\
  & Diver-Embed~\cite{diver} & 1.7B & \cmark & \href{https://huggingface.co/AQ-MedAI/Diver-Retriever-1.7B}{HF: AQ-MedAI/Diver-Retriever-1.7B}\\
 & Diver-Embed~\cite{diver} & 4B & \cmark & \href{https://huggingface.co/AQ-MedAI/Diver-Retriever-4B}{HF: AQ-MedAI/Diver-Retriever-4B}\\
 & ReasonIR~\cite{reasonir} & 8B & \cmark & \href{https://huggingface.co/reasonir/ReasonIR-8B}{HF: reasonir/ReasonIR-8B}\\
\midrule

\multirow{13}{*}{\begin{tabular}{c}\textsc{Multi}\\\textsc{modal}\end{tabular}}
 & CLIP~\cite{clip} & 428M & & \href{https://huggingface.co/openai/clip-vit-large-patch14-336}{HF: openai/clip-vit-large-patch14-336}\\
 & BGE-VL~\cite{zhou2025megapairs} & 400M & & \href{https://huggingface.co/BAAI/BGE-VL-large}{HF: BAAI/BGE-VL-large}\\
 & SigLIP2~\cite{tschannen2025siglip} & 0.9B & & \href{https://huggingface.co/google/siglip2-so400m-patch14-384}{HF: google/siglip2-so400m-patch14-384}\\
 & MetaCLIP2~\cite{chuang2025meta} & 2B & & \href{https://huggingface.co/facebook/metaclip-2-worldwide-huge-quickgelu}{HF: facebook/metaclip-2-worldwide-huge-quickgelu}\\
 & EVA-CLIP~\cite{EVA-CLIP-18B} & 18B & & \href{https://huggingface.co/BAAI/EVA-CLIP-18B}{HF: BAAI/EVA-CLIP-18B}\\
 & VLM2VecV2~\cite{meng2025vlm2vec} & 2B & & \href{https://huggingface.co/VLM2Vec/VLM2Vec-V2.0}{HF: VLM2Vec/VLM2Vec-V2.0}\\
 & LamRA-Ret~\cite{lamra} & 7B & & \href{https://huggingface.co/code-kunkun/LamRA-Ret}{HF: code-kunkun/LamRA-Ret}\\
 & RzenEmbed~\cite{jian2025rzenembed} & 7B & & \href{https://huggingface.co/qihoo360/RzenEmbed}{HF: qihoo360/RzenEmbed}\\
  & Ops-MM-Embedding~\cite{opensearch_ops_mm_embedding_2025} & 2B & & \href{https://huggingface.co/OpenSearch-AI/Ops-MM-Embedding-v1-2B}{HF: OpenSearch-AI/Ops-MM-Embedding-v1-2B}\\
 & Ops-MM-Embedding~\cite{opensearch_ops_mm_embedding_2025} & 7B & & \href{https://huggingface.co/OpenSearch-AI/Ops-MM-Embedding-v1-7B}{HF: OpenSearch-AI/Ops-MM-Embedding-v1-7B}\\
  & Qwen3-VL-embedding~\cite{li2026qwen3} & 2B & & \href{https://huggingface.co/Qwen/Qwen3-VL-Embedding-2B}{HF: Qwen/Qwen3-VL-Embedding-2B}\\
 & Qwen3-VL-embedding~\cite{li2026qwen3} & 8B & & \href{https://huggingface.co/Qwen/Qwen3-VL-Embedding-8B}{HF: Qwen/Qwen3-VL-Embedding-8B}\\
 & Seed1.6-embedding & - & & \href{https://console.volcengine.com/ark/region:ark+cn-beijing/model/detail?Id=doubao-embedding-vision}{API: Doubao-Embedding-Vision} \\
\bottomrule
\end{tabular}
}
}
\end{table}

\begin{table}[!htb]
\caption{{Overview of Evaluated Reranker Models.} }
\newcommand{\cmark}{\ding{51}}
\newcommand{\xmark}{\ding{55}}
\label{tab:rerank_models}
\centering
\tablestyle{12pt}{1.1}{
\begin{tabular}{c l c l}
\toprule
\textbf{\textsc{Modality}} & \textbf{\textsc{Model Name}} & \textbf{\textsc{Param.}}
& \textbf{Model Link}\\
\midrule

{\textsc{Text}} & BGE-Reranker~\cite{bgem3} & 600M & \href{https://huggingface.co/BAAI/bge-reranker-v2-m3}{HF: BAAI/bge-reranker-v2-m3}\\

\midrule
\multirow{2}{*}{\textsc{Multi-modal}}
 & Qwen3-VL-Reranker~\cite{li2026qwen3} & 8B & \href{https://huggingface.co/Qwen/Qwen3-VL-Reranker-8B}{HF: Qwen/Qwen3-VL-Reranker-8B}\\
 & Jina-Reranker~\cite{jina_reranker_m0_2025} & 2B & \href{https://huggingface.co/jinaai/jina-reranker-m0}{HF: jinaai/jina-reranker-m0}\\

\bottomrule
\end{tabular}
}
\end{table}

\begin{table}[!htb]
\caption{Overview of Evaluated Agentic Models. A-RAG combines GPT with Qwen3 embedding models.
Since GPT is shared across settings, we distinguish the two A-RAG variants by the parameter size of the underlying Qwen3 retriever.
}
\newcommand{\cmark}{\ding{51}}
\newcommand{\xmark}{\ding{55}}
\label{tab:agentic_models}
\centering
\tablestyle{14pt}{1.1}{
\begin{tabular}{c l c l}
\toprule
\textbf{\textsc{Modality}} & \textbf{\textsc{Model Name}} & \textbf{\textsc{Param.}}
& \textbf{Model Link}\\
\midrule

\multirow{3}{*}{\textsc{Text}}
 & {A-RAG~\cite{a-rag}} & 0.6B &

 \href{https://github.com/Ayanami0730/arag}{GPT-5-mini + Qwen3-Embedding-0.6B}\\
 & {A-RAG~\cite{a-rag}} & 8B &  \href{https://github.com/Ayanami0730/arag}{GPT-5-mini + Qwen3-Embedding-8B}\\
 & Agentic-R~\cite{Agentic-R} & 7B & \href{https://huggingface.co/liuwenhan/triviaqa_hotpotqa_train-search-r1-ppo-qwen2.5-7b-em-iter1}{HF: liuwenhan/search-agent}\\

\bottomrule
\end{tabular}
}
\end{table}

\clearpage
\section{Evaluation with Query Rewriting and Re-ranking}
\label{appendix:rewrite_rerank}

Due to space constraints in the main paper, we report additional results for \emph{query rewriting} and \emph{re-ranking} in Tables~\ref{table:Reranker and Query Rewrite} and~\ref{table:Reranker and Query Rewrite for reasoning}.

\begin{table*}[h]
\caption{{
Performance of retrieval models with query rewriting and re-ranking on ARK.
R@k and N@k denote Recall@k and nDCG@k, respectively.
Scores are macro-averaged across subtypes under the knowledge-axis grouping and across reasoning skills under the reasoning-axis grouping.
}}
\centering
\renewcommand{\arraystretch}{1}
\resizebox{\textwidth}{!}
{
\tablestyle{6pt}{1.2}
{
\begin{tabular}{l|cccccc|cccccc}
\toprule
\multirow{2}{*}{\textbf{Model}} & \multicolumn{6}{c|}{\textbf{Knowledge perspective}}
& \multicolumn{6}{c}{\textbf{Reasoning perspective}} \\
 & R@1 & R@5 & R@10 & N@5 & N@10 & N@20 & R@1 & R@5 & R@10 & N@5 & N@10 & N@20  \\
\midrule
\multicolumn{13}{c}{\textit{\textbf{Query Rewrite}}} \\
\midrule
Seed1.6-embedding & 19.32 & 37.60 & 48.43 & 28.79& 32.28&34.77&15.30&31.69&40.70&23.80&26.71&29.18 \\
\rowcolor{lightgray} \ + \textit{Rewrite} & 23.42 & 43.85 & 54.49 & 34.04 & 37.52 & 39.87 & 17.49 & 35.09 & 44.12 & 26.59 & 29.53 & 31.96 \\
\midrule
\multicolumn{13}{c}{\textit{\textbf{Reranker Models Based on Seed1.6-embedding}}} \\
\midrule
BGE-Reranker & 12.23 & 23.95 & 32.47 & 18.29 & 21.05 & 24.11 & 8.19 & 18.35 & 26.01 & 13.40 & 15.88 & 18.51 \\
\rowcolor{lightgray} \ + \textit{Rewrite} & 15.25 & 27.29 & 35.97 & 21.50 & 24.29 & 27.49 & 11.64 & 21.39 & 28.98 & 16.72 & 19.15 & 21.70 \\
Jina-Reranker & 23.90 & 43.06 & 52.88 & 33.95 & 37.16 & 39.46 & 17.14 & 33.48 & 42.91 & 25.74 & 28.81 & 30.91 \\
\rowcolor{lightgray} \ + \textit{Rewrite} & 27.77 & 47.47 & 57.69 & 38.17 & 41.50 & 43.32 & 21.21 & 38.27 & 47.32 & 30.20 & 33.14 & 34.76 \\
Qwen3-VL-Reranker & \underline{29.33} & \underline{49.20} & \underline{58.25} & \underline{39.92} & \underline{42.86} & \underline{44.46} & \underline{23.89} & \underline{41.76} & \underline{49.56} & \underline{33.39} & \underline{35.90} & \underline{37.28} \\
\rowcolor{lightgray} \ + \textit{Rewrite} & \textbf{34.51} & \textbf{51.45} & \textbf{59.38} & \textbf{43.66} & \textbf{46.23} & \textbf{47.88} & \textbf{28.31} & \textbf{42.76} & \textbf{49.58} & \textbf{36.12} & \textbf{38.29} & \textbf{39.86} \\
\bottomrule
\end{tabular}
}
}
\label{table:Reranker and Query Rewrite}
\end{table*}

\begin{table*}[h]
\caption{
Recall@1 on ARK by reasoning skill with query rewriting and re-ranking.
}
\centering
\renewcommand{\arraystretch}{1}
\resizebox{\textwidth}{!}
{
\tablestyle{6pt}{1.2}
{
\begin{tabular}{l|ccccccc}
\toprule
\textbf{Model}
 & \makecell{\textbf{Knowledge}\\\textbf{Reasoning}}
& \makecell{\textbf{Conceptual}\\\textbf{Abstraction}}
& \makecell{\textbf{Fine-Grained}\\\textbf{Visual Reasoning}}
& \makecell{\textbf{Logical}\\\textbf{Reasoning}}
& \makecell{\textbf{Spatial}\\\textbf{Reasoning}}
& \makecell{\textbf{Symbolic}\\\textbf{Reasoning}}
& \textbf{Average}

 \\
\midrule
\multicolumn{8}{c}{\textit{\textbf{Query Rewrite}}} \\
\midrule
Seed1.6-embedding & 22.08&33.33&4.35&10.71&6.19&15.16&15.30 \\
\rowcolor{lightgray} \ + \textit{Rewrite} & 24.96 & 31.00 & 10.00 & 18.40 & 5.00 & 15.58 & 17.49 \\
\midrule
\multicolumn{8}{c}{\textit{\textbf{Reranker Models Based on Seed1.6-embedding}}} \\
\midrule
BGE-Reranker & 14.55 & 21.40 & 3.85 & 3.07 & 1.54 & 4.76 & 8.19 \\
\rowcolor{lightgray} \ + \textit{Rewrite} & 16.98 & 21.40 & 5.38 & 17.18 & 3.08 & 5.84 & 11.64 \\
Jina-Reranker & 27.82 & \textbf{41.48} & 10.00 & 4.91 & 6.92 & 11.69 & 17.14 \\
\rowcolor{lightgray} \ + \textit{Rewrite} & 29.24 & {38.86} & {14.62} & \underline{21.47} & 7.69 & 15.37 & 21.21 \\
Qwen3-VL-Reranker & \underline{33.95} & 37.12 & \textbf{30.00} & 5.52 & \textbf{10.77} & \underline{25.97} & \underline{23.89} \\
\rowcolor{lightgray} \ + \textit{Rewrite} & \textbf{38.09} & \underline{39.74} & \underline{26.15} & \textbf{23.31} & \underline{9.23} & \textbf{33.33} & \textbf{28.31} \\
\bottomrule
\end{tabular}
}
}
\label{table:Reranker and Query Rewrite for reasoning}
\end{table*}

\clearpage
\section{Results on ARK under different evaluation metrics.}
\label{appendix:results_metrics}

In addition to the main Recall@1 results reported in Tables~\ref{table:knowledge_score} and \ref{table:reasoning_score}, we provide a broader view of retrieval quality under multiple metrics.
Tables~\ref{table:knowledge_score_recall@5}--\ref{table:reasoning_score_ndcg@20} report Recall@5/10 and nDCG@5/10/20. Recall@$k$ measures whether the ground-truth item appears within the top-$k$ retrieved candidates, while nDCG@$k$ further accounts for its rank position, assigning higher credit when the correct item is retrieved closer to the top.

\begin{table*}[h]
\caption{Domain-wise Recall@5 performance on ARK (knowledge axis). Columns correspond to ARK's five knowledge domains and their subtypes. ``Avg.'' denotes the macro-average across all subtypes.
}
\centering
\renewcommand{\arraystretch}{1.2}
\resizebox{\textwidth}{!}
{
{
\begin{tabular}{l|cccc|ccc|cc|ccccc|ccc|c}
\toprule
\multirow{2}{*}{\textbf{Model}} & \multicolumn{4}{c|}{\textbf{Visual Cognition}}
& \multicolumn{3}{c|}{\textbf{Natural Science}}
& \multicolumn{2}{c|}{\textbf{Formal Science}}
& \multicolumn{5}{c|}{\textbf{Humanities \& Social Science}}
& \multicolumn{3}{c|}{\textbf{Engineering \& Technology}}
& \multirow{2}{*}{\textbf{Avg.}} \\
 & Daily. & Fine. & SCog. & S3D. & Bio. & Phy. & Che. & Mat. & Cod. & Eco. & Comic. & Art & Cal. & Met. & Compu. & Mecha. & Ele. &  \\
\midrule
\multicolumn{19}{c}{\textit{\textbf{Text Embedding Models}}} \\
\midrule
BGE-M3-0.6B & 57.63 & 6.52 & 0.00 & 1.53 & 63.86 & 30.00 & 8.57 & 10.53 & 4.00 & 56.25 & 3.45 & 9.00 & 7.50 & 31.25 & 31.25 & 10.98 & 0.99 & 19.61 \\
Diver-Embed-1.7B & 48.73 & 7.61 & 0.00 & 0.76 & 57.83 & 46.25 & 9.52 & 26.97 & 7.00 & 50.00 & 2.30 & 7.00 & 10.00 & 30.00 & 52.50 & 14.63 & 1.98 & 21.95 \\
Diver-Embed-4B & 61.02 & 6.52 & 0.00 & 0.76 & 68.67 & 40.00 & 15.24 & 31.58 & 15.00 & \underline{75.00} & 5.75 & 11.00 & 7.50 & 38.75 & 60.00 & 15.85 & 4.95 & 26.92 \\
Qwen3-Embed-4B & 57.20 & 7.61 & 4.94 & 2.29 & 71.08 & 41.25 & 14.29 & 28.95 & 13.00 & 63.75 & 3.45 & 9.00 & 7.50 & 23.75 & 51.25 & 20.73 & 6.93 & 25.12 \\
\midrule
Qwen3-Embed-8B & 61.86 & \underline{10.87} & 3.70 & 2.29 & 73.49 & \underline{50.00} & 17.14 & 30.92 & 14.00 & \textbf{77.50} & 6.90 & 7.00 & \underline{16.25} & 27.50 & 58.75 & 21.95 & 9.90 & 28.82 \\
ReasonIR-8B & 59.75 & 8.70 & 0.00 & 1.53 & 62.65 & 38.75 & 15.24 & 21.71 & 6.00 & 73.75 & 10.34 & 5.00 & 7.50 & 41.25 & 58.75 & 17.07 & 3.96 & 25.41 \\
\midrule
\multicolumn{19}{c}{\textit{\textbf{Multimodal Embedding Models}}} \\
\midrule
CLIP-0.4B & 61.44 & 8.70 & 3.70 & 10.69 & 61.45 & 20.00 & 11.43 & 5.92 & 2.00 & 27.50 & 9.20 & 29.00 & 10.00 & 23.75 & 16.25 & 17.07 & \underline{11.88} & 19.41 \\
BGE-VL-0.4B & 72.46 & 9.78 & 0.00 & 0.00 & 67.47 & 8.75 & 0.00 & 7.24 & 8.00 & 10.00 & 1.15 & 13.00 & 3.75 & 13.75 & 16.25 & 12.20 & 9.90 & 14.92 \\
SigLIP2-0.9B & \textbf{88.98} & 6.52 & 2.47 & 4.58 & 66.27 & 15.00 & 15.24 & 11.84 & 1.00 & 31.25 & 9.20 & 15.00 & 11.25 & 5.00 & 35.00 & \underline{26.83} & 5.94 & 20.67 \\
\midrule
MetaCLIP2-2B & 78.39 & \textbf{13.04} & 4.94 & 6.87 & \underline{78.31} & 38.75 & 8.57 & 9.21 & 8.00 & 42.50 & 4.60 & \textbf{41.00} & 7.50 & 12.50 & 28.75 & 25.61 & 10.89 & 24.67 \\
VLM2VecV2-2B & 65.25 & 8.70 & 13.58 & 6.11 & 63.86 & 27.50 & 23.81 & 28.95 & 10.00 & 58.75 & 3.45 & 10.00 & 7.50 & 30.00 & 37.50 & 20.73 & 5.94 & 24.80 \\
Ops-MM-Embed-v1-2B & 83.05 & 8.70 & 6.17 & 7.63 & 74.70 & 32.50 & \textbf{40.95} & 25.66 & 12.00 & 48.75 & 13.79 & 18.00 & 6.25 & 31.25 & 50.00 & 18.29 & 8.91 & 28.62 \\
Qwen3-VL-Embed-2B & 85.17 & \underline{10.87} & 14.81 & \underline{12.21} & 73.49 & 36.25 & 26.67 & 32.89 & 9.00 & 70.00 & 13.79 & 9.00 & 3.75 & \underline{45.00} & 47.50 & 20.73 & \textbf{16.83} & 31.06 \\
\midrule
LamRA-Ret-7B & 78.39 & 8.70 & 1.23 & 3.82 & \underline{78.31} & 13.75 & 5.71 & 25.00 & 24.00 & 62.50 & 9.20 & 17.00 & 3.75 & 30.00 & 6.25 & 14.63 & 3.96 & 22.72 \\
RzenEmbed-7B & 84.75 & \underline{10.87} & \underline{20.99} & 4.58 & 71.08 & 45.00 & 35.24 & 33.55 & 30.00 & 67.50 & 14.94 & 19.00 & 12.50 & 43.75 & \textbf{72.50} & 23.17 & \textbf{16.83} & 35.66 \\
Ops-MM-Embed-v1-7B & 83.05 & \textbf{13.04} & \textbf{25.93} & \textbf{13.74} & 75.90 & 43.75 & \textbf{40.95} & \underline{35.53} & 20.00 & 68.75 & 17.24 & 23.00 & 10.00 & 43.75 & 60.00 & 21.95 & \textbf{16.83} & 36.08 \\
Qwen3-VL-Embed-8B & \underline{86.86} & 9.78 & 6.17 & \textbf{13.74} & \underline{78.31} & \textbf{51.25} & \underline{39.05} & 30.92 & \textbf{45.00} & \textbf{77.50} & \underline{19.54} & 12.00 & 11.25 & 43.75 & 56.25 & \underline{26.83} & \textbf{16.83} & \underline{36.77} \\
\midrule
EVA-CLIP-18B & 83.05 & \underline{10.87} & 0.00 & \underline{12.21} & \textbf{86.75} & 35.00 & 10.48 & 1.32 & 9.00 & 52.50 & 6.90 & \underline{38.00} & 6.25 & 15.00 & 23.75 & \textbf{29.27} & 5.94 & 25.08 \\
Seed1.6-embedding & 84.75 & \textbf{13.04} & 7.41 & 8.40 & 66.27 & 45.00 & \textbf{40.95} & \textbf{38.16} & \underline{36.00} & \underline{75.00} & \textbf{21.84} & 26.00 & \textbf{27.50} & \textbf{47.50} & \underline{71.25} & 23.17 & 6.93 & \textbf{37.60} \\

\bottomrule
\end{tabular}
}
}
\label{table:knowledge_score_recall@5}
\end{table*}

\begin{table*}[h]
\caption{Domain-wise Recall@10 performance on ARK (knowledge axis). Columns correspond to ARK's five knowledge domains and their subtypes.}
\centering
\renewcommand{\arraystretch}{1.2}
\resizebox{\textwidth}{!}
{
{
\begin{tabular}{l|cccc|ccc|cc|ccccc|ccc|c}
\toprule
\multirow{2}{*}{\textbf{Model}} & \multicolumn{4}{c|}{\textbf{Visual Cognition}}
& \multicolumn{3}{c|}{\textbf{Natural Science}}
& \multicolumn{2}{c|}{\textbf{Formal Science}}
& \multicolumn{5}{c|}{\textbf{Humanities \& Social Science}}
& \multicolumn{3}{c|}{\textbf{Engineering \& Technology}}
& \multirow{2}{*}{\textbf{Avg.}} \\
 & Daily. & Fine. & SCog. & S3D. & Bio. & Phy. & Che. & Mat. & Cod. & Eco. & Comic. & Art & Cal. & Met. & Compu. & Mecha. & Ele. &  \\
\midrule
\multicolumn{19}{c}{\textit{\textbf{Text Embedding Models}}} \\
\midrule
BGE-M3-0.6B & 63.98 & 11.96 & 1.23 & 1.53 & 68.67 & 37.50 & 17.14 & 15.13 & 11.00 & 67.50 & 6.90 & 13.00 & 7.50 & 45.00 & 43.75 & 14.63 & 4.95 & 25.37 \\
Diver-Embed-1.7B & 56.78 & 9.78 & 0.00 & 1.53 & 72.29 & 50.00 & 19.05 & 31.58 & 14.00 & 57.50 & 2.30 & 12.00 & 12.50 & 50.00 & 58.75 & 19.51 & 7.92 & 27.97 \\
Diver-Embed-4B & 68.64 & 13.04 & 0.00 & 0.76 & 78.31 & 48.75 & 23.81 & 38.82 & 23.00 & \underline{82.50} & 6.90 & 17.00 & 10.00 & 53.75 & 67.50 & 19.51 & 6.93 & 32.90 \\
Qwen3-Embed-4B & 65.68 & 15.22 & 16.05 & 3.82 & 79.52 & 48.75 & 19.05 & 37.50 & 19.00 & 76.25 & 5.75 & 13.00 & 8.75 & 50.00 & 58.75 & 23.17 & 14.85 & 32.65 \\
\midrule
Qwen3-Embed-8B & 70.34 & 15.22 & 3.70 & 6.11 & 83.13 & 53.75 & 27.62 & 41.45 & 21.00 & \textbf{85.00} & 10.34 & 14.00 & 21.25 & 50.00 & 65.00 & 25.61 & 13.86 & 35.73 \\
ReasonIR-8B & 69.49 & 17.39 & 3.70 & 3.05 & 73.49 & 45.00 & 21.90 & 32.24 & 9.00 & \underline{82.50} & 13.79 & 7.00 & 10.00 & 61.25 & 66.25 & 23.17 & 6.93 & 32.13 \\
\midrule
\multicolumn{19}{c}{\textit{\textbf{Multimodal Embedding Models}}} \\
\midrule
CLIP-0.4B & 69.49 & 18.48 & 7.41 & 12.98 & 85.54 & 30.00 & 14.29 & 10.53 & 6.00 & 42.50 & 12.64 & 40.00 & 13.75 & 35.00 & 21.25 & 20.73 & 14.85 & 26.79 \\
BGE-VL-0.4B & 81.78 & 14.13 & 0.00 & 0.00 & 85.54 & 12.50 & 0.00 & 13.82 & 10.00 & 22.50 & 1.15 & 18.00 & 6.25 & 33.75 & 18.75 & 15.85 & 10.89 & 20.29 \\
SigLIP2-0.9B & \textbf{92.80} & 7.61 & 7.41 & 5.34 & 75.90 & 23.75 & 20.95 & 17.11 & 1.00 & 37.50 & 10.34 & 17.00 & \underline{23.75} & 16.25 & 42.50 & 31.71 & 7.92 & 25.81 \\
\midrule
MetaCLIP2-2B & 86.02 & 18.48 & 7.41 & 10.69 & \textbf{96.39} & 51.25 & 11.43 & 13.82 & 13.00 & 50.00 & 5.75 & \textbf{55.00} & 8.75 & 31.25 & 32.50 & \underline{35.37} & 12.87 & 31.76 \\
VLM2VecV2-2B & 74.15 & 15.22 & 35.80 & 9.92 & 74.70 & 33.75 & 36.19 & 42.76 & 19.00 & 65.00 & 5.75 & 14.00 & 10.00 & 48.75 & 41.25 & 23.17 & 10.89 & 32.96 \\
Ops-MM-Embed-v1-2B & 90.68 & 16.30 & 23.46 & 12.21 & 90.36 & 45.00 & 55.24 & 36.84 & 20.00 & 58.75 & 16.09 & 25.00 & 6.25 & 56.25 & 57.50 & 23.17 & 17.82 & 38.29 \\
Qwen3-VL-Embed-2B & 89.83 & 16.30 & 30.86 & 14.50 & 83.13 & 48.75 & 42.86 & 40.79 & 17.00 & 76.25 & 19.54 & 17.00 & 5.00 & \underline{63.75} & 50.00 & 24.39 & \textbf{24.75} & 39.10 \\
\midrule
LamRA-Ret-7B & 88.14 & 14.13 & 12.35 & 7.63 & 91.57 & 27.50 & 8.57 & 31.58 & 37.00 & 75.00 & 13.79 & 27.00 & 3.75 & 45.00 & 7.50 & 15.85 & 3.96 & 30.02 \\
RzenEmbed-7B & \textbf{92.80} & 15.22 & \underline{44.44} & 6.11 & 87.95 & 52.50 & \textbf{63.81} & 42.76 & 42.00 & \underline{82.50} & 17.24 & 23.00 & 18.75 & 56.25 & \underline{76.25} & 29.27 & \underline{19.80} & 45.33 \\
Ops-MM-Embed-v1-7B & 89.83 & \textbf{22.83} & \textbf{56.79} & \underline{19.08} & 92.77 & 51.25 & \textbf{63.81} & \underline{45.39} & 30.00 & 75.00 & 19.54 & 32.00 & 13.75 & \textbf{67.50} & 62.50 & 28.05 & 16.83 & \underline{46.29} \\
Qwen3-VL-Embed-8B & \underline{91.95} & 14.13 & 34.57 & \textbf{19.85} & 92.77 & \textbf{60.00} & 58.10 & 44.74 & \underline{55.00} & \underline{82.50} & \underline{21.84} & 20.00 & 15.00 & 58.75 & 61.25 & 31.71 & 16.83 & 45.82 \\
\midrule
EVA-CLIP-18B & 88.56 & \underline{21.74} & 2.47 & 16.03 & \underline{93.98} & 46.25 & 14.29 & 5.26 & 11.00 & 61.25 & 8.05 & \underline{50.00} & 6.25 & 26.25 & 33.75 & \textbf{37.80} & 11.88 & 31.46 \\
Seed1.6-embedding & 91.10 & 18.48 & 30.86 & 12.98 & 83.13 & \underline{56.25} & \underline{60.95} & \textbf{46.71} & \textbf{58.00} & 81.25 & \textbf{27.59} & 30.00 & \textbf{38.75} & 61.25 & \textbf{82.50} & 31.71 & 11.88 & \textbf{48.43} \\

\bottomrule
\end{tabular}
}
}
\label{table:knowledge_score_recall@10}
\end{table*}

\begin{table*}[h]
\caption{Domain-wise NDCG@5 performance on ARK (knowledge axis). Columns correspond to ARK's five knowledge domains and their subtypes.}
\centering
\renewcommand{\arraystretch}{1.2}
\resizebox{\textwidth}{!}
{
{
\begin{tabular}{l|cccc|ccc|cc|ccccc|ccc|c}
\toprule
\multirow{2}{*}{\textbf{Model}} & \multicolumn{4}{c|}{\textbf{Visual Cognition}}
& \multicolumn{3}{c|}{\textbf{Natural Science}}
& \multicolumn{2}{c|}{\textbf{Formal Science}}
& \multicolumn{5}{c|}{\textbf{Humanities \& Social Science}}
& \multicolumn{3}{c|}{\textbf{Engineering \& Technology}}
& \multirow{2}{*}{\textbf{Avg.}} \\
 & Daily. & Fine. & SCog. & S3D. & Bio. & Phy. & Che. & Mat. & Cod. & Eco. & Comic. & Art & Cal. & Met. & Compu. & Mecha. & Ele. &  \\
\midrule
\multicolumn{19}{c}{\textit{\textbf{Text Embedding Models}}} \\
\midrule
BGE-M3-0.6B & 48.34 & 4.27 & 0.00 & 0.76 & 48.88 & 23.44 & 5.13 & 7.10 & 2.89 & 47.21 & 2.22 & 5.62 & 6.73 & 19.37 & 28.37 & 10.23 & 0.99 & 15.39 \\
Diver-Embed-1.7B & 42.55 & 4.62 & 0.00 & 0.30 & 41.85 & 36.70 & 6.37 & 19.97 & 4.27 & 37.19 & 1.64 & 5.43 & 9.54 & 17.94 & 45.96 & 10.33 & 0.88 & 16.80 \\
Diver-Embed-4B & 54.27 & 3.70 & 0.00 & 0.76 & 55.07 & 32.82 & 8.98 & \underline{23.41} & 8.35 & \underline{64.75} & 4.09 & 8.25 & 7.04 & 24.23 & 53.74 & 13.81 & 2.48 & 21.51 \\
Qwen3-Embed-4B & 50.07 & 5.20 & 4.32 & 1.63 & 50.08 & 32.21 & 8.31 & 20.23 & 8.10 & 51.32 & 2.60 & 6.31 & 6.16 & 13.77 & 42.66 & 16.43 & 4.97 & 19.08 \\
\midrule
Qwen3-Embed-8B & 54.28 & 6.13 & 1.68 & 1.73 & 56.86 & \underline{37.89} & 10.65 & 22.22 & 8.94 & 62.72 & 4.69 & 5.56 & \underline{11.69} & 14.92 & 53.09 & 15.52 & 6.12 & 22.04 \\
ReasonIR-8B & 50.58 & 4.83 & 0.00 & 0.76 & 45.79 & 29.48 & 8.17 & 16.84 & 3.56 & 60.16 & 7.10 & 2.88 & 5.34 & 27.97 & 49.51 & 13.59 & 2.36 & 19.35 \\
\midrule
\multicolumn{19}{c}{\textit{\textbf{Multimodal Embedding Models}}} \\
\midrule
CLIP-0.4B & 49.96 & 3.97 & 2.24 & 6.59 & 45.22 & 13.93 & 7.88 & 3.83 & 1.50 & 17.63 & 6.56 & 20.22 & 7.52 & 12.90 & 12.14 & 12.47 & 8.13 & 13.69 \\
BGE-VL-0.4B & 62.79 & 4.51 & 0.00 & 0.00 & 50.66 & 4.39 & 0.00 & 4.16 & 3.87 & 8.04 & 0.57 & 9.59 & 2.66 & 7.30 & 13.45 & 10.24 & 6.09 & 11.08 \\
SigLIP2-0.9B & \textbf{79.40} & 4.41 & 0.96 & 2.88 & 50.33 & 9.05 & 8.36 & 7.93 & 0.43 & 22.66 & 7.46 & 8.71 & 8.83 & 3.29 & 28.01 & \underline{21.09} & 3.43 & 15.72 \\
\midrule
MetaCLIP2-2B & 69.59 & \underline{7.75} & 2.33 & 5.36 & 56.65 & 29.14 & 4.95 & 4.98 & 5.34 & 31.81 & 4.17 & \textbf{31.26} & 4.62 & 7.97 & 21.12 & 19.92 & 6.54 & 18.44 \\
VLM2VecV2-2B & 54.85 & 4.87 & 5.91 & 3.51 & 43.95 & 16.67 & 13.69 & 18.53 & 6.97 & 46.34 & 2.79 & 7.19 & 5.02 & 19.77 & 30.94 & 14.27 & 3.12 & 17.55 \\
Ops-MM-Embed-v1-2B & 73.93 & 4.42 & 3.59 & 5.00 & 57.18 & 22.02 & 21.73 & 15.73 & 7.39 & 36.53 & 10.99 & 12.19 & 5.16 & 22.86 & 38.74 & 14.82 & 5.75 & 21.06 \\
Qwen3-VL-Embed-2B & 76.09 & 5.82 & 6.62 & 8.76 & 50.68 & 29.89 & 13.72 & 20.43 & 6.47 & 56.29 & 9.26 & 6.01 & 3.12 & 25.73 & 37.81 & 14.53 & 10.13 & 22.43 \\
\midrule
LamRA-Ret-7B & 69.57 & 6.87 & 0.53 & 1.95 & 58.92 & 7.57 & 4.47 & 17.63 & 14.50 & 50.39 & 6.13 & 12.35 & 3.12 & 17.71 & 5.79 & 11.45 & 2.67 & 17.15 \\
RzenEmbed-7B & 77.79 & 6.56 & \underline{10.29} & 3.20 & 52.16 & 28.09 & 20.79 & 22.94 & 18.30 & 56.80 & 9.16 & 12.03 & 9.17 & 25.77 & \underline{60.63} & 19.45 & \underline{11.57} & 26.16 \\
Ops-MM-Embed-v1-7B & 73.37 & 6.66 & \textbf{14.83} & \textbf{9.53} & \underline{60.12} & 30.13 & \underline{22.34} & 22.84 & 12.76 & 55.37 & 12.05 & 17.90 & 7.58 & \underline{30.96} & 52.47 & 17.40 & 11.15 & 26.91 \\
Qwen3-VL-Embed-8B & \underline{78.35} & 5.19 & 2.83 & \underline{8.95} & 59.87 & \textbf{40.18} & 20.87 & 20.89 & \textbf{29.38} & \textbf{65.62} & \underline{15.73} & 8.54 & 7.44 & 25.26 & 51.54 & 19.93 & \textbf{12.11} & \underline{27.80} \\
\midrule
EVA-CLIP-18B & 74.28 & 6.60 & 0.00 & 7.61 & \textbf{67.24} & 24.70 & 8.65 & 0.91 & 5.84 & 36.89 & 3.97 & \underline{28.31} & 5.33 & 9.89 & 20.65 & \textbf{25.87} & 5.01 & 19.51 \\
Seed1.6-embedding & 75.56 & \textbf{8.86} & 3.31 & 5.86 & 49.10 & 33.69 & \textbf{22.43} & \textbf{25.96} & \underline{27.12} & 61.55 & \textbf{17.45} & 17.97 & \textbf{20.44} & \textbf{32.13} & \textbf{64.57} & 17.91 & 5.51 & \textbf{28.79} \\
\bottomrule
\end{tabular}
}
}
\label{table:knowledge_score_ndcg_5}
\end{table*}

\begin{table*}[h]
\caption{Domain-wise NDCG@10 performance on ARK (knowledge axis). Columns correspond to ARK's five knowledge domains and their subtypes. }
\centering
\renewcommand{\arraystretch}{1.2}
\resizebox{\textwidth}{!}
{

{
\begin{tabular}{l|cccc|ccc|cc|ccccc|ccc|c}
\toprule
\multirow{2}{*}{\textbf{Model}} & \multicolumn{4}{c|}{\textbf{Visual Cognition}}
& \multicolumn{3}{c|}{\textbf{Natural Science}}
& \multicolumn{2}{c|}{\textbf{Formal Science}}
& \multicolumn{5}{c|}{\textbf{Humanities \& Social Science}}
& \multicolumn{3}{c|}{\textbf{Engineering \& Technology}}
& \multirow{2}{*}{\textbf{Avg.}} \\
 & Daily. & Fine. & SCog. & S3D. & Bio. & Phy. & Che. & Mat. & Cod. & Eco. & Comic. & Art & Cal. & Met. & Compu. & Mecha. & Ele. &  \\
\midrule
\multicolumn{19}{c}{\textit{\textbf{Text Embedding Models}}} \\
\midrule
BGE-M3-0.6B & 50.39 & 6.01 & 0.39 & 0.76 & 50.49 & 25.81 & 7.93 & 8.63 & 5.14 & 50.90 & 3.36 & 6.82 & 6.73 & 23.71 & 32.54 & 11.38 & 2.31 & 17.25 \\
Diver-Embed-1.7B & 45.15 & 5.25 & 0.00 & 0.55 & 46.45 & 37.91 & 9.47 & 21.46 & 6.48 & 39.67 & 1.64 & 7.10 & 10.35 & 24.25 & 48.01 & 11.89 & 2.85 & 18.73 \\
Diver-Embed-4B & 56.74 & 5.78 & 0.00 & 0.76 & 58.24 & 35.64 & 11.72 & 25.83 & 10.93 & \textbf{67.15} & 4.50 & 10.15 & 7.90 & 29.10 & 56.26 & 14.96 & 3.13 & 23.46 \\
Qwen3-Embed-4B & 52.81 & 7.54 & 7.82 & 2.12 & 52.82 & 34.58 & 9.87 & 22.94 & 10.01 & 55.59 & 3.26 & 7.55 & 6.52 & 22.21 & 45.21 & 17.18 & 7.51 & 21.50 \\
\midrule
Qwen3-Embed-8B & 57.04 & 7.57 & 1.68 & 2.93 & 59.96 & \underline{39.15} & 13.93 & 25.67 & 11.21 & \underline{65.25} & 5.89 & 7.89 & \underline{13.33} & 21.97 & 55.11 & 16.69 & 7.34 & 24.27 \\
ReasonIR-8B & 53.78 & 7.56 & 1.12 & 1.20 & 49.29 & 31.62 & 10.33 & 20.26 & 4.57 & 62.91 & 8.21 & 3.54 & 6.12 & 34.33 & 51.99 & 15.43 & 3.34 & 21.51 \\
\midrule
\multicolumn{19}{c}{\textit{\textbf{Multimodal Embedding Models}}} \\
\midrule
CLIP-0.4B & 52.55 & 6.98 & 3.43 & 7.32 & 53.06 & 17.03 & 8.73 & 5.26 & 2.82 & 22.40 & 7.70 & 23.57 & 8.76 & 16.46 & 13.70 & 13.59 & 9.12 & 16.03 \\
BGE-VL-0.4B & 65.77 & 5.99 & 0.00 & 0.00 & 56.80 & 5.59 & 0.00 & 6.28 & 4.46 & 12.26 & 0.57 & 11.27 & 3.47 & 13.64 & 14.27 & 11.35 & 6.37 & 12.83 \\
SigLIP2-0.9B & \textbf{80.65} & 4.74 & 2.46 & 3.15 & 53.44 & 11.80 & 10.25 & 9.65 & 0.43 & 24.66 & 7.79 & 9.36 & 12.87 & 6.94 & 30.47 & 22.63 & 4.09 & 17.38 \\
\midrule
MetaCLIP2-2B & 72.03 & 9.51 & 3.06 & 6.59 & 62.59 & 33.16 & 5.81 & 6.44 & 6.86 & 34.24 & 4.58 & \textbf{35.79} & 4.99 & 13.76 & 22.33 & \underline{23.12} & 7.15 & 20.71 \\
VLM2VecV2-2B & 57.69 & 7.00 & 13.28 & 4.67 & 47.37 & 18.77 & 17.73 & 22.92 & 9.97 & 48.35 & 3.56 & 8.54 & 5.80 & 25.80 & 32.14 & 15.06 & 4.80 & 20.20 \\
Ops-MM-Embed-v1-2B & 76.43 & 6.83 & 9.20 & 6.44 & 62.20 & 26.11 & 26.32 & 19.39 & 9.99 & 39.68 & 11.67 & 14.36 & 5.16 & 30.75 & 41.13 & 16.43 & 8.52 & 24.15 \\
Qwen3-VL-Embed-2B & 77.62 & 7.58 & 11.75 & 9.49 & 53.89 & 33.81 & 18.99 & 23.13 & 9.07 & 58.39 & 10.98 & 8.68 & 3.57 & 31.77 & 38.58 & 15.70 & \textbf{12.72} & 25.04 \\
\midrule
LamRA-Ret-7B & 72.74 & 8.66 & 3.97 & 3.26 & 63.35 & 11.91 & 5.41 & 19.70 & 18.61 & 54.46 & 7.63 & 15.45 & 3.12 & 22.42 & 6.23 & 11.83 & 2.67 & 19.50 \\
RzenEmbed-7B & \underline{80.43} & 7.98 & \underline{17.86} & 3.71 & 57.88 & 30.53 & \textbf{30.09} & 26.02 & 22.16 & 61.79 & 9.88 & 13.36 & 11.19 & 29.86 & \underline{61.75} & 21.37 & \underline{12.57} & 29.32 \\
Ops-MM-Embed-v1-7B & 75.54 & 9.85 & \textbf{24.66} & \textbf{11.24} & \underline{65.70} & 32.52 & \underline{29.72} & \underline{26.04} & 15.95 & 57.37 & 12.87 & 20.80 & 8.71 & \textbf{38.87} & 53.31 & 19.36 & 11.15 & 30.22 \\
Qwen3-VL-Embed-8B & 80.00 & 6.63 & 11.91 & \underline{10.97} & 64.67 & \textbf{43.16} & 27.19 & 25.46 & \underline{32.69} & \textbf{67.15} & \underline{16.47} & 11.19 & 8.72 & 30.27 & 53.17 & 21.53 & 12.11 & \underline{30.78} \\
\midrule
EVA-CLIP-18B & 76.10 & \underline{10.27} & 0.80 & 8.81 & \textbf{69.65} & 28.30 & 9.90 & 2.24 & 6.48 & 39.87 & 4.38 & \underline{32.02} & 5.33 & 13.34 & 23.90 & \textbf{28.65} & 6.80 & 21.58 \\
Seed1.6-embedding & 77.69 & \textbf{10.58} & 10.97 & 7.42 & 54.66 & 37.25 & 28.97 & \textbf{28.69} & \textbf{34.21} & 63.52 & \textbf{19.32} & 19.18 & \textbf{23.90} & \underline{36.41} & \textbf{68.36} & 20.60 & 7.06 & \textbf{32.28} \\
\bottomrule
\end{tabular}
}
}
\label{table:knowledge_score_ndcg_10}
\end{table*}

\begin{table*}[t]
\caption{Domain-wise NDCG@20 performance on ARK (knowledge axis). Columns correspond to ARK's five knowledge domains and their subtypes.}
\centering
\renewcommand{\arraystretch}{1.2}
\resizebox{\textwidth}{!}
{
{
\begin{tabular}{l|cccc|ccc|cc|ccccc|ccc|c}
\toprule
\multirow{2}{*}{\textbf{Model}} & \multicolumn{4}{c|}{\textbf{Visual Cognition}}
& \multicolumn{3}{c|}{\textbf{Natural Science}}
& \multicolumn{2}{c|}{\textbf{Formal Science}}
& \multicolumn{5}{c|}{\textbf{Humanities \& Social Science}}
& \multicolumn{3}{c|}{\textbf{Engineering \& Technology}}
& \multirow{2}{*}{\textbf{Avg.}} \\
 & Daily. & Fine. & SCog. & S3D. & Bio. & Phy. & Che. & Mat. & Cod. & Eco. & Comic. & Art & Cal. & Met. & Compu. & Mecha. & Ele. &  \\
\midrule
\multicolumn{19}{c}{\textit{\textbf{Text Embedding Models}}} \\
\midrule
BGE-M3-0.6B & 52.68 & 7.61 & 1.92 & 1.16 & 52.64 & 27.62 & 10.59 & 11.11 & 6.14 & 54.08 & 4.25 & 7.76 & 6.73 & 29.55 & 33.71 & 13.54 & 2.54 & 19.04 \\
Diver-Embed-1.7B & 47.83 & 6.64 & 0.60 & 1.13 & 48.88 & 39.80 & 11.89 & 23.79 & 8.49 & 42.31 & 1.95 & 8.58 & 10.94 & 29.00 & 49.27 & 12.49 & 4.07 & 20.45 \\
Diver-Embed-4B & 58.72 & 7.38 & 0.32 & 0.95 & 60.07 & 37.29 & 12.92 & 28.04 & 12.67 & \textbf{68.75} & 4.50 & 11.65 & 10.77 & 33.72 & 56.93 & 15.92 & 4.15 & 24.99 \\
Qwen3-Embed-4B & 54.43 & 7.84 & 9.41 & 2.89 & 54.90 & 35.86 & 11.78 & 23.98 & 11.27 & 57.47 & 3.57 & 8.84 & 7.10 & 28.58 & 46.74 & 18.12 & 8.99 & 23.05 \\
Qwen3-Embed-8B & 59.31 & 9.23 & 3.20 & 3.11 & 62.11 & \underline{40.62} & 15.88 & 28.20 & 12.71 & 65.56 & 6.53 & 9.90 & \underline{14.57} & 26.91 & 56.39 & 17.02 & 9.61 & 25.93 \\
ReasonIR-8B & 55.72 & 9.36 & 2.99 & 1.58 & 51.44 & 32.86 & 11.47 & 21.90 & 6.32 & 63.60 & 9.38 & 5.53 & 6.79 & 37.60 & 52.95 & 16.62 & 5.01 & 23.01 \\
\midrule
\multicolumn{19}{c}{\textit{\textbf{Multimodal Embedding Models}}} \\
\midrule
CLIP-0.4B & 54.77 & 7.83 & 5.29 & 8.47 & 55.46 & 19.73 & 10.64 & 8.28 & 4.29 & 25.37 & 9.39 & 26.09 & 10.92 & 23.84 & 14.65 & 16.09 & 9.39 & 18.26 \\
BGE-VL-0.4B & 67.17 & 7.60 & 0.00 & 0.18 & 59.10 & 6.56 & 0.00 & 7.94 & 5.67 & 13.79 & 1.15 & 14.10 & 4.40 & 22.21 & 14.27 & 13.17 & 7.45 & 14.40 \\
SigLIP2-0.9B & \textbf{81.93} & 5.59 & 5.64 & 3.56 & 54.08 & 13.66 & 12.46 & 11.30 & 0.43 & 26.55 & 8.65 & 11.10 & 14.14 & 8.44 & 32.05 & 24.13 & 4.33 & 18.71 \\
\midrule
MetaCLIP2-2B & 73.14 & \textbf{12.89} & 4.81 & 7.35 & 63.50 & 34.44 & 6.97 & 9.31 & 8.20 & 36.78 & 5.20 & \textbf{37.83} & 4.99 & 21.45 & 24.59 & \underline{24.99} & 7.62 & 22.59 \\
VLM2VecV2-2B & 59.54 & 7.81 & 16.73 & 6.43 & 49.23 & 20.65 & 19.88 & 25.05 & 13.53 & 50.56 & 4.48 & 10.31 & 6.74 & 29.88 & 33.67 & 16.00 & 5.07 & 22.09 \\
Ops-MM-Embed-v1-2B & 77.19 & 7.63 & 15.44 & 8.25 & 63.78 & 27.44 & 29.64 & 20.59 & 12.49 & 41.65 & 12.52 & 17.53 & 6.74 & 39.23 & 41.79 & 18.32 & 9.31 & 26.44 \\
Qwen3-VL-Embed-2B & 78.84 & 8.64 & 18.74 & 11.23 & 56.70 & 35.05 & 22.15 & 25.60 & 11.00 & 59.06 & 12.46 & 10.16 & 3.87 & 36.01 & 40.44 & 16.88 & \textbf{13.51} & 27.08 \\
\midrule
LamRA-Ret-7B & 74.52 & 11.09 & 6.73 & 3.82 & 64.32 & 14.50 & 7.29 & 21.50 & 21.62 & 55.78 & 9.65 & 17.18 & 4.15 & 27.59 & 8.07 & 12.40 & 2.93 & 21.36 \\
RzenEmbed-7B & \underline{81.39} & 10.72 & \underline{23.55} & 5.24 & 59.99 & 32.43 & \textbf{32.79} & \underline{28.50} & 26.79 & 63.07 & 11.32 & 15.17 & 12.74 & 35.03 & \underline{62.69} & 23.86 & \underline{12.84} & 31.65 \\
Ops-MM-Embed-v1-7B & 76.94 & 10.99 & \textbf{28.14} & \textbf{12.41} & \underline{66.97} & 34.45 & \underline{31.69} & 27.87 & 20.74 & 59.88 & 13.78 & 22.79 & 10.26 & \textbf{43.11} & 56.17 & 20.85 & 11.66 & 32.28 \\
Qwen3-VL-Embed-8B & 81.17 & 9.30 & 13.81 & \underline{11.74} & 65.90 & \textbf{43.75} & 30.95 & 26.98 & \underline{37.35} & \underline{67.82} & \underline{18.77} & 14.75 & 10.98 & 36.58 & 55.67 & 22.74 & 12.11 & \underline{32.96} \\
\midrule
EVA-CLIP-18B & 77.06 & 11.92 & 1.08 & 10.39 & \textbf{70.88} & 28.91 & 12.27 & 4.02 & 7.20 & 41.81 & 5.83 & \underline{34.04} & 6.55 & 20.62 & 25.17 & \textbf{30.43} & 7.79 & 23.29 \\
Seed1.6-embedding & 78.55 & \underline{12.84} & 16.74 & 7.82 & 57.77 & 38.28 & 31.43 & \textbf{30.50} & \textbf{38.50} & 65.07 & \textbf{20.27} & 22.70 & \textbf{25.76} & \underline{40.95} & \textbf{69.62} & 24.30 & 10.01 & \textbf{34.77} \\

\bottomrule
\end{tabular}
}
}
\label{table:knowledge_score_ndcg_20}
\end{table*}

\begin{table*}[t]
\caption{Recall@5 on ARK by cognitive demand and reasoning skill (reasoning axis). ``Both'' denotes instances that are both knowledge-intensive and reasoning-intensive. }

\renewcommand{\arraystretch}{1.1}
\centering
\resizebox{\textwidth}{!}{
\tablestyle{2.5pt}{1.1}{
\begin{tabular}{l|cccc|ccccccc}
\toprule

\multirow{2}{*}{\textbf{Model}}
& \multicolumn{4}{c|}{\textbf{Cognitive Demand}}
& \multicolumn{7}{c}{\textbf{Reasoning Skill}} \\
& \makecell{\textbf{Perception-}\\\textbf{Only}}
& \makecell{\textbf{Knowledge-}\\\textbf{Intensive}}
& \makecell{\textbf{Reasoning-}\\\textbf{Intensive}}
& \textbf{Both}
& \makecell{\textbf{Knowledge}\\\textbf{Reasoning}}
& \makecell{\textbf{Conceptual}\\\textbf{Abstraction}}
& \makecell{\textbf{Fine-Grained}\\\textbf{Visual Reasoning}}
& \makecell{\textbf{Logical}\\\textbf{Reasoning}}
& \makecell{\textbf{Spatial}\\\textbf{Reasoning}}
& \makecell{\textbf{Symbolic}\\\textbf{Reasoning}}
& \textbf{Average}
\\

\midrule
\multicolumn{12}{c}{\textit{\textbf{Text Embedding Models}}} \\
\midrule

BGE-M3-0.6B & 57.63 & 24.44 & 19.07 & 22.00 & 23.25 & 31.88 & 12.31 & 14.11 & 11.92 & 14.07 & 17.92 \\
Diver-Embed-1.7B & 49.15 & 28.74 & 21.82 & 26.84 & 28.67 & 37.55 & 13.85 & 15.34 & 8.46 & 21.21 & 20.85 \\
Diver-Embed-4B & 61.02 & 32.68 & 24.73 & 30.91 & 33.38 & 43.67 & 13.85 & \underline{17.79} & 8.46 & 24.68 & 23.64 \\
Qwen3-Embed-4B & 57.63 & 25.96 & 18.38 & 22.66 & 26.68 & 35.81 & 7.69 & 7.36 & 8.46 & 16.23 & 17.04 \\
\midrule
Qwen3-Embed-8B & 61.44 & 30.17 & 22.05 & 27.28 & 31.81 & 44.54 & 12.31 & 13.50 & 8.46 & 18.83 & 21.57 \\
ReasonIR-8B & 59.75 & 29.36 & 22.43 & 27.72 & 29.53 & 44.98 & 14.62 & 11.04 & 8.46 & 20.56 & 21.53 \\
\midrule
 \multicolumn{12}{c}{\textit{\textbf{Multimodal Embedding Models}}} \\
\midrule
CLIP-0.4B & 61.44 & 17.46 & 13.02 & 13.75 & 15.69 & 17.47 & 11.54 & 10.43 & 11.54 & 7.58 & 12.37 \\
BGE-VL-0.4B & 72.46 & 12.53 & 8.04 & 9.24 & 9.27 & 8.73 & 12.31 & 7.36 & 1.92 & 6.28 & 7.65 \\
SigLIP2-0.9B & \textbf{88.98} & 18.98 & 13.09 & 16.39 & 20.40 & 19.65 & 6.92 & 7.98 & 4.62 & 9.31 & 11.48 \\
\midrule
MetaCLIP2-2B & 78.39 & 23.46 & 16.16 & 18.59 & 21.68 & 24.02 & 13.85 & 9.20 & 8.46 & 8.87 & 14.35 \\
VLM2VecV2-2B & 64.83 & 23.99 & 19.60 & 22.22 & 27.10 & 31.44 & 13.85 & 7.98 & 13.08 & 18.40 & 18.64 \\
Ops-MM-Embed-v1-2B & 83.05 & 28.38 & 21.82 & 25.85 & 30.53 & 32.31 & 14.62 & 8.59 & 11.54 & 22.08 & 19.95 \\
Qwen3-VL-Embed-2B & 85.59 & 29.19 & 24.27 & 26.62 & 30.81 & 39.74 & 20.77 & 12.27 & \underline{18.08} & 22.29 & 23.99 \\
\midrule
LamRA-Ret-7B & 79.24 & 22.20 & 16.77 & 19.91 & 21.83 & 29.26 & 13.85 & 4.91 & 8.08 & 16.02 & 15.66 \\
RzenEmbed-7B & 84.75 & 36.08 & 29.56 & 34.76 & 38.52 & 45.41 & \textbf{23.85} & 15.95 & 14.23 & 29.65 & 27.94 \\
Ops-MM-Embed-v1-7B & 83.05 & 35.99 & \underline{30.93} & 34.54 & \underline{39.66} & 45.41 & \underline{23.08} & 15.34 & \textbf{22.31} & 30.30 & \underline{29.35} \\
Qwen3-VL-Embed-8B & \underline{86.44} & \underline{38.05} & 30.40 & \underline{36.19} & 38.09 & \underline{47.16} & 20.77 & 15.34 & 15.77 & \underline{33.55} & 28.45 \\
\midrule
EVA-CLIP-18B & 83.05 & 23.28 & 15.39 & 17.49 & 20.68 & 26.64 & 12.31 & 6.13 & 9.23 & 6.28 & 13.55 \\
Seed1.6-embedding & 84.32 & \textbf{45.41} & \textbf{35.35} & \textbf{44.31} & \textbf{45.88} & \textbf{55.38} & 14.13 & \textbf{19.64} & 14.16 & \textbf{40.96} & \textbf{31.69} \\
\bottomrule
\end{tabular}
}
}
\label{table:reasoning_score_recall@5}
\end{table*}

\begin{table*}[t]
\caption{Recall@10 on ARK by cognitive demand and reasoning skill (reasoning axis). ``Both'' denotes instances that are both knowledge-intensive and reasoning-intensive. }

\renewcommand{\arraystretch}{1.1}
\centering
\resizebox{\textwidth}{!}{
\tablestyle{2.5pt}{1.1}{
\begin{tabular}{l|cccc|ccccccc}
\toprule

\multirow{2}{*}{\textbf{Model}}
& \multicolumn{4}{c|}{\textbf{Cognitive Demand}}
& \multicolumn{7}{c}{\textbf{Reasoning Skill}} \\
& \makecell{\textbf{Perception-}\\\textbf{Only}}
& \makecell{\textbf{Knowledge-}\\\textbf{Intensive}}
& \makecell{\textbf{Reasoning-}\\\textbf{Intensive}}
& \textbf{Both}
& \makecell{\textbf{Knowledge}\\\textbf{Reasoning}}
& \makecell{\textbf{Conceptual}\\\textbf{Abstraction}}
& \makecell{\textbf{Fine-Grained}\\\textbf{Visual Reasoning}}
& \makecell{\textbf{Logical}\\\textbf{Reasoning}}
& \makecell{\textbf{Spatial}\\\textbf{Reasoning}}
& \makecell{\textbf{Symbolic}\\\textbf{Reasoning}}
& \textbf{Average}
\\

\midrule
\multicolumn{12}{c}{\textit{\textbf{Text Embedding Models}}} \\
\midrule

BGE-M3-0.6B & 63.98 & 29.81 & 23.74 & 27.17 & 27.67 & 37.12 & 17.69 & 17.18 & 14.62 & 20.56 & 22.47 \\
Diver-Embed-1.7B & 56.78 & 34.65 & 26.80 & 31.90 & 33.38 & 40.61 & 21.54 & 19.02 & 11.54 & 27.06 & 25.53 \\
Diver-Embed-4B & 69.92 & 39.03 & 30.70 & 37.29 & 39.94 & 49.34 & 23.08 & \underline{23.31} & 11.54 & 30.52 & 29.62 \\
Qwen3-Embed-4B & 65.25 & 32.59 & 26.34 & 29.81 & 33.95 & 43.23 & 20.77 & 12.88 & 17.31 & 23.81 & 25.32 \\
\midrule
Qwen3-Embed-8B & 70.34 & 37.42 & 29.56 & 35.20 & 40.37 & 50.22 & 23.85 & 18.40 & 13.08 & 27.49 & 28.90 \\
ReasonIR-8B & 69.49 & 35.90 & 29.48 & 34.32 & 36.38 & 48.91 & 27.69 & 17.18 & 13.46 & 28.35 & 28.66 \\
\midrule
 \multicolumn{12}{c}{\textit{\textbf{Multimodal Embedding Models}}} \\
\midrule
CLIP-0.4B & 69.07 & 24.62 & 19.30 & 19.91 & 22.82 & 24.45 & 22.31 & 13.50 & 15.77 & 11.47 & 18.39 \\
BGE-VL-0.4B & 81.78 & 17.19 & 12.25 & 12.98 & 13.69 & 13.54 & 20.00 & 8.59 & 6.15 & 9.31 & 11.88 \\
SigLIP2-0.9B & \underline{92.80} & 24.17 & 17.92 & 21.34 & 26.53 & 24.89 & 14.62 & 14.11 & 7.31 & 13.20 & 16.78 \\
\midrule
MetaCLIP2-2B & 86.02 & 30.62 & 22.13 & 24.42 & 28.25 & 30.13 & 22.31 & 10.43 & 14.23 & 13.42 & 19.80 \\
VLM2VecV2-2B & 74.58 & 31.33 & 28.64 & 30.25 & 35.52 & 35.81 & 26.15 & 12.27 & 24.23 & 28.57 & 27.09 \\
Ops-MM-Embed-v1-2B & 90.68 & 37.51 & 31.39 & 34.32 & 39.23 & 38.43 & \underline{29.23} & 14.11 & 22.69 & 33.12 & 29.47 \\
Qwen3-VL-Embed-2B & 90.68 & 37.15 & 32.77 & 34.76 & 38.94 & 44.54 & 27.69 & 18.40 & 28.85 & 33.12 & 31.92 \\
\midrule
LamRA-Ret-7B & 88.56 & 28.65 & 23.43 & 26.07 & 27.96 & 35.81 & 21.54 & 4.91 & 16.15 & 21.65 & 21.34 \\
RzenEmbed-7B & \textbf{93.22} & 45.84 & 39.20 & \underline{44.77} & \underline{49.22} & \underline{52.84} & 29.23 & 21.47 & 25.77 & 42.42 & 36.82 \\
Ops-MM-Embed-v1-7B & 90.25 & 44.05 & \underline{40.74} & 42.35 & 48.36 & 49.78 & \textbf{34.62} & 15.95 & \textbf{39.23} & 39.83 & \underline{37.96} \\
Qwen3-VL-Embed-8B & 91.53 & \underline{46.37} & 39.74 & 44.22 & 47.08 & 51.53 & 25.38 & 17.18 & \underline{31.92} & \underline{43.94} & 36.17 \\
\midrule
EVA-CLIP-18B & 88.56 & 29.27 & 20.83 & 22.55 & 26.11 & 32.75 & 23.08 & 9.82 & 13.85 & 10.17 & 19.30 \\
Seed1.6-embedding & 89.83 & \textbf{56.30} & \textbf{45.51} & \textbf{55.03} & \textbf{55.83} & \textbf{61.29} & 18.48 & \textbf{30.36} & 24.78 & \textbf{53.46} & \textbf{40.70} \\
\bottomrule
\end{tabular}
}
}
\label{table:reasoning_score_recall@10}
\end{table*}

\begin{table*}[t]
\caption{NDCG@5 on ARK by cognitive demand and reasoning skill (reasoning axis). ``Both'' denotes instances that are both knowledge-intensive and reasoning-intensive. }

\renewcommand{\arraystretch}{1.1}
\centering
\resizebox{\textwidth}{!}{
\tablestyle{2.5pt}{1.1}{
\begin{tabular}{l|cccc|ccccccc}
\toprule
\multirow{2}{*}{\textbf{Model}}
& \multicolumn{4}{c|}{\textbf{Cognitive Demand}}
& \multicolumn{7}{c}{\textbf{Reasoning Skill}} \\
& \makecell{\textbf{Perception-}\\\textbf{Only}}
& \makecell{\textbf{Knowledge-}\\\textbf{Intensive}}
& \makecell{\textbf{Reasoning-}\\\textbf{Intensive}}
& \textbf{Both}
& \makecell{\textbf{Knowledge}\\\textbf{Reasoning}}
& \makecell{\textbf{Conceptual}\\\textbf{Abstraction}}
& \makecell{\textbf{Fine-Grained}\\\textbf{Visual Reasoning}}
& \makecell{\textbf{Logical}\\\textbf{Reasoning}}
& \makecell{\textbf{Spatial}\\\textbf{Reasoning}}
& \makecell{\textbf{Symbolic}\\\textbf{Reasoning}}
& \textbf{Average}
\\

\midrule
\multicolumn{12}{c}{\textit{\textbf{Text Embedding Models}}} \\
\midrule

BGE-M3-0.6B & 48.34 & 18.58 & 13.65 & 16.35 & 17.73 & 25.26 & 7.34 & 9.35 & 7.26 & 9.27 & 12.70 \\
Diver-Embed-1.7B & 42.77 & 22.21 & 16.18 & 20.44 & 22.13 & 30.47 & 8.15 & 10.41 & 5.75 & 15.01 & 15.32 \\
Diver-Embed-4B & 54.21 & 25.68 & 18.65 & 23.84 & 26.17 & 36.12 & 8.97 & \underline{12.15} & 5.07 & 17.20 & 17.61 \\
Qwen3-Embed-4B & 49.92 & 19.48 & 13.40 & 16.73 & 19.90 & 27.62 & 4.98 & 5.84 & 5.60 & 11.08 & 12.50 \\
\midrule
Qwen3-Embed-8B & 53.86 & 23.16 & 15.99 & 20.44 & 24.20 & 34.66 & 6.41 & 9.15 & 5.09 & 12.98 & 15.42 \\
ReasonIR-8B & 50.58 & 22.52 & 16.97 & 21.28 & 23.25 & 37.02 & 9.20 & 7.22 & 6.37 & 14.19 & 16.21 \\
\midrule
 \multicolumn{12}{c}{\textit{\textbf{Multimodal Embedding Models}}} \\
\midrule
CLIP-0.4B & 50.07 & 12.30 & 8.58 & 9.46 & 10.81 & 11.71 & 5.74 & 7.26 & 7.03 & 5.15 & 7.95 \\
BGE-VL-0.4B & 63.14 & 8.94 & 5.31 & 6.46 & 6.80 & 6.36 & 6.34 & 4.77 & 1.06 & 3.63 & 4.83 \\
SigLIP2-0.9B & \textbf{79.42} & 13.63 & 9.23 & 11.69 & 14.63 & 14.15 & 4.86 & 5.58 & 2.66 & 5.75 & 7.94 \\
\midrule
MetaCLIP2-2B & 69.78 & 16.98 & 11.37 & 13.37 & 15.74 & 17.92 & 8.44 & 5.63 & 5.83 & 5.36 & 9.82 \\
VLM2VecV2-2B & 55.10 & 16.62 & 13.22 & 15.43 & 19.01 & 24.28 & 8.77 & 4.66 & 7.40 & 11.40 & 12.59 \\
Ops-MM-Embed-v1-2B & 74.19 & 19.98 & 14.99 & 17.86 & 21.28 & 24.10 & 9.10 & 6.06 & 8.03 & 13.16 & 13.62 \\
Qwen3-VL-Embed-2B & 76.35 & 20.90 & 16.63 & 19.17 & 22.34 & 32.50 & 11.99 & 8.12 & \underline{10.48} & 13.81 & 16.54 \\
\midrule
LamRA-Ret-7B & 69.65 & 16.08 & 11.89 & 14.43 & 16.26 & 22.18 & 9.22 & 3.73 & 4.62 & 10.75 & 11.13 \\
RzenEmbed-7B & 77.94 & 25.81 & 20.25 & 24.57 & 27.65 & 34.81 & \underline{14.49} & 11.43 & 8.06 & 19.04 & 19.25 \\
Ops-MM-Embed-v1-7B & 73.97 & 26.32 & \underline{21.65} & 24.76 & \underline{28.69} & 35.02 & \textbf{14.63} & 10.71 & \textbf{14.44} & 18.68 & \underline{20.36} \\
Qwen3-VL-Embed-8B & \underline{78.00} & \underline{28.24} & 21.49 & \underline{26.51} & 28.37 & \underline{39.53} & 10.93 & 11.30 & 9.77 & \underline{21.38} & 20.21 \\
\midrule
EVA-CLIP-18B & 74.28 & 17.75 & 11.21 & 13.13 & 15.49 & 18.82 & 7.66 & 4.96 & 6.01 & 4.80 & 9.62 \\
Seed1.6-embedding & 74.21 & \textbf{34.29} & \textbf{26.52} & \textbf{33.54} & \textbf{34.60} & \textbf{45.18} & 9.36 & \textbf{14.99} & 10.37 & \textbf{28.33} & \textbf{23.80} \\
\bottomrule
\end{tabular}
}
}
\label{table:reasoning_score_ndcg@5}
\end{table*}

\begin{table*}[t]
\caption{NDCG@10 on ARK by cognitive demand and reasoning skill (reasoning axis). ``Both'' denotes instances that are both knowledge-intensive and reasoning-intensive. }

\renewcommand{\arraystretch}{1.1}
\centering
\resizebox{\textwidth}{!}{
\tablestyle{2.5pt}{1.1}{
\begin{tabular}{l|cccc|ccccccc}
\toprule
\multirow{2}{*}{\textbf{Model}}
& \multicolumn{4}{c|}{\textbf{Cognitive Demand}}
& \multicolumn{7}{c}{\textbf{Reasoning Skill}} \\
& \makecell{\textbf{Perception-}\\\textbf{Only}}
& \makecell{\textbf{Knowledge-}\\\textbf{Intensive}}
& \makecell{\textbf{Reasoning-}\\\textbf{Intensive}}
& \textbf{Both}
& \makecell{\textbf{Knowledge}\\\textbf{Reasoning}}
& \makecell{\textbf{Conceptual}\\\textbf{Abstraction}}
& \makecell{\textbf{Fine-Grained}\\\textbf{Visual Reasoning}}
& \makecell{\textbf{Logical}\\\textbf{Reasoning}}
& \makecell{\textbf{Spatial}\\\textbf{Reasoning}}
& \makecell{\textbf{Symbolic}\\\textbf{Reasoning}}
& \textbf{Average}
\\

\midrule
\multicolumn{12}{c}{\textit{\textbf{Text Embedding Models}}} \\
\midrule

BGE-M3-0.6B & 50.39 & 20.33 & 15.18 & 18.04 & 19.19 & 26.99 & 9.04 & 10.37 & 8.17 & 11.41 & 14.20 \\
Diver-Embed-1.7B & 45.23 & 24.12 & 17.78 & 22.07 & 23.64 & 31.43 & 10.53 & 11.62 & 6.77 & 16.90 & 16.81 \\
Diver-Embed-4B & 57.07 & 27.73 & 20.60 & 25.90 & 28.29 & 37.94 & 12.02 & \underline{13.92} & 6.09 & 19.12 & 19.56 \\
Qwen3-Embed-4B & 52.47 & 21.63 & 15.98 & 19.06 & 22.28 & 30.10 & 9.11 & 7.59 & 8.45 & 13.52 & 15.17 \\
\midrule
Qwen3-Embed-8B & 56.74 & 25.52 & 18.41 & 23.01 & 27.00 & 36.55 & 10.05 & 10.70 & 6.55 & 15.77 & 17.77 \\
ReasonIR-8B & 53.78 & 24.66 & 19.26 & 23.44 & 25.49 & 38.36 & 13.34 & 9.20 & 8.02 & 16.72 & 18.52 \\
\midrule
 \multicolumn{12}{c}{\textit{\textbf{Multimodal Embedding Models}}} \\
\midrule
CLIP-0.4B & 52.54 & 14.56 & 10.58 & 11.42 & 13.06 & 13.97 & 9.05 & 8.29 & 8.39 & 6.38 & 9.86 \\
BGE-VL-0.4B & 66.15 & 10.47 & 6.67 & 7.67 & 8.25 & 7.98 & 8.79 & 5.14 & 2.42 & 4.58 & 6.19 \\
SigLIP2-0.9B & \underline{80.67} & 15.31 & 10.79 & 13.29 & 16.61 & 15.83 & 7.35 & 7.60 & 3.50 & 7.02 & 9.65 \\
\midrule
MetaCLIP2-2B & 72.22 & 19.29 & 13.27 & 15.25 & 17.87 & 19.89 & 10.98 & 6.00 & 7.66 & 6.77 & 11.53 \\
VLM2VecV2-2B & 58.27 & 18.99 & 16.15 & 18.03 & 21.71 & 25.71 & 12.71 & 6.08 & 11.04 & 14.70 & 15.33 \\
Ops-MM-Embed-v1-2B & 76.64 & 22.91 & 18.06 & 20.58 & 24.08 & 26.00 & 13.71 & 7.77 & 11.65 & 16.73 & 16.66 \\
Qwen3-VL-Embed-2B & 77.98 & 23.52 & 19.41 & 21.86 & 25.04 & 34.09 & 14.17 & 10.16 & 13.96 & 17.39 & 19.13 \\
\midrule
LamRA-Ret-7B & 72.71 & 18.12 & 14.01 & 16.38 & 18.20 & 24.28 & 11.72 & 3.73 & 7.18 & 12.52 & 12.94 \\
RzenEmbed-7B & \textbf{80.71} & 29.02 & 23.41 & 27.86 & 31.18 & 37.25 & \underline{16.25} & 13.26 & 11.81 & 23.25 & 22.17 \\
Ops-MM-Embed-v1-7B & 76.29 & 28.94 & \underline{24.83} & 27.30 & \underline{31.53} & 36.45 & \textbf{18.46} & 10.88 & \textbf{19.87} & 21.75 & \underline{23.16} \\
Qwen3-VL-Embed-8B & 79.65 & \underline{30.98} & 24.55 & \underline{29.15} & 31.32 & \underline{40.94} & 12.48 & 11.93 & \underline{15.01} & \underline{24.79} & 22.75 \\
\midrule
EVA-CLIP-18B & 76.10 & 19.68 & 12.94 & 14.74 & 17.22 & 20.82 & 11.20 & 6.07 & 7.44 & 6.03 & 11.46 \\
Seed1.6-embedding & 76.01 & \textbf{37.85} & \textbf{29.84} & \textbf{37.04} & \textbf{37.87} & \textbf{47.03} & 10.73 & \textbf{18.30} & 13.88 & \textbf{32.44} & \textbf{26.71} \\
\bottomrule
\end{tabular}
}
}
\label{table:reasoning_score_ndcg@10}
\end{table*}

\begin{table*}[t]
\caption{NDCG@20 on ARK by cognitive demand and reasoning skill (reasoning axis). ``Both'' denotes instances that are both knowledge-intensive and reasoning-intensive. }

\renewcommand{\arraystretch}{1.1}
\centering
\resizebox{\textwidth}{!}{
\tablestyle{2.5pt}{1.1}{
\begin{tabular}{l|cccc|ccccccc}
\toprule
\multirow{2}{*}{\textbf{Model}}
& \multicolumn{4}{c|}{\textbf{Cognitive Demand}}
& \multicolumn{7}{c}{\textbf{Reasoning Skill}} \\
& \makecell{\textbf{Perception-}\\\textbf{Only}}
& \makecell{\textbf{Knowledge-}\\\textbf{Intensive}}
& \makecell{\textbf{Reasoning-}\\\textbf{Intensive}}
& \textbf{Both}
& \makecell{\textbf{Knowledge}\\\textbf{Reasoning}}
& \makecell{\textbf{Conceptual}\\\textbf{Abstraction}}
& \makecell{\textbf{Fine-Grained}\\\textbf{Visual Reasoning}}
& \makecell{\textbf{Logical}\\\textbf{Reasoning}}
& \makecell{\textbf{Spatial}\\\textbf{Reasoning}}
& \makecell{\textbf{Symbolic}\\\textbf{Reasoning}}
& \textbf{Average}
\\

\midrule
\multicolumn{12}{c}{\textit{\textbf{Text Embedding Models}}} \\
\midrule

BGE-M3-0.6B & 52.68 & 22.31 & 17.11 & 19.98 & 21.26 & 28.72 & 11.54 & 12.22 & 9.52 & 13.67 & 16.16 \\
Diver-Embed-1.7B & 47.88 & 25.87 & 19.48 & 23.89 & 25.48 & 32.76 & 12.32 & 13.47 & 7.85 & 19.05 & 18.49 \\
Diver-Embed-4B & 58.53 & 29.15 & 22.03 & 27.36 & 29.81 & 38.41 & 14.16 & \underline{14.87} & 7.09 & 20.95 & 20.88 \\
Qwen3-Embed-4B & 54.32 & 22.99 & 17.51 & 20.39 & 23.61 & 31.19 & 11.39 & 8.79 & 10.36 & 15.02 & 16.73 \\
\midrule
Qwen3-Embed-8B & 59.11 & 27.11 & 20.04 & 24.56 & 28.46 & 37.21 & 12.48 & 12.57 & 8.02 & 17.93 & 19.44 \\
ReasonIR-8B & 55.73 & 26.16 & 20.62 & 24.84 & 26.98 & 39.46 & 15.22 & 10.28 & 9.03 & 18.08 & 19.84 \\
\midrule
 \multicolumn{12}{c}{\textit{\textbf{Multimodal Embedding Models}}} \\
\midrule
CLIP-0.4B & 54.88 & 16.59 & 12.70 & 13.37 & 15.28 & 16.06 & 11.85 & 9.22 & 10.63 & 8.14 & 11.86 \\
BGE-VL-0.4B & 67.55 & 11.76 & 8.12 & 8.82 & 9.45 & 8.87 & 13.10 & 6.27 & 3.74 & 5.58 & 7.83 \\
SigLIP2-0.9B & \textbf{81.85} & 16.62 & 12.03 & 14.54 & 18.16 & 17.38 & 8.12 & 8.20 & 4.99 & 8.12 & 10.83 \\
\midrule
MetaCLIP2-2B & 73.34 & 20.82 & 15.16 & 16.76 & 19.57 & 21.32 & 15.82 & 6.29 & 9.63 & 8.20 & 13.47 \\
VLM2VecV2-2B & 59.89 & 20.73 & 18.04 & 19.75 & 23.40 & 26.93 & 14.63 & 6.56 & 13.69 & 16.93 & 17.02 \\
Ops-MM-Embed-v1-2B & 77.42 & 24.73 & 20.48 & 22.43 & 25.97 & 27.48 & 16.58 & 8.56 & 16.12 & 18.73 & 18.91 \\
Qwen3-VL-Embed-2B & 79.08 & 25.40 & 21.66 & 23.81 & 27.21 & 35.97 & 16.01 & 10.65 & \underline{17.59} & 19.60 & 21.17 \\
\midrule
LamRA-Ret-7B & 74.20 & 19.68 & 15.85 & 17.96 & 19.82 & 25.97 & 16.26 & 4.06 & 8.61 & 14.31 & 14.84 \\
RzenEmbed-7B & \underline{81.56} & 31.07 & 25.76 & 29.77 & 32.95 & 38.45 & \underline{19.42} & 13.88 & 15.45 & 25.75 & 24.32 \\
Ops-MM-Embed-v1-7B & 77.47 & 30.97 & \underline{26.85} & 29.20 & 33.20 & 38.01 & \textbf{20.66} & 11.65 & \textbf{22.26} & 24.03 & \underline{24.97} \\
Qwen3-VL-Embed-8B & 80.94 & \underline{33.00} & 26.78 & \underline{31.21} & \underline{33.34} & \underline{42.49} & 16.91 & 12.73 & 16.76 & \underline{27.10} & 24.89 \\
\midrule
EVA-CLIP-18B & 77.06 & 21.16 & 14.72 & 16.22 & 18.87 & 22.39 & 14.33 & 6.83 & 9.68 & 7.53 & 13.27 \\
Seed1.6-embedding & 77.29 & \textbf{40.25} & \textbf{32.17} & \textbf{39.33} & \textbf{39.64} & \textbf{48.52} & 13.74 & \textbf{21.63} & 16.27 & \textbf{35.28} & \textbf{29.18} \\
\bottomrule
\end{tabular}
}
}
\label{table:reasoning_score_ndcg@20}
\end{table*}

\clearpage
\section{Bootstrap Uncertainty Analysis of Main Results}
\label{appendix:bootstrap_ci}

To quantify uncertainty in the main retrieval results, we estimate 95\% confidence intervals for Recall@1 using 1,000 query-level bootstrap resamples. Tables~\ref{tab:bootstrap_ci_knowledge} and~\ref{tab:bootstrap_ci_reasoning} report the point estimates and percentile-based bootstrap intervals for eight representative text and multimodal embedding models on the knowledge and reasoning axes, respectively. The intervals are reported as R@1 [95\% CI], in percentage points.

\begin{sidewaystable}[p]
\centering
\caption{Recall@1 with 95\% bootstrap confidence intervals on the knowledge axis. Each entry is reported as R@1 [lower, upper] in percentage points.}
\label{tab:bootstrap_ci_knowledge}
\setlength{\tabcolsep}{3.2pt}
\renewcommand{\arraystretch}{1.4}
\scriptsize
\begin{tabular}{lcccc|cccc}
\toprule
\multirow{2}{*}{\textbf{Subtype}} & \multicolumn{4}{c|}{\textbf{Text Embedding Models}} & \multicolumn{4}{c}{\textbf{Multimodal Embedding Models}} \\
\cmidrule(lr){2-5}\cmidrule(lr){6-9}
& \makecell{\textbf{Diver-}\textbf{Embed-4B}} & \makecell{\textbf{Qwen3-}\textbf{Embed-4B}} & \makecell{\textbf{Qwen3-}\textbf{Embed-8B}} & \textbf{ReasonIR-8B} & \textbf{RzenEmbed-7B} & \makecell{\textbf{Ops-MM-}\textbf{Embed-v1-7B}} & \makecell{\textbf{Qwen3-VL-}\textbf{Embed-8B}} & \makecell{\textbf{Seed1.6-}\textbf{embedding}} \\
\midrule
Daily-Life       & 46.19 [39.83, 52.54] & 42.37 [36.02, 48.73] & 46.19 [39.83, 52.97] & 40.25 [33.90, 46.61] & 69.92 [64.41, 75.00] & 62.29 [56.36, 68.22] & 69.07 [63.56, 74.58] & 65.25 [59.32, 71.61] \\
Fine-Grained     & 1.09 [0.00, 3.26] & 2.17 [0.00, 5.43] & 1.09 [0.00, 4.35] & 1.09 [0.00, 3.26] & 2.17 [0.00, 6.52] & 1.09 [0.00, 3.26] & 0.00 [0.00, 0.00] & 4.35 [1.09, 8.70] \\
Spatial-Cogmap   & 0.00 [0.00, 0.00] & 3.70 [0.00, 8.64] & 0.00 [0.00, 0.00] & 0.00 [0.00, 0.00] & 1.23 [0.00, 3.70] & 1.23 [0.00, 3.70] & 0.00 [0.00, 0.00] & 0.00 [0.00, 0.00] \\
Spatial-3D       & 0.76 [0.00, 2.29] & 0.76 [0.00, 2.29] & 0.76 [0.00, 3.05] & 0.00 [0.00, 0.00] & 1.53 [0.00, 3.82] & 5.34 [1.53, 9.16] & 3.82 [0.76, 7.63] & 3.82 [0.76, 7.63] \\
\midrule
Biology          & 39.76 [28.92, 50.60] & 27.71 [18.07, 37.35] & 40.96 [31.33, 51.81] & 28.92 [18.07, 37.35] & 32.53 [22.89, 42.17] & 42.17 [31.33, 53.01] & 39.76 [28.92, 50.60] & 31.33 [21.69, 42.17] \\
Physics          & 22.50 [13.75, 31.25] & 21.25 [12.50, 31.25] & 22.50 [13.75, 32.50] & 20.00 [12.47, 28.75] & 10.00 [3.75, 17.50] & 15.00 [7.50, 21.25] & 26.25 [16.25, 36.25] & 18.75 [11.25, 27.50] \\
Chemistry        & 2.86 [0.00, 5.71] & 1.90 [0.00, 4.76] & 4.76 [0.95, 9.52] & 1.90 [0.00, 4.76] & 5.71 [1.90, 10.48] & 3.81 [0.95, 7.62] & 3.81 [0.95, 7.62] & 3.81 [0.95, 7.62] \\
\midrule
Mathematics      & 14.47 [8.55, 20.39] & 10.53 [5.92, 15.79] & 12.50 [7.24, 18.42] & 10.53 [5.92, 15.79] & 11.84 [7.22, 17.76] & 9.21 [4.61, 13.82] & 9.87 [5.92, 15.13] & 11.84 [6.58, 17.11] \\
Code-Drawing     & 2.00 [0.00, 5.00] & 3.00 [0.00, 7.00] & 4.00 [1.00, 8.00] & 1.00 [0.00, 3.00] & 7.00 [2.98, 12.00] & 6.00 [2.00, 11.00] & 14.00 [7.00, 21.00] & 18.00 [11.00, 26.00] \\
\midrule
Economics        & 55.00 [43.75, 65.00] & 36.25 [26.25, 47.50] & 47.50 [36.25, 60.00] & 42.50 [32.50, 53.75] & 43.75 [32.50, 55.00] & 41.25 [30.00, 52.50] & 51.25 [40.00, 62.50] & 45.00 [33.75, 56.25] \\
Comic            & 2.30 [0.00, 5.75] & 1.15 [0.00, 3.45] & 2.30 [0.00, 5.75] & 4.60 [1.15, 9.20] & 2.30 [0.00, 5.75] & 6.90 [2.30, 12.64] & 11.49 [5.75, 18.39] & 12.64 [5.75, 20.69] \\
Art              & 5.00 [1.00, 10.00] & 4.00 [1.00, 8.00] & 4.00 [1.00, 8.00] & 1.00 [0.00, 3.00] & 4.00 [1.00, 8.00] & 13.00 [7.00, 20.00] & 4.00 [1.00, 8.00] & 10.00 [4.00, 17.00] \\
Calendar         & 6.25 [1.25, 12.50] & 5.00 [1.25, 10.00] & 7.50 [2.50, 13.75] & 3.75 [0.00, 8.75] & 6.25 [1.25, 12.50] & 5.00 [1.25, 11.25] & 3.75 [0.00, 8.75] & 13.75 [7.50, 21.25] \\
Metro            & 10.00 [4.97, 17.50] & 3.75 [0.00, 8.75] & 2.50 [0.00, 6.25] & 13.75 [6.25, 21.28] & 7.50 [2.50, 13.75] & 17.50 [9.97, 26.25] & 7.50 [2.50, 13.75] & 16.25 [8.75, 25.00] \\
\midrule
Computer-Science & 46.25 [36.25, 57.50] & 33.75 [23.75, 43.75] & 46.25 [36.25, 57.53] & 36.25 [25.00, 46.25] & 48.75 [37.50, 58.75] & 42.50 [32.50, 53.75] & 46.25 [35.00, 57.50] & 57.50 [46.25, 67.50] \\
Mechanical       & 10.98 [4.88, 18.29] & 10.98 [4.88, 18.29] & 8.54 [3.66, 14.63] & 9.76 [3.66, 15.85] & 15.85 [8.54, 23.17] & 12.20 [4.88, 19.51] & 13.41 [6.10, 20.73] & 12.20 [6.10, 19.51] \\
Electronics      & 0.00 [0.00, 0.00] & 2.97 [0.00, 5.94] & 1.98 [0.00, 4.95] & 0.99 [0.00, 2.97] & 4.95 [0.99, 9.90] & 5.94 [1.98, 10.89] & 6.93 [1.98, 11.88] & 3.96 [0.99, 7.92] \\
\midrule
\textbf{Macro avg.} & \textbf{15.61 [14.09, 17.10]} & \textbf{12.43 [10.96, 13.84]} & \textbf{14.90 [13.48, 16.45]} & \textbf{12.72 [11.13, 14.18]} & \textbf{16.19 [14.79, 17.55]} & \textbf{17.08 [15.50, 18.72]} & \textbf{18.30 [16.66, 19.94]} & \textbf{19.32 [17.52, 21.00]} \\
\bottomrule
\end{tabular}
\end{sidewaystable}

\paragraph{Knowledge-axis reliability.}
The bootstrap results support the domain- and subtype-level findings underlying Observation~2. Across the evaluated models, performance remains substantially stronger on Daily-Life than on Fine-Grained, Spatial-Cogmap, and Spatial-3D, and is generally higher on Biology, Economics, and Computer Science than on Chemistry. Although confidence intervals are wider for some smaller subtypes, the large and consistently observed subtype-level gaps remain clear. Our conclusion therefore concerns recurring domain- and subtype-level variation across models.

\begin{sidewaystable}[p]
\centering
\caption{Recall@1 with 95\% bootstrap confidence intervals on the reasoning axis. Each entry is reported as R@1 [lower, upper] in percentage points.}
\label{tab:bootstrap_ci_reasoning}
\setlength{\tabcolsep}{3.2pt}
\renewcommand{\arraystretch}{1.4}
\scriptsize
\begin{tabular}{lcccc|cccc}
\toprule
\multirow{2}{*}{\textbf{Group}} & \multicolumn{4}{c|}{\textbf{Text Embedding Models}} & \multicolumn{4}{c}{\textbf{Multimodal Embedding Models}} \\
\cmidrule(lr){2-5}\cmidrule(lr){6-9}
& \makecell{\textbf{Diver-}\textbf{Embed-4B}} & \makecell{\textbf{Qwen3-}\textbf{Embed-4B}} & \makecell{\textbf{Qwen3-}\textbf{Embed-8B}} & \textbf{ReasonIR-8B} & \textbf{RzenEmbed-7B} & \makecell{\textbf{Ops-MM-}\textbf{Embed-v1-7B}} & \makecell{\textbf{Qwen3-VL-}\textbf{Embed-8B}} & \makecell{\textbf{Seed1.6-}\textbf{embedding}} \\
\midrule
Perception-Only & 46.19 [40.08, 52.89] & 41.53 [35.29, 47.93] & 45.76 [39.20, 51.99] & 40.25 [34.24, 47.01] & 70.34 [64.38, 75.89] & 63.56 [57.56, 69.49] & 68.64 [62.60, 74.03] & 62.29 [56.05, 68.02] \\
Knowledge-Intensive & 17.73 [15.71, 19.98] & 12.09 [10.27, 14.05] & 15.67 [13.69, 17.91] & 15.31 [13.36, 17.36] & 15.04 [13.02, 17.33] & 16.11 [14.07, 18.27] & 17.73 [15.72, 20.02] & 22.33 [20.05, 24.69] \\
Reasoning-Intensive & 11.94 [10.38, 13.72] & 7.73 [6.33, 9.21] & 9.57 [8.00, 11.20] & 11.18 [9.56, 12.85] & 10.57 [8.91, 12.29] & 12.02 [10.14, 13.80] & 11.94 [10.32, 13.59] & 16.85 [14.85, 18.90] \\
Both & 15.95 [13.77, 18.35] & 9.90 [8.11, 11.86] & 13.09 [11.03, 15.37] & 14.41 [12.27, 16.59] & 13.75 [11.56, 16.06] & 14.63 [12.27, 16.96] & 15.95 [13.69, 18.29] & 21.70 [19.08, 24.40] \\
\midrule
Knowledge Reasoning & 18.12 [15.54, 21.08] & 11.98 [9.72, 14.48] & 15.98 [13.49, 18.68] & 16.41 [13.74, 19.03] & 16.12 [13.56, 18.88] & 17.12 [14.31, 20.03] & 17.55 [14.91, 20.52] & 22.08 [19.17, 25.29] \\
Conceptual Abstraction & 27.07 [21.86, 32.87] & 17.90 [13.15, 23.08] & 23.58 [18.40, 29.32] & 28.38 [22.77, 33.93] & 22.71 [17.41, 28.76] & 23.58 [17.70, 29.24] & 30.13 [24.68, 36.41] & 33.33 [27.62, 38.97] \\
Fine-Grained Visual Reasoning & 3.85 [0.79, 7.14] & 1.54 [0.00, 4.03] & 0.77 [0.00, 2.52] & 3.85 [0.81, 7.09] & 4.62 [1.50, 8.57] & 6.92 [2.88, 11.30] & 0.77 [0.00, 2.56] & 4.35 [0.70, 8.67] \\
Logical Reasoning & 6.13 [2.65, 10.00] & 4.29 [1.31, 7.74] & 4.91 [1.95, 8.19] & 3.07 [0.62, 5.85] & 6.13 [2.86, 10.13] & 6.13 [2.56, 10.00] & 6.75 [3.31, 11.11] & 10.71 [7.05, 14.84] \\
Spatial Reasoning & 1.54 [0.37, 3.14] & 2.69 [0.81, 4.89] & 1.54 [0.37, 3.32] & 4.23 [1.79, 6.85] & 2.31 [0.75, 4.33] & 5.77 [2.98, 8.66] & 3.85 [1.76, 6.09] & 6.19 [3.60, 9.24] \\
Symbolic Reasoning & 9.31 [6.72, 12.12] & 5.19 [3.23, 7.43] & 6.71 [4.40, 9.21] & 7.58 [5.26, 9.96] & 8.01 [5.39, 10.55] & 6.93 [4.64, 9.23] & 8.87 [6.35, 11.51] & 15.16 [12.19, 18.54] \\
\midrule
\textbf{Average} & \textbf{11.00 [9.36, 12.67]} & \textbf{7.26 [5.89, 8.80]} & \textbf{8.92 [7.41, 10.54]} & \textbf{10.59 [8.89, 12.26]} & \textbf{9.98 [8.28, 11.80]} & \textbf{11.08 [9.23, 12.94]} & \textbf{11.32 [9.71, 13.01]} & \textbf{15.30 [13.38, 17.29]} \\
\bottomrule
\end{tabular}
\end{sidewaystable}

\paragraph{Reasoning-axis reliability.}
The reasoning-axis intervals reinforce the main diagnostic conclusions. For every evaluated model, Perception-Only retrieval is markedly stronger than Reasoning-Intensive retrieval. Across reasoning skills, Knowledge Reasoning and Conceptual Abstraction consistently outperform Fine-Grained Visual Reasoning and Spatial Reasoning, whose estimates remain low even after accounting for bootstrap uncertainty. This recurring pattern supports the conclusion that vision-centric reasoning is a persistent bottleneck rather than an artifact of a single model. These marginal bootstrap intervals quantify sampling variability within each group; they should not be interpreted as formal paired significance tests between models.

\clearpage
\section{Multi-round Retrieval Results for Agent}
\label{appendix:agent_multiround}

Recent agentic methods extend conventional one-shot retrieval by decomposing a user query into iterative search steps, inspecting intermediate results, and reformulating subsequent retrieval actions based on accumulated evidence~\cite{prism,a-rag,Agentic-R}.
This paradigm is particularly relevant to ARK, since many queries require multi-step reasoning and visual verification rather than simple semantic matching.

In this section, we evaluate two representative multi-round agent methods, A-RAG~\cite{a-rag} and Agentic-R~\cite{Agentic-R}.
As shown in Table~\ref{table:multi_round_retrieval}, these agentic retrievers do not outperform strong multimodal embedding models in per-round Recall@1, indicating that their single-step retrieval accuracy remains limited.
However, their cumulative Hits@1 improves consistently as the number of retrieval rounds increases, showing that iterative search can recover additional correct targets over multiple attempts.
These results suggest that current agentic retrievers are more promising as components in multi-stage retrieval pipelines than as direct replacements for strong multimodal retrievers.

\vspace{5pt}

\begin{table*}[bh]
\caption{{Multi-round retrieval performance.}
Text and multimodal embedding models are included as single-round baselines (reported with R@1 only). Agentic retrievers support multi-round retrieval, and we report both per-round R@1 and cumulative Hits@1 across rounds. Here, R@1 measures success in the current round only, while Hits@1 accumulates successful retrievals over all rounds. Agentic retrievers show consistent gains as the number of retrieval rounds increases.
}
\centering
\tablestyle{8pt}{1.02}
{
\begin{tabular}{l|cc|cc}
\toprule
\multirow{2}{*}{\textbf{Model}} & \multicolumn{2}{c|}{\textbf{Knowledge domain avg.}} & \multicolumn{2}{c}{\textbf{Reasoning skill avg.}} \\
\cmidrule(lr){2-3} \cmidrule(lr){4-5}
 & \textbf{R@1} & \textbf{Hits@1} & \textbf{R@1} & \textbf{Hits@1} \\
\midrule
\multicolumn{5}{c}{\textit{\textbf{Text Embedding Models}}} \\
\midrule
BGE-M3-0.6B & 10.71 & - & 7.27 & - \\
Diver-Embed-1.7B &  11.38 & - & 9.16 & - \\
Diver-Embed-4B & \textbf{15.61} & - & \textbf{11.00} & - \\
Qwen3-Embed-4B & 12.43 & - & 7.26 & - \\
\midrule
Qwen3-Embed-8B & \textbf{14.90} & - & 8.92 & - \\
ReasonIR-8B & 12.72 & - & \textbf{10.59}& - \\
\midrule
\multicolumn{5}{c}{\textit{\textbf{Multimodal Embedding Models}}} \\
\midrule
CLIP-0.4B & 7.66 & - & 3.52 & - \\
BGE-VL-0.4B & 7.12 & - & 2.15 & - \\
SigLIP2-0.9B & \textbf{10.68} & - &\textbf{4.33} & - \\
\midrule
LamRA-Ret-7B & 11.26 & - & 6.39 & - \\
RzenEmbed-7B & 16.19 & - & 9.98 & - \\
Ops-MM-Embed-v1-7B & 17.08 & - & 11.08 & - \\
Qwen3-VL-Embed-8B & \textbf{18.30} & - & \textbf{11.32} & - \\
\midrule
\multicolumn{5}{c}{\textit{\textbf{Agentic Retrieval Models}}} \\

\midrule
A-RAG-0.6B (round 1) & 12.64 & 12.64 & 8.37 & 8.37 \\
A-RAG-0.6B (round 3) & 13.53 & 21.42 & 9.18 & 14.65 \\
A-RAG-0.6B (round 5) & \textbf{14.39} & \textbf{27.66} & \textbf{9.74} & \textbf{20.48} \\
\midrule
A-RAG-8B (round 1)   & 13.09 & 13.09 & 9.96 & 9.96 \\
A-RAG-8B (round 3)   & 14.11 & 22.08 & 11.12 & 17.34 \\
A-RAG-8B (round 5)   & \textbf{14.87} & \textbf{28.42} & \textbf{11.90} & \textbf{22.83} \\
\midrule
Agentic-R-7B (round 1) & 12.81 & 12.81 & 8.73 & 8.73 \\
Agentic-R-7B (round 3) & 13.56 & 22.71 & 9.36 & 18.33 \\
Agentic-R-7B (round 5) & \textbf{14.11} & \textbf{26.52} & \textbf{9.67} & \textbf{21.79} \\
\bottomrule
\end{tabular}
}
\label{table:multi_round_retrieval}
\end{table*}

\clearpage
\section{Captioning Robustness and Prompt Details}
\label{appendix:prompts}

We first evaluate the robustness of text-only retrieval baselines to different image-captioning models. We then provide the prompts used for caption generation and query rewriting for inference-time enhancement in Tables~\ref{prompt:caption_generation} and~\ref{prompt:query_rewrite}, respectively.

\subsection{Robustness to Captioning Models}
\label{appendix:captioner_robustness}

To examine whether using Qwen-generated image captions systematically favors Qwen-family text embedding models, we reran all six text-only embedding models on the full benchmark using captions generated by two non-Qwen models: \href{https://huggingface.co/moonshotai/Kimi-VL-A3B-Instruct}{Kimi-VL-A3B-Instruct} (16B parameters, 3B active) and \href{https://huggingface.co/zai-org/GLM-4.6V-Flash}{GLM-4.6V-Flash} (9B parameters).

Table~\ref{tab:captioner_knowledge_robustness} reports the macro-average Recall@1 on the knowledge axis. Model rankings remain highly stable across captioners: Diver-Embed-4B performs best under all three settings, and the Spearman rank correlations are $\rho=1.00$ between Qwen and Kimi and $\rho=0.94$ between Qwen and GLM. Moreover, Qwen3-Embed-4B performs better with GLM captions than with Qwen captions (13.30 vs. 12.43).

\begin{table*}[ht]
\centering
\caption{Macro-average Recall@1 (\%) on the knowledge axis using captions generated by different models. Differences are measured relative to Qwen captions.}
\label{tab:captioner_knowledge_robustness}
\tablestyle{6pt}{1.08}{
\begin{tabular}{lccccc}
\toprule
\textbf{Text embedding model} & \textbf{Qwen} & \textbf{Kimi} & \textbf{GLM} & \textbf{Kimi$-$Qwen} & \textbf{GLM$-$Qwen} \\
\midrule
BGE-M3-0.6B      & 10.71 & 8.97  & 10.08 & $-1.74$ & $-0.63$ \\
Diver-Embed-1.7B & 11.38 & 9.70  & 10.40 & $-1.68$ & $-0.98$ \\
Diver-Embed-4B   & \textbf{15.61} & \textbf{12.97} & \textbf{14.77} & $-2.64$ & $-0.84$ \\
Qwen3-Embed-4B   & 12.43 & 11.05 & 13.30 & $-1.38$ & $+0.87$ \\
Qwen3-Embed-8B   & 14.90 & 11.95 & 13.67 & $-2.95$ & $-1.23$ \\
ReasonIR-8B      & 12.72 & 11.09 & 12.37 & $-1.63$ & $-0.35$ \\
\midrule
\textbf{Model average} & 12.96 & 10.96 & 12.43 & $-2.00$ & $-0.53$ \\
\bottomrule
\end{tabular}}
\end{table*}

To directly assess family-specific compatibility, we define the Qwen-caption gain for each embedding model as its performance with Qwen captions minus its average performance with the two non-Qwen captioners. On the knowledge axis, the mean gain is 1.17 percentage points for the two Qwen embedding models, compared with 1.31 points for the four non-Qwen models.

Table~\ref{tab:captioner_reasoning_robustness} reports the macro-average Recall@1 across the six reasoning skills. These fine-grained results are more sensitive to caption quality: the rank correlations decrease to approximately $\rho=0.60$ for Qwen versus Kimi and $\rho=0.49$ for Qwen versus GLM. Nevertheless, Diver-Embed-4B remains the strongest model with all three captioners. The Qwen-caption gain is 0.16 points for Qwen embedding models and 2.30 points for non-Qwen embedding models, indicating that the relative benefit of Qwen captions is not concentrated among Qwen-family retrievers.

\begin{table*}[ht]
\centering
\caption{Macro-average Recall@1 (\%) across the six reasoning skills using captions generated by different models.}
\label{tab:captioner_reasoning_robustness}
\tablestyle{9pt}{1.08}{
\begin{tabular}{lccc}
\toprule
\textbf{Text embedding model} & \textbf{Qwen} & \textbf{Kimi} & \textbf{GLM} \\
\midrule
BGE-M3-0.6B      & 7.27  & 5.42 & 6.12 \\
Diver-Embed-1.7B & 9.16  & 5.29 & 5.64 \\
Diver-Embed-4B   & \textbf{11.00} & \textbf{9.29} & \textbf{9.54} \\
Qwen3-Embed-4B   & 7.26  & 7.18 & 8.35 \\
Qwen3-Embed-8B   & 8.92  & 7.63 & 8.58 \\
ReasonIR-8B      & 10.59 & 7.93 & 8.44 \\
\midrule
\textbf{Model average} & 9.03 & 7.12 & 7.78 \\
\bottomrule
\end{tabular}}
\end{table*}

Overall, these results provide no evidence of a systematic Qwen-family compatibility advantage. The knowledge-axis conclusions and overall model rankings remain largely stable across captioners, whereas the fine-grained reasoning-skill results are more sensitive to caption quality. This contrast further highlights the value of separating knowledge and reasoning in ARK.

\begin{table*}[ht]
\centering
\newcommand{\VarSty}[1]{\textnormal{\ttfamily\color{blue!90!black}#1}\unskip}
\newcommand{\var}{\texttt}

\begin{minipage}{0.99\columnwidth}\vspace{0mm}    \centering
\caption{The prompt for image caption generation.}
\begin{tcolorbox}
    \centering
    \small
    \hspace{-6mm}
    \begin{tabular}{p{0.99\columnwidth}}

\begin{minipage}{0.99\columnwidth}\vspace{0mm}

\VarSty{prompt} = f\var{"""} \\

Please generate a caption for this image, covering the scene context, the main objects, their relationships, actions, and attributes. The caption should be approximately 100 words and focus only on the visible content without adding speculation.\\

\var{"""}
\end{minipage}

    \end{tabular}
\end{tcolorbox}

\vspace{-2mm}
\label{prompt:caption_generation}
\end{minipage}
\end{table*}

\begin{table*}[h]\centering
\newcommand{\VarSty}[1]{\textnormal{\ttfamily\color{blue!90!black}#1}\unskip}
\newcommand{\var}{\texttt}

\begin{minipage}{0.99\columnwidth}\vspace{0mm} \centering
\caption{The prompt for query rewriting.}
\begin{tcolorbox}
    \centering
    \small
    \hspace{-6mm}
    \begin{tabular}{p{0.99\columnwidth}}
\begin{minipage}{0.99\columnwidth}\vspace{0mm}

\textbf{Input Format:}\\
\quad - Original Query: \textless query\_text\textgreater\\
\quad - Original Question Images: \textless query\_image\textgreater\\[2pt]

\VarSty{prompt} = f\var{"""}\\

Task Description:\\
You are an expert in multimodal retrieval. Your goal is to rewrite the query to bridge the gap between a user's raw query and the complex visual target by explicitly externalizing implicit domain knowledge and reasoning. Perform the following execution flow to generate the refined query.\\[2pt]

Execution Flow:\\
1. \textbf{Dual-Axis Analysis}: Identify the required knowledge domain and reasoning type.\\
2. \textbf{Step-by-Step Reasoning}:

\quad - Infer implicit background knowledge or domain concepts not explicitly stated in the text.

\quad - Translate the query's abstract reasoning logic into concrete, observable visual cues. Determine what specific objects, attributes, scene configurations, or structural patterns must be visibly present to support the query's intent.

\quad - Identify critical constraints that separate the ideal target from hard negatives.\\
3. \textbf{Synthesize Target Image}: Based on the reasoning above, describe the visual content and layout of the ideal target image.\\
4. \textbf{Query Rewrite}: Integrate the above information into a single, concise text search query. (\textasciitilde100 words)\\[2pt]

Output Requirement:\\
Only output the final refined query.\\

\var{"""}
\end{minipage}
    \end{tabular}
\end{tcolorbox}

\vspace{-2mm}
\label{prompt:query_rewrite}
\end{minipage}
\end{table*}

\clearpage

\section{Impact Statement}
\label{appendix:impact_statement}

ARK is designed to support the development of more reliable multimodal retrieval.
It is constructed exclusively from publicly available materials and is intended for non-commercial academic research.
Following standard dataset curation practices~\cite{mr2,mmmu}, annotators prioritize sources with explicit reuse permissions and refrain from using materials from websites that prohibit copying or redistribution.
We also conduct best-effort screening to remove private or sensitive information and to filter out harmful content, aiming to facilitate socially beneficial research.
Accordingly, ARK is primarily constructed from openly reusable resources, complemented by selected existing datasets released under compatible public licenses.
ARK has been released under the CC BY-NC-SA 4.0 license to promote transparent and responsible downstream use.
Overall, we do not anticipate significant ethical concerns.

{

\section{Annotation Protocol and Consistency Analysis for Reasoning-Related Labels}\label{appendix:annotation_protocol}

In this section, we provide additional details on both the data construction workflow and the instance-level annotation protocol.
Figure~\ref{fig:annotation workflow} summarizes the overall curation pipeline, including data collection, reasoning-label annotation, and arbitration.
Table~\ref{tab:data_collection_protocol} further specifies the data collection and curation rules used to construct ARK.
The original data content and target pairs are constructed from reliable sources as described in Appendix~\ref{appendix:dataset_construction_details} and the subsequent annotation protocol focuses on assigning and validating labels such as cognitive demand and reasoning skill.
We use a structured human annotation pipeline. Annotators first receive unified training with shared annotation guidelines, then annotate instances independently, and finally resolve all disagreement cases through panel adjudication.

\textbf{Instance-level annotation protocol.}
We provide the instance-level annotation guidelines and the corresponding agreement statistics for reasoning-related annotation in Tables~\ref{tab:reasoning_demand}--\ref{tab:handling_ambiguities}. The annotation pipeline and user interface are shown in Figures~\ref{fig:annotation workflow}--\ref{fig:anno_panel_arbitration}.

\textbf{Annotation process.}
The reasoning-related annotation was conducted by 6 PhD annotators with diverse academic backgrounds. Annotators assign reasoning-related labels based on the reasoning pattern required by each retrieval instance, including its query, target, and hard negatives.
During annotation, annotators check whether the rewritten query is consistent with the target sample and whether the assigned reasoning-related labels reflect the intended reasoning requirement. Gemini 3 Pro is used as an auxiliary tool for difficult cases. In all cases, the final annotation decision is made by human annotators, not by the model.

\textbf{Agreement and adjudication.}
The corresponding agreement statistics, including per-annotator accuracy and inter-annotator agreement such as Cohen's kappa, are reported in Table~\ref{tab:annotation_consistance}. Across annotators, Cohen's kappa values are consistently in the substantial-to-almost-perfect range (all $>0.61$), indicating that the reasoning-related labels are generally reliable rather than heavily affected by annotation ambiguity. Remaining disagreement cases are resolved through panel adjudication.

\clearpage

\begin{center}
    \begin{protocolbox}{Annotation Protocol: Annotation Workflow}
    \centering
    \includegraphics[width=0.8\textwidth]{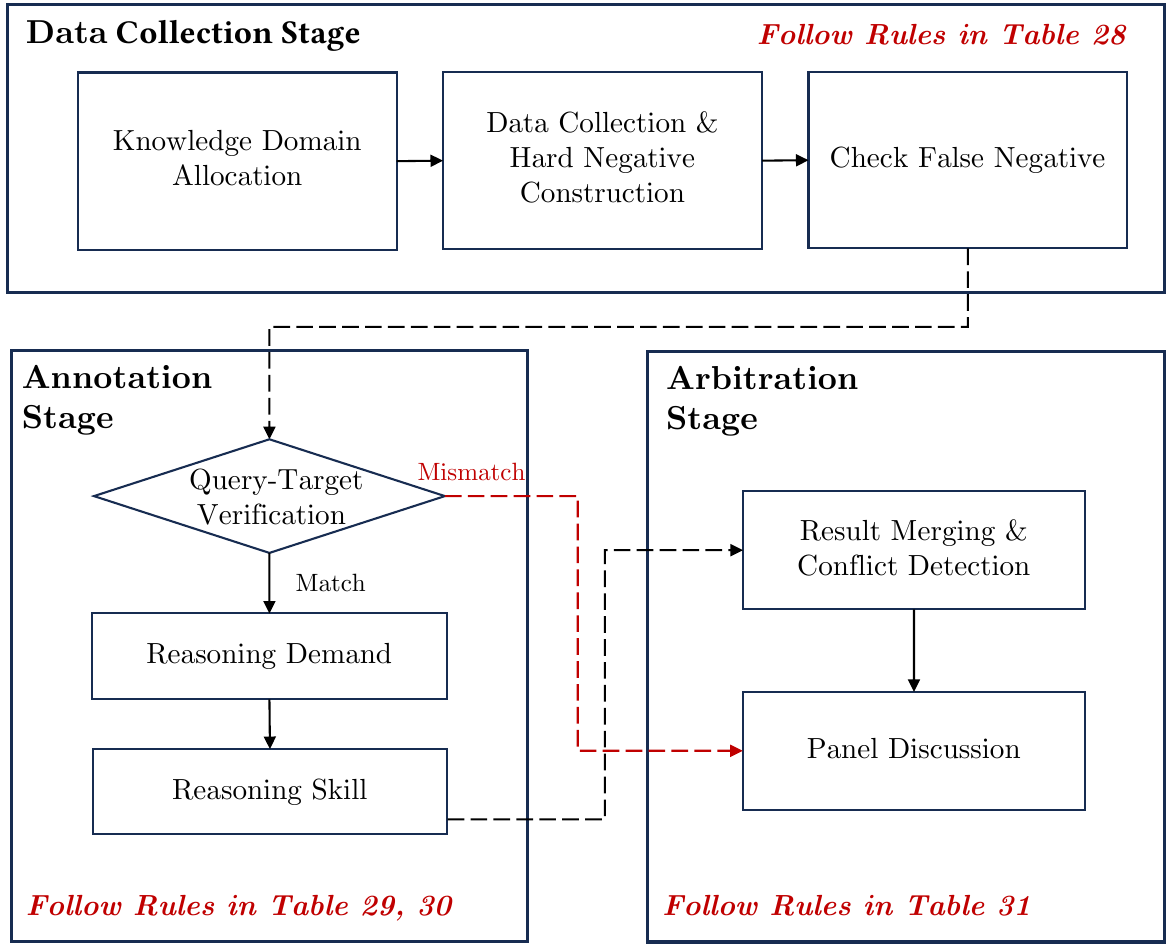}
    \end{protocolbox}
    \captionof{figure}{{The three-stage construction workflow of the ARK benchmark.} The pipeline begins with data collection and targeted hard negative construction. During the Annotation Stage, instances first undergo a strict Query-Target Verification step where mismatched or erroneous pairs are immediately flagged and routed to a Panel Discussion.
    Valid matches then proceed through systematic instance-level labeling (Reasoning Demand and Skill). Finally, the Arbitration Stage merges independent results to identify annotation conflicts. Any flagged instances undergo a Panel Discussion to resolve discrepancies and finalize the ground-truth labels. Each phase strictly adheres to predefined protocol rules (detailed in Tables~\ref{tab:data_collection_protocol}--\ref{tab:handling_ambiguities}) to ensure maximum dataset consistency and validity.}
\label{fig:annotation workflow}
\end{center}

\clearpage
\begin{center}
    \captionof{table}{The comprehensive data collection and curation protocol for the ARK benchmark.}
    \label{tab:data_collection_protocol}
    \begin{protocolbox}{Annotation Protocol: Data Collection}
    The principles for collecting and curating the multimodal query-target pairs for the ARK benchmark is given as:

\vspace{4pt}

    \textbf{1.1 Source of Data}

    \vspace{4pt}

    \begin{itemize}
        \item \textbf{Authoritativeness:} Annotators must collect instances from authoritative open-source platforms (e.g., Wikipedia, World Bank Open Data, arXiv) or expert-level academic datasets.
        \item \textbf{Diversity:} Annotators should prioritize finding diverse sources instead of collecting query-target pairs from a single domain to prevent structural bias.
        \item \textbf{Domains:} To comprehensively evaluate multimodal retrieval, annotators must ensure the collected data align with our 5 predefined meta-domains.
    \end{itemize}

    \vspace{6pt}

    \textbf{1.2 Ethical Considerations}
    \begin{itemize}
        \item \textbf{Copyright:} Annotators should prioritize collecting materials with explicit open-source reuse permissions, such as the CC BY-NC-SA 4.0 license. Data from sources that explicitly prohibit copying, scraping, or redistribution \textbf{MUST} be strictly avoided.
        \item \textbf{Privacy and Safety:} Annotators must conduct a best-effort manual screening during the collection phase to identify and permanently filter out any images or texts containing personally identifiable information, sensitive content, or harmful materials to ensure the dataset remains safe for socially beneficial research.
    \end{itemize}
    \textbf{1.3 Hard Negatives}
    \begin{itemize}
        \item \textbf{High Surface-Level Similarity:} A valid hard negative must share significant semantic or visual overlap with the ground truth target.
        \item \textbf{Strict Logical Violation:} The negative \textbf{MUST} definitively violate at least one core constraint specified in the query. (e.g., a slightly swapped coordinate in a cognitive map)
        \item \textbf{No Ambiguity:} Annotators must verify that the negative is unambiguously incorrect based {only} on the provided textual and visual evidence. If a distractor could be arguably correct under a different interpretation, it must be discarded.
    \end{itemize}
    \end{protocolbox}

\end{center}

\clearpage
\begin{center}
\captionof{table}{Protocol for labeling specific reasoning demand in ARK dataset.}
\label{tab:reasoning_demand}
\begin{protocolbox}{Annotation Protocol: Decision Rules for Reasoning Demand}
\textbf{1. Knowledge-Intensive}

\vspace{4pt}

\begin{itemize}
    \item \textbf{Keyword:} Expert-level, domain-specific concepts
    \item \textbf{Decision Rules:} Assign this label if distinguishing the target from hard negatives primarily depends on recognizing \textbf{expert-level, domain-specific concepts rather than general, everyday knowledge}. The matching process relies heavily on internal factual recall rather than multi-step logical deduction.
    \item \textbf{Examples:} Retrieving an image of an ``Amur leopard''. Distinguishing it from hard negatives (like jaguars, cheetahs, or other leopard subspecies) requires expert-level biological knowledge of specific coat patterns and taxonomy, rather than basic ``animal'' recognition.
\end{itemize}

\vspace{4pt}

\textbf{2. Reasoning-Intensive}
\begin{itemize}
    \item \textbf{Keyword:} Explicitly stated characteristics; common sense; complex reasoning
    \item \textbf{Decision Rules:} Assign this label if the required domain knowledge is everyday common sense or if the target's characteristics are explicitly provided in the query. The core bottleneck is not recalling facts, but \textbf{applying multi-step inference, relational understanding, spatial alignment, or structural abstraction} to select the correct target over confusing distractors.
    \item \textbf{Examples:}
    A query such as: ``Find the desk where the notebook is between the mug and the lamp, and the number of red pens is greater than the number of blue pens.'' Here, the model does not need specialized domain knowledge. Instead, it must integrate multiple relational and counting constraints to distinguish the target from highly similar distractors.
\end{itemize}

\textbf{3. Both}
\begin{itemize}
    \item \textbf{Keyword:} Two-step cognitive process; expert-level knowledge; complex reasoning
    \item \textbf{Decision Rules:} Assign this label if the task strictly requires a \textbf{two-step cognitive process}: the model must first recall the correct domain rules or specialized concepts mentioned in the query, and then apply those rules through a deductive, spatial, or symbolic process to distinguish the ground truth from highly plausible hard negatives.
    \item \textbf{Examples:} Mathematical proofs, physics phenomena, or chemistry reactions where the query involves symbolic formulas or specialized terms.
\end{itemize}

\textbf{4. None}
\begin{itemize}
    \item \textbf{Keyword:} Perception; common sense
    \item \textbf{Decision Rules:} Assign this label if the task requires neither specialized domain knowledge nor complex reasoning. The retrieval can be easily solved via \textbf{basic surface-level semantic matching}. It relies entirely on everyday common sense, general object recognition, and basic vocabulary, with no deceptive hard negatives that require multi-step verification.
    \item \textbf{Examples:} In a daily-life image retrieval scenario, the query is simply ``a man riding a bicycle in a park.'' The model relies entirely on general common sense to align the basic text concepts with the visual elements. There are no challenging distractors that require domain expertise or structural deduction to filter out.
\end{itemize}
\end{protocolbox}
\end{center}

\clearpage
\begin{center}
    \captionof{table}{Protocol for labeling specific reasoning skills in ARK dataset.}
    \label{tab:reasoning_skills}
    \begin{protocolbox}{Annotation Protocol: Decision Rules for Reasoning Skills}
    \textbf{1. Knowledge Reasoning}

    \vspace{4pt}
    \begin{itemize}
        \item \textbf{Decision Rules:} Assign this label if the retrieval requires inference grounded in background domain knowledge, \textbf{going explicitly beyond pure recognition or factual recall}. This skill frequently co-occurs with other reasoning types. The model must apply domain rules, facts, or historical contexts to deduce the correct target, rather than simply identifying a named entity.
        \item \textbf{Example:} {In a chemistry retrieval scenario}, the query asks for the immediate product of a specific chemical reaction. This inherently involves \textbf{both} {Knowledge Reasoning} (recalling and applying specific chemical transformation principles) and {Symbolic Reasoning} (deducing the resulting structural changes).
    \end{itemize}

 \vspace{4pt}

    \textbf{2. Spatial Reasoning}
    \begin{itemize}
        \item \textbf{Decision Rules:} Assign this label if the task strictly requires understanding geometric transformations, topological relations, or precise relative positioning. The bottleneck \textbf{must involve mentally manipulating spatial structures, such as 3D folding/rotation}.
        \item \textbf{Examples:} {In a 3D geometry scenario}, the query provides a 3D block structure and a viewing direction arrow. The model must mentally perform a 3D-to-2D projection (Spatial Reasoning) to distinguish the correct 2D silhouette.
    \end{itemize}

    \textbf{3. Fine-Grained Visual Reasoning}
    \begin{itemize}
        \item \textbf{Decision Rules:} Assign this label if the target needs to be identified by subtle, localized, or detail-sensitive visual cues (e.g., tiny objects, specific textures, or small text) within high-resolution images. The model must perform a meticulous ``\textbf{thinking with images}'' inspection process to verify these localized conditions.
        \item \textbf{Examples:} {In a high-resolution natural landscape scenario}, the query asks for a specific ``German nature park'' featuring a small, localized informational sign. The model cannot rely on the overall forest appearance; it must zoom in to verify the tiny text on the signboard (Fine-Grained Visual Reasoning) to reject hard negatives showing similar ruins or parks.
    \end{itemize}

    \textbf{4. Logical Reasoning}
    \begin{itemize}
        \item \textbf{Decision Rules:} Assign this label if correct retrieval depends on \textbf{structured deduction over explicit conditions, relations, or Boolean logic} (e.g., conjunction, disjunction, negation, implication).
        \item \textbf{Examples:} {In a digital circuit scenario}, the query describes a logical expression \texttt{(A xor B) nand (C xor D)}. The model must trace the logic gates (Logical Reasoning) to select the correct circuit diagram over hard negatives that have identical visual layouts but swap an XOR gate for an OR gate.
    \end{itemize}

    \textbf{5. Symbolic Reasoning}
    \begin{itemize}
        \item \textbf{Decision Rules:} Assign this label if the task involves interpreting \textbf{structured symbolic representations} (e.g., mathematical formulas, chemical equations, or code snippets). This typically requires mapping structured symbolic representations to diagrams or graphical plots.
        \item \textbf{Examples}: In a code-driven graphics scenario, the query provides a raw TikZ code snippet specifying a geometric shape. The model must interpret the symbolic coordinate syntax and programmatic commands to map them to the correct rendered drawing.
    \end{itemize}
    \end{protocolbox}
\end{center}

\begin{center}
    \addtocounter{table}{-1}
    \captionof{table}{Protocol for labeling specific reasoning skills in ARK dataset. Continued.}
    \begin{protocolbox}{Annotation Protocol: Decision Rules for Reasoning Skills Continued}
    \textbf{6. Conceptual Abstraction}
    \begin{itemize}
        \item \textbf{Decision Rules:} Assign this label if the task requires higher-level inductive generalization, interpreting metaphors, or summarizing abstract takeaways from visual trends. The model must \textbf{move from concrete visual elements} (like chart lines or cartoon characters) \textbf{to abstract concepts} (like economic shifts or historical viewpoints).
        \item \textbf{Examples:} {In an economics scenario}, the query asks for a chart demonstrating how ``near-term net payoff is governed chiefly by the match between skills and demand.'' The model must abstract the high-level economic conclusion from numerical trends.
    \end{itemize}
    \end{protocolbox}
\end{center}

\begin{center}
    \captionof{table}{Conflict resolution pipeline and guidelines for handling annotation ambiguities.}
    \label{tab:handling_ambiguities}
    \begin{protocolbox}{Annotation Protocol: Handling Ambiguities}
    To ensure the high validity and consistency of the ARK benchmark, we adopt the following conflict resolution pipeline for any disagreements:

 \vspace{3pt}

    \begin{itemize}
        \item \textbf{Merging:} Once all annotators have completed their batches, any instance with conflicting labels is automatically flagged and exported to a centralized merging tool for arbitration by the core authors.
        \item \textbf{Arbitration:} For each disputed instance, the core authors review the conflicting labels alongside the raw data to reach a consensus based on the established decision rules.
        \item \textbf{Discarding:} If the core authors determine during the discussion that an instance is inherently ambiguous, the query and target are fundamentally mismatched, or the hard negative cannot be definitively eliminated without subjective assumptions---the instance is immediately discarded from the benchmark. \textbf{We prioritize annotation certainty over dataset size.}
    \end{itemize}
    \end{protocolbox}
\end{center}

\begin{figure}[ht]
    \centering
    \includegraphics[width=\linewidth]{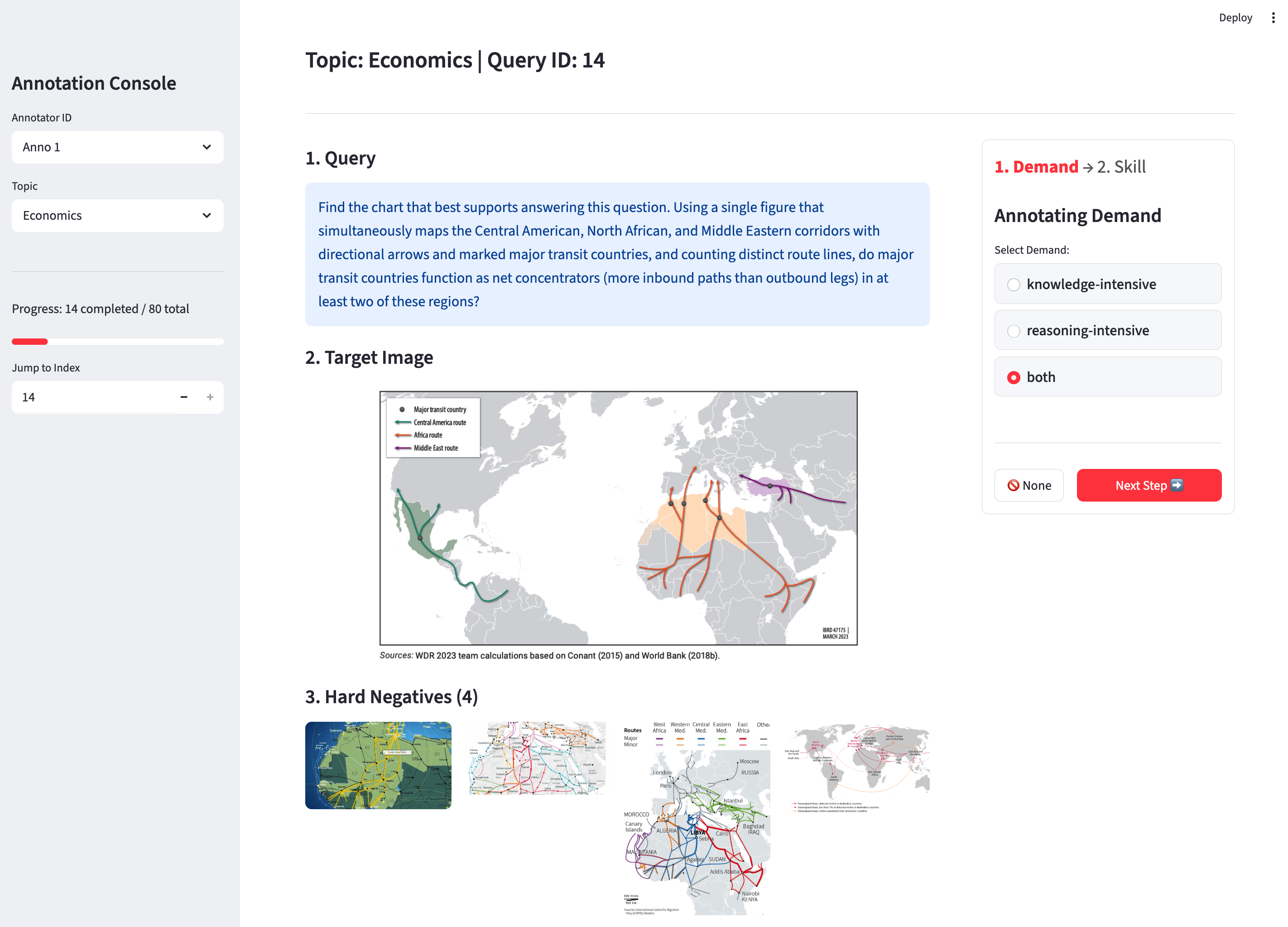}
    \caption{\textbf{Annotation Interface: Reasoning Demand}. The interface displays the text query, target image, and corresponding hard negatives. Annotators select the appropriate demand label on the right and click ``Next Step'' to proceed. Instances labeled as ``Reasoning-Intensive'' or ``Both'' proceed to the Reasoning Skill annotation stage, while instances labeled as ``Knowledge-Intensive'' or ``None'' immediately advance to the next sample without requiring reasoning-skill annotation. The left panel provides topic navigation and tracks overall annotation progress.}
    \label{fig:anno_panel_demand}
\end{figure}
\vspace{-5pt}
\begin{figure}[htbp]
    \centering
    \includegraphics[width=\linewidth]{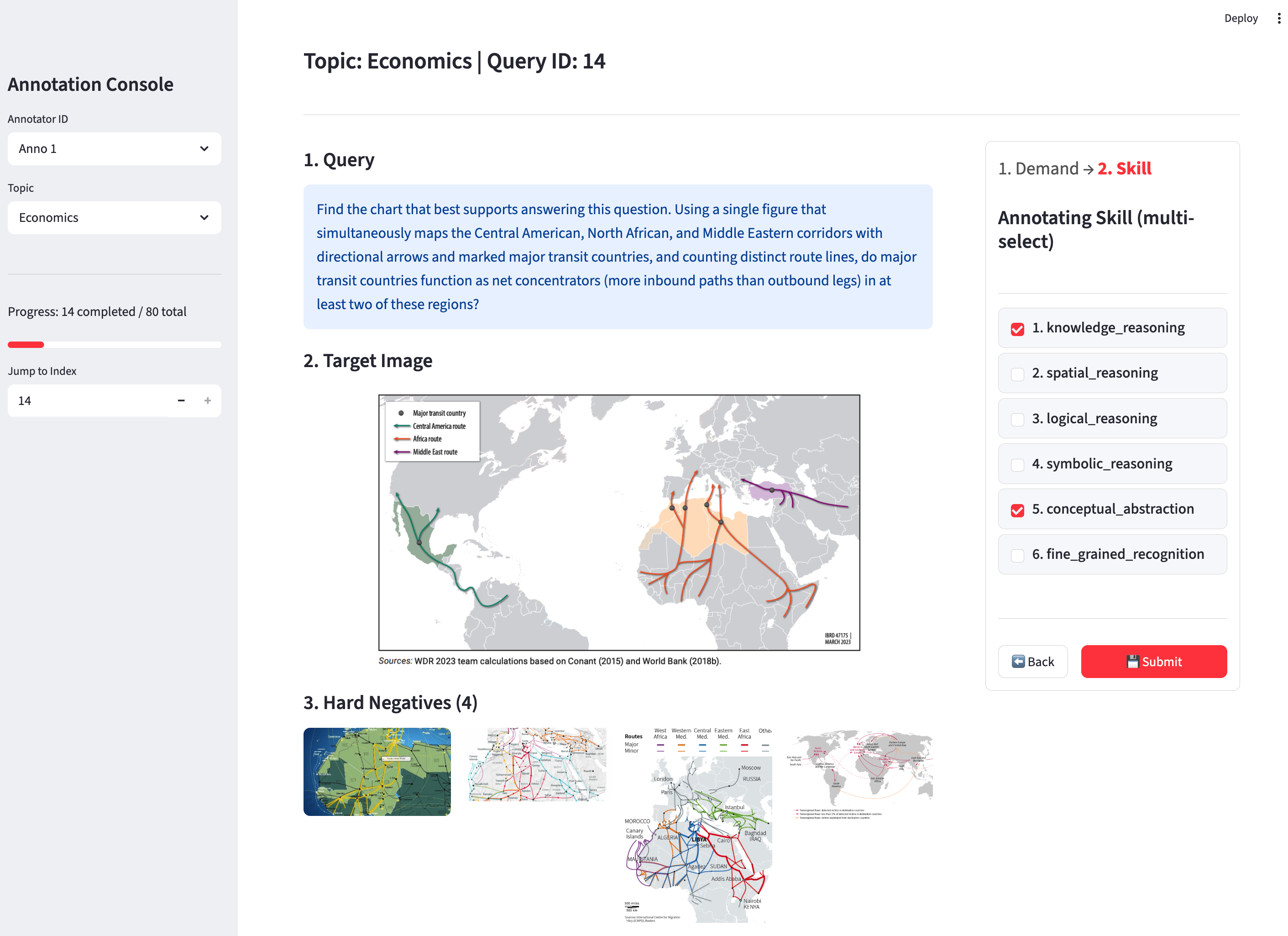}
    \caption{\textbf{Annotation Interface: Reasoning Skill.} The interface displays the text query, target image, and corresponding hard negatives. On the right panel, the annotators must select at least one applicable reasoning skill to activate the ``Next Step'' button and proceed to the Primary Evidence annotations.}
    \label{fig:anno_panel_skill}
\end{figure}

\begin{figure}[htbp]
    \centering
    \includegraphics[width=\linewidth]{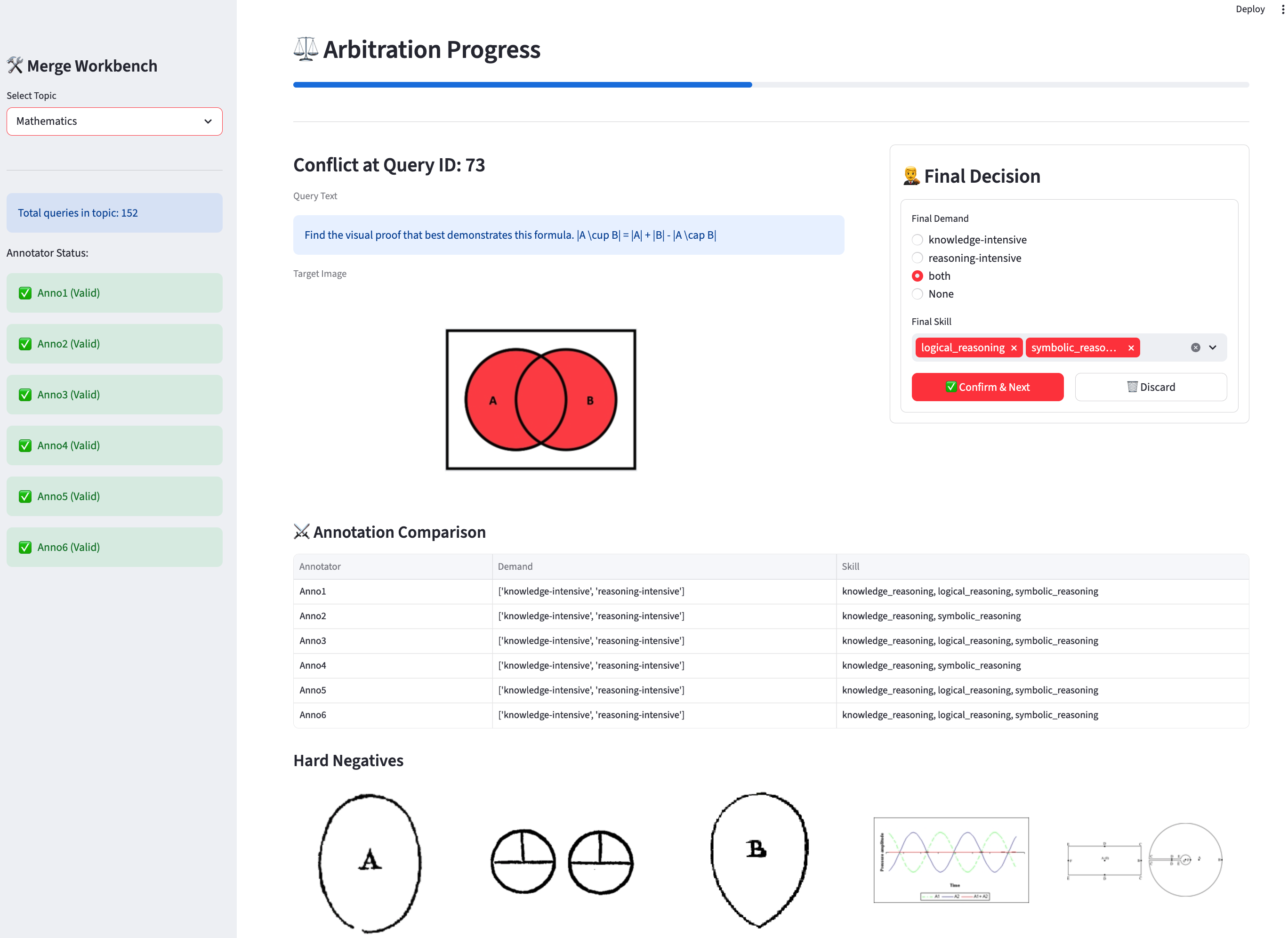}
    \caption{\textbf{Arbitration Interface}. It displays the text query, target image, hard negatives, and the conflicting label records from different annotators. Authors use this interface to finalize the ground-truth labels based on established guidelines. If an instance is inherently ambiguous or mismatched, the panel clicks ``Discard'' to eliminate the sample and ensure dataset validity.}
    \label{fig:anno_panel_arbitration}
\end{figure}

\begin{figure}[htbp]
    \centering
    \captionof{table}{\textbf{Annotation consistency analysis}. The table reports the accuracy, Cohen's kappa coefficients, and qualitative agreement levels of six annotators across two distinct tasks: Reasoning Demand (4-class) and Reasoning Skill (6-class). Here, Cohen's kappa measures the level of agreement between the annotators' assigned labels and the final consensus labels. Qualitative agreement levels follow standard interpretation guidelines (Poor: $<0.00$, Slight: $0.00$-$0.20$, Fair: $0.21$-$0.40$, Moderate: $0.41$-$0.60$, Substantial: $0.61$-$0.80$, Almost Perfect: $0.81$-$1.00$). The leftmost column lists the annotators and their academic backgrounds: CS (Computer Science), BE (Biomedical Engineering), and ME (Mechanical Engineering). Confidence intervals for the kappa coefficients are provided in parentheses.}
    \includegraphics[width=\linewidth]{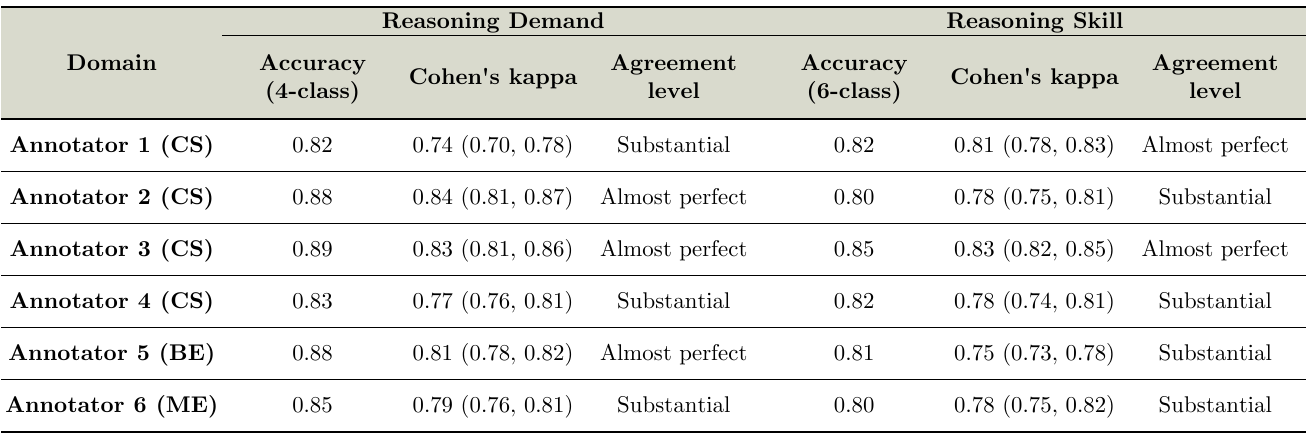}
    \label{tab:annotation_consistance}
\end{figure}
}

\clearpage

\section{Case Study}
\label{appendix:case_study}

\begin{figure*}[!th]
\centering
    \includegraphics[width=0.6\linewidth]{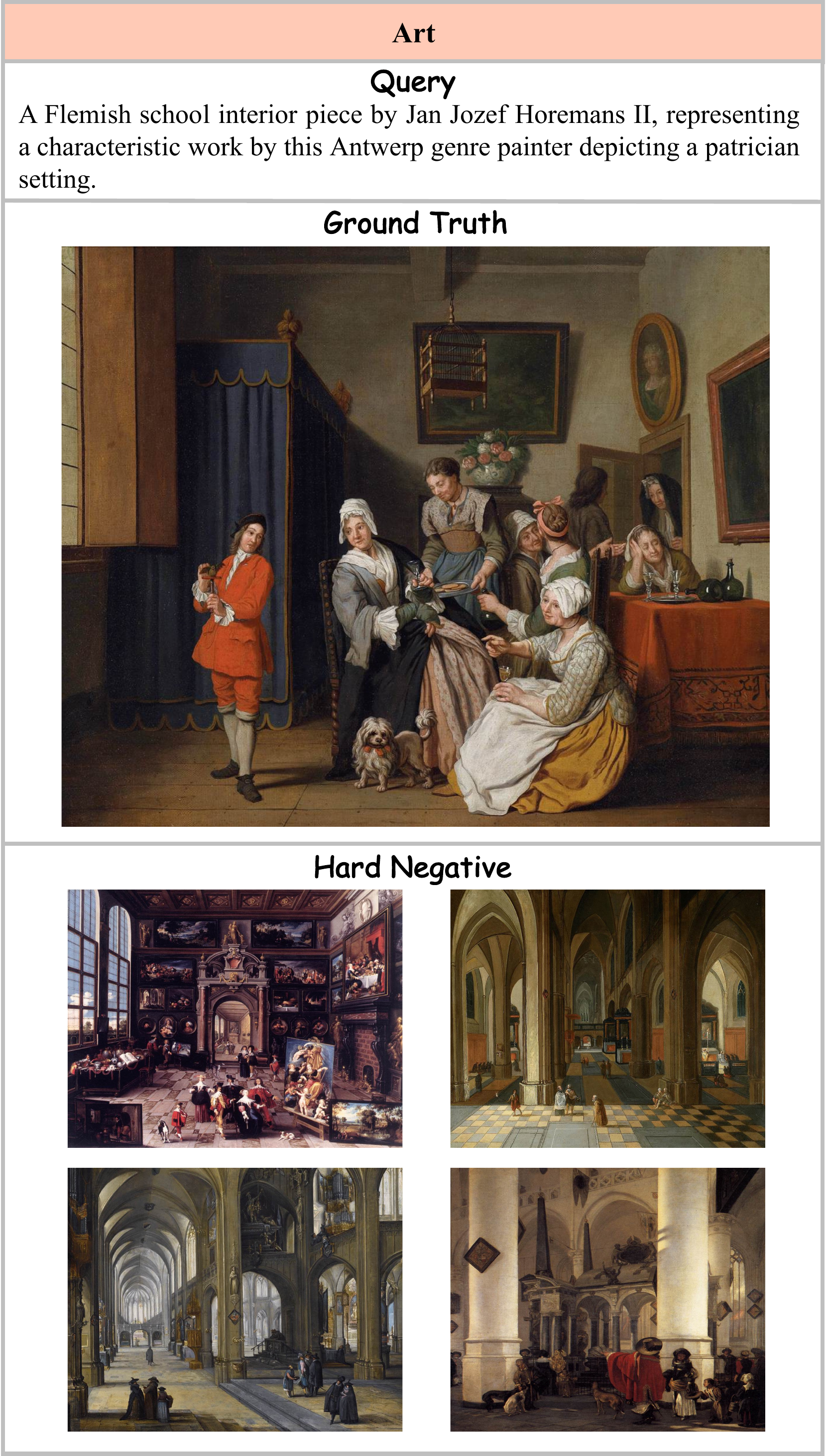}
  \caption{Visualized example of Art subtype.}
  \label{fig: case_Art}
\end{figure*}

\begin{figure*}[!th]
\centering
    \includegraphics[width=0.6\linewidth]{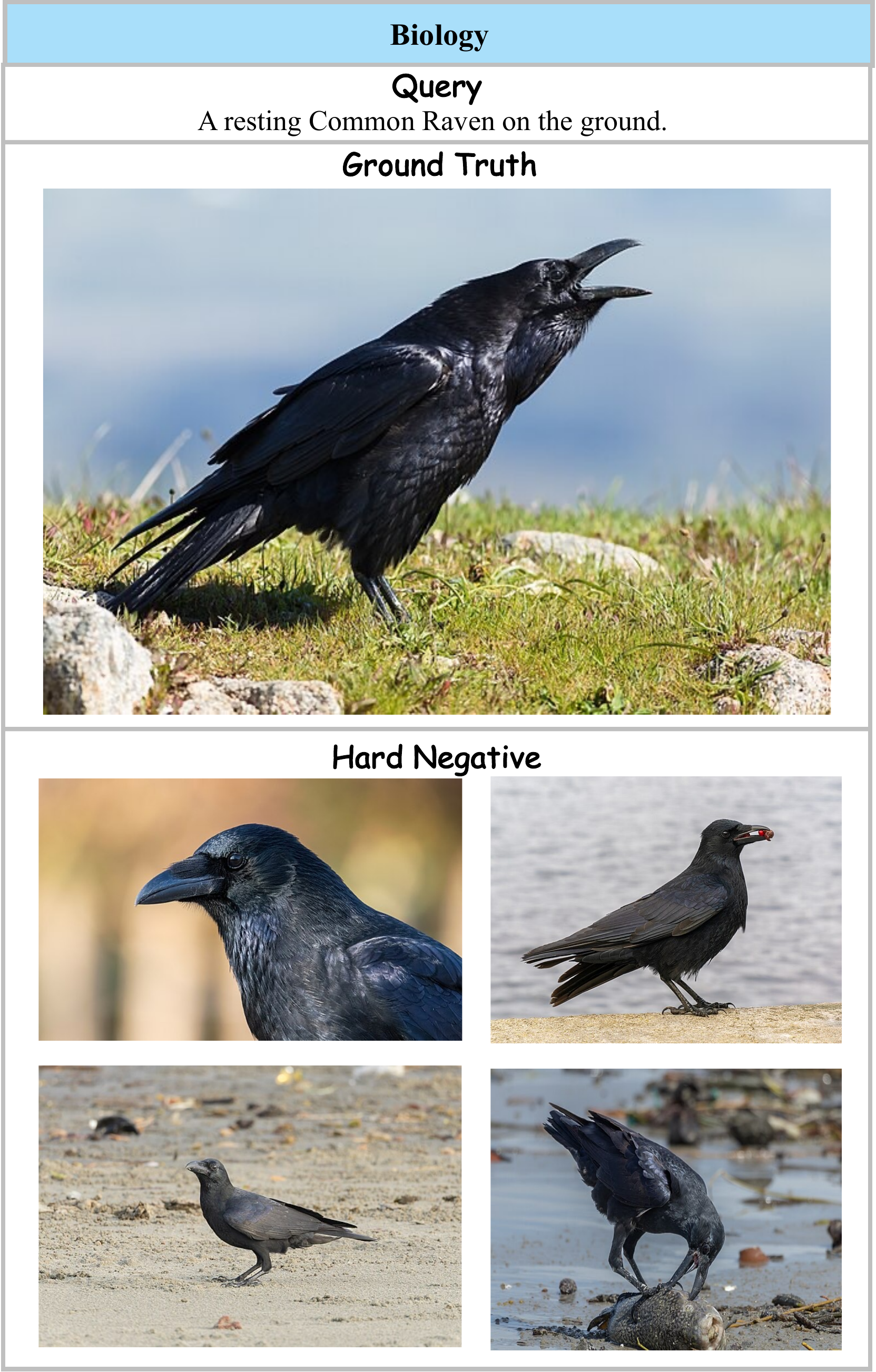}
  \caption{Visualized example of Biology subtype.}
  \label{fig: case_Biology}
\end{figure*}

\begin{figure*}[!th]
\centering
    \includegraphics[width=0.6\linewidth]{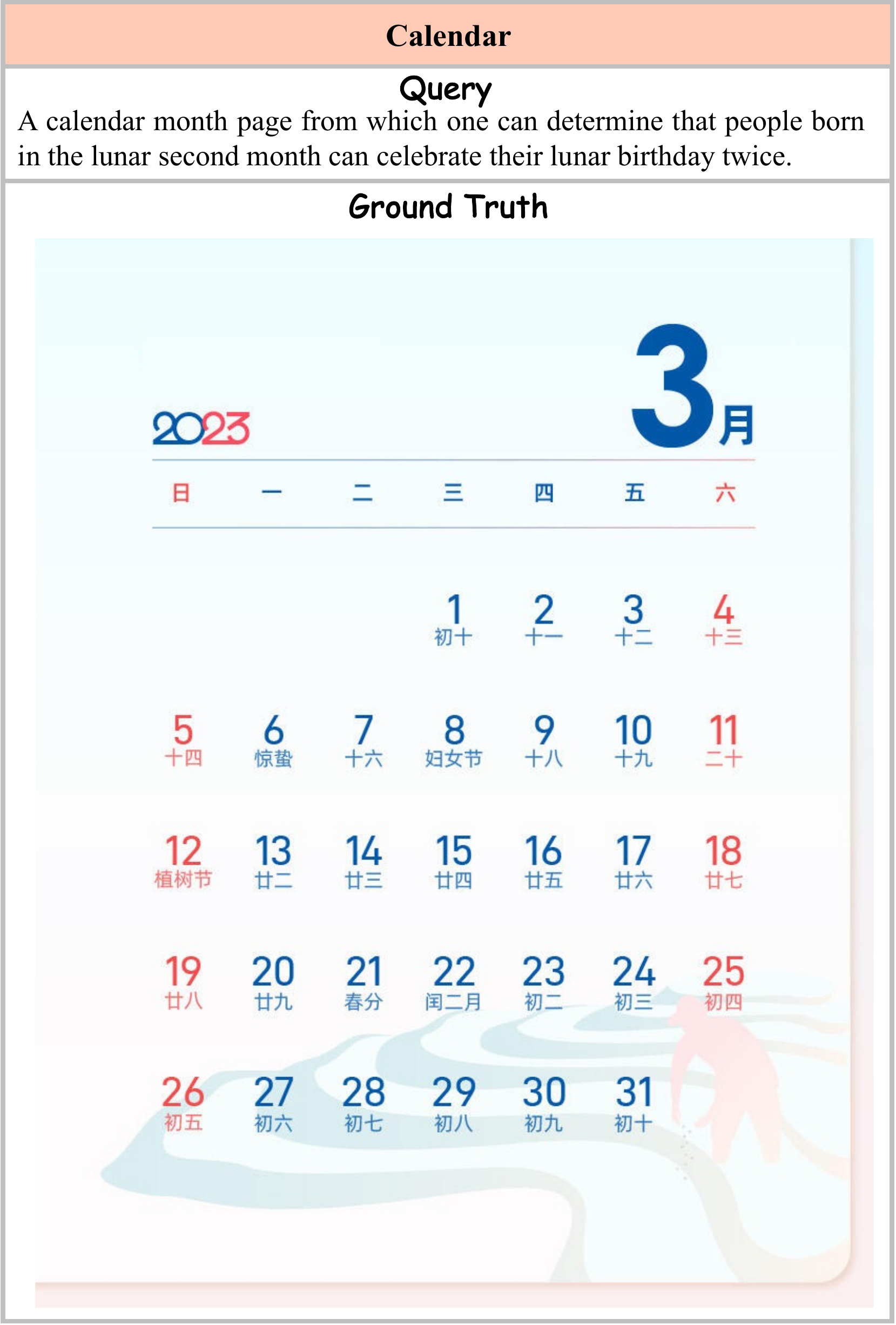}
  \caption{Visualized example of Calendar subtype.}
  \label{fig: case_Calendar}
\end{figure*}

\begin{figure*}[!th]
\centering
    \includegraphics[width=0.6\linewidth]{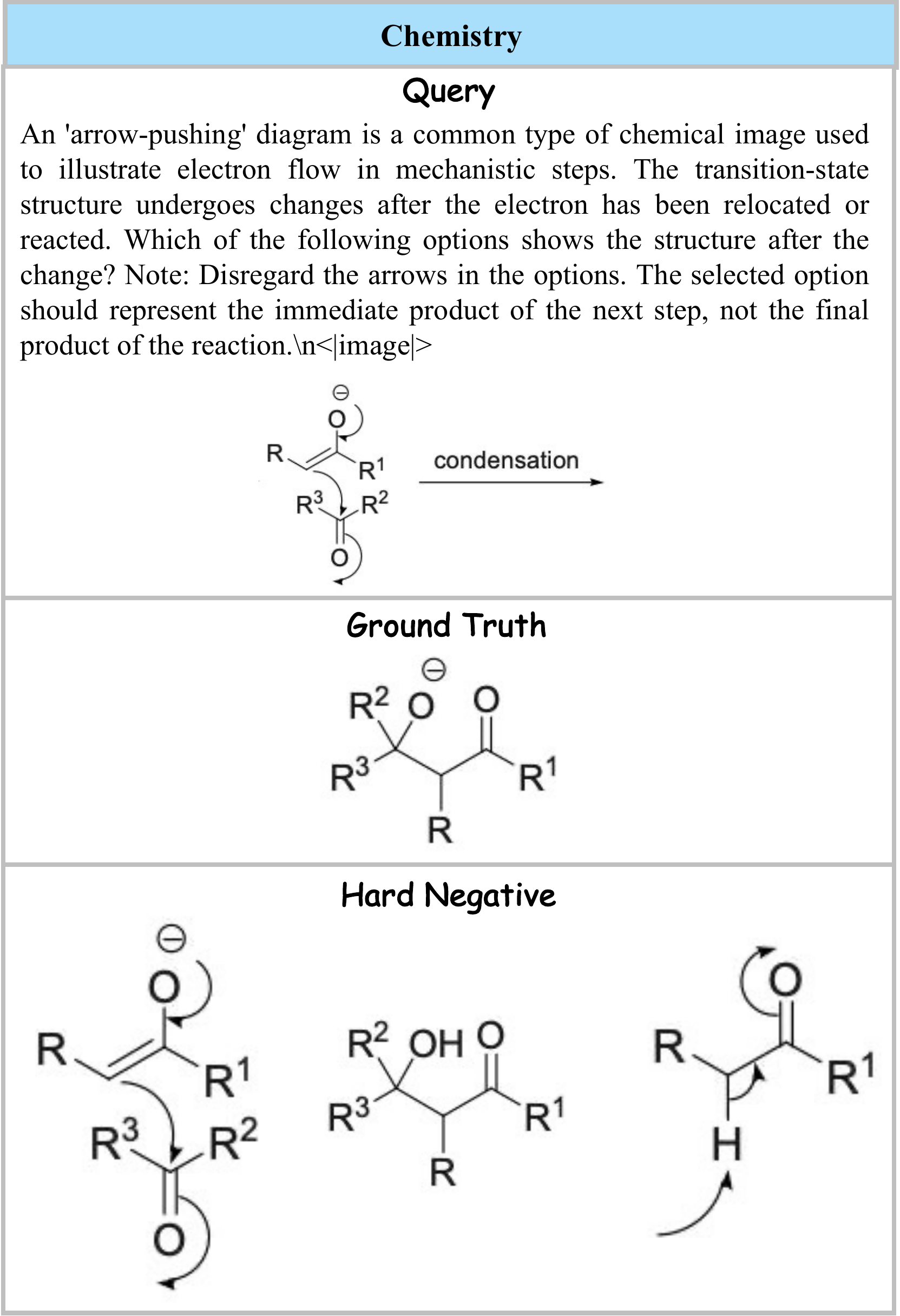}
  \caption{Visualized example of Chemistry subtype.}
  \label{fig: case_chemistry}
\end{figure*}

\begin{figure*}[!th]
\centering
    \includegraphics[width=0.6\linewidth]{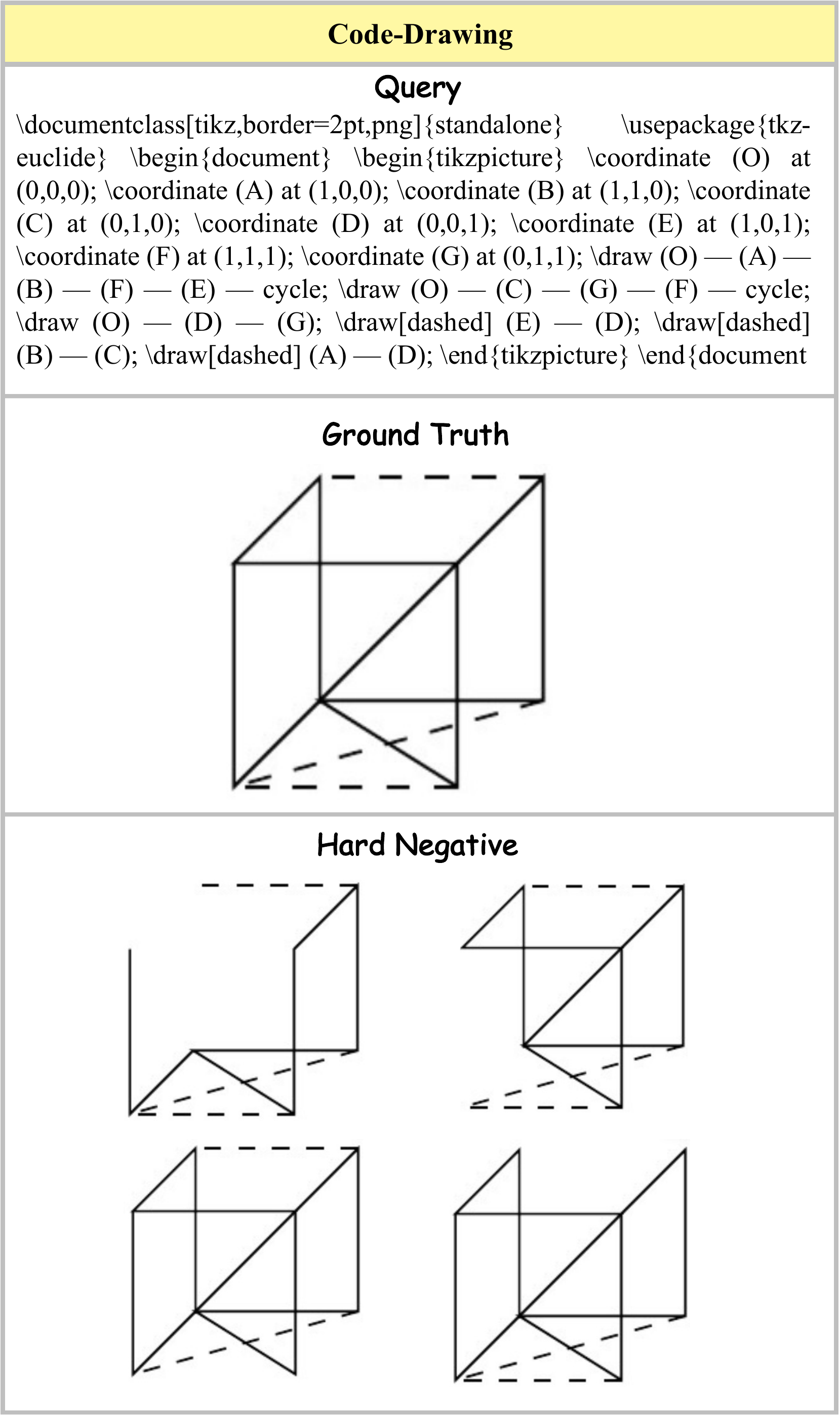}
  \caption{Visualized example of Code-Drawing subtype.}
  \label{fig: case_code_drawing}
\end{figure*}

\begin{figure*}[!th]
\centering
    \includegraphics[width=0.6\linewidth]{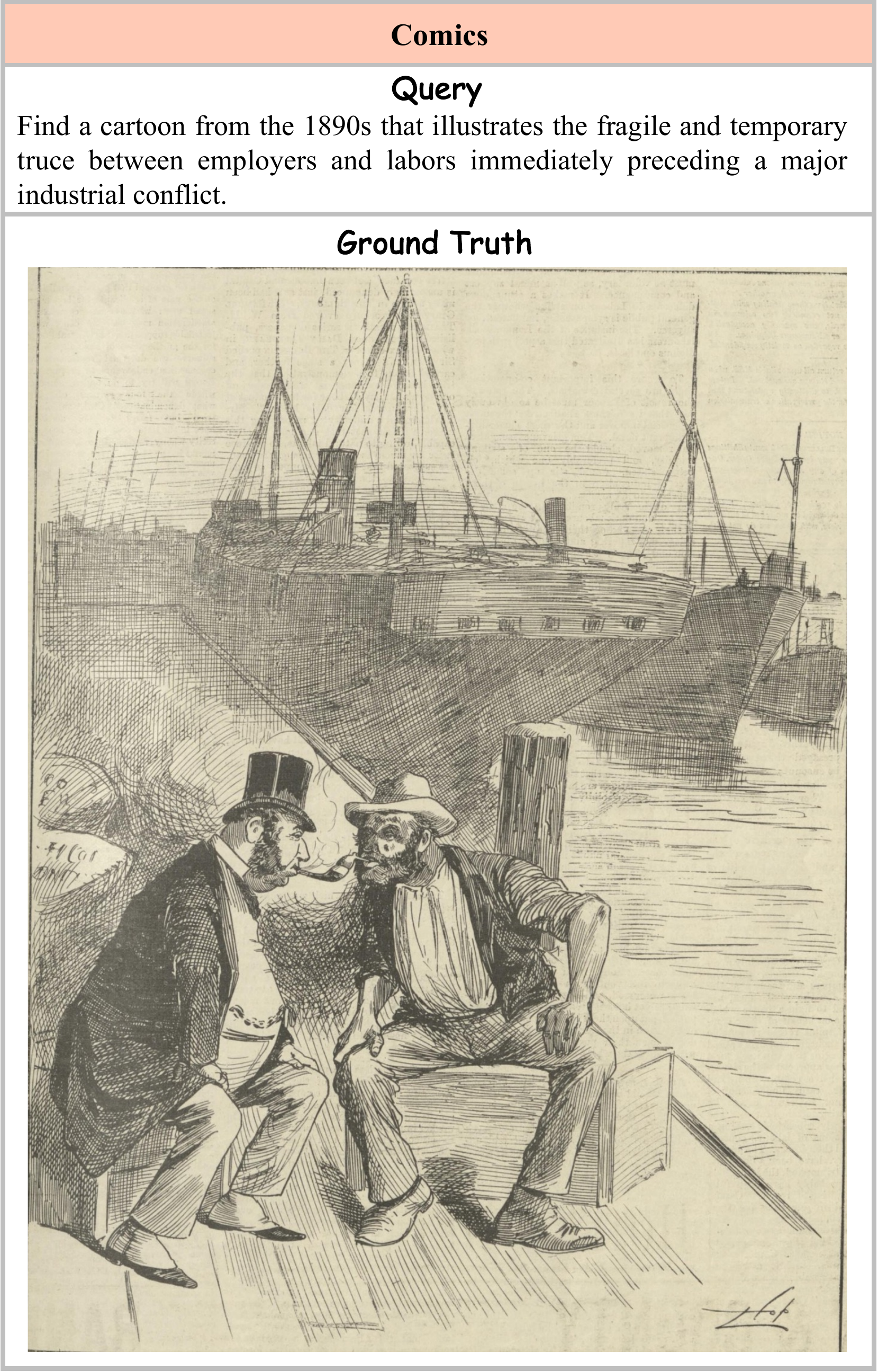}
  \caption{Visualized example of Comic subtype.}
  \label{fig: case_Comic}
\end{figure*}

\begin{figure*}[!th]
\centering
    \includegraphics[width=0.6\linewidth]{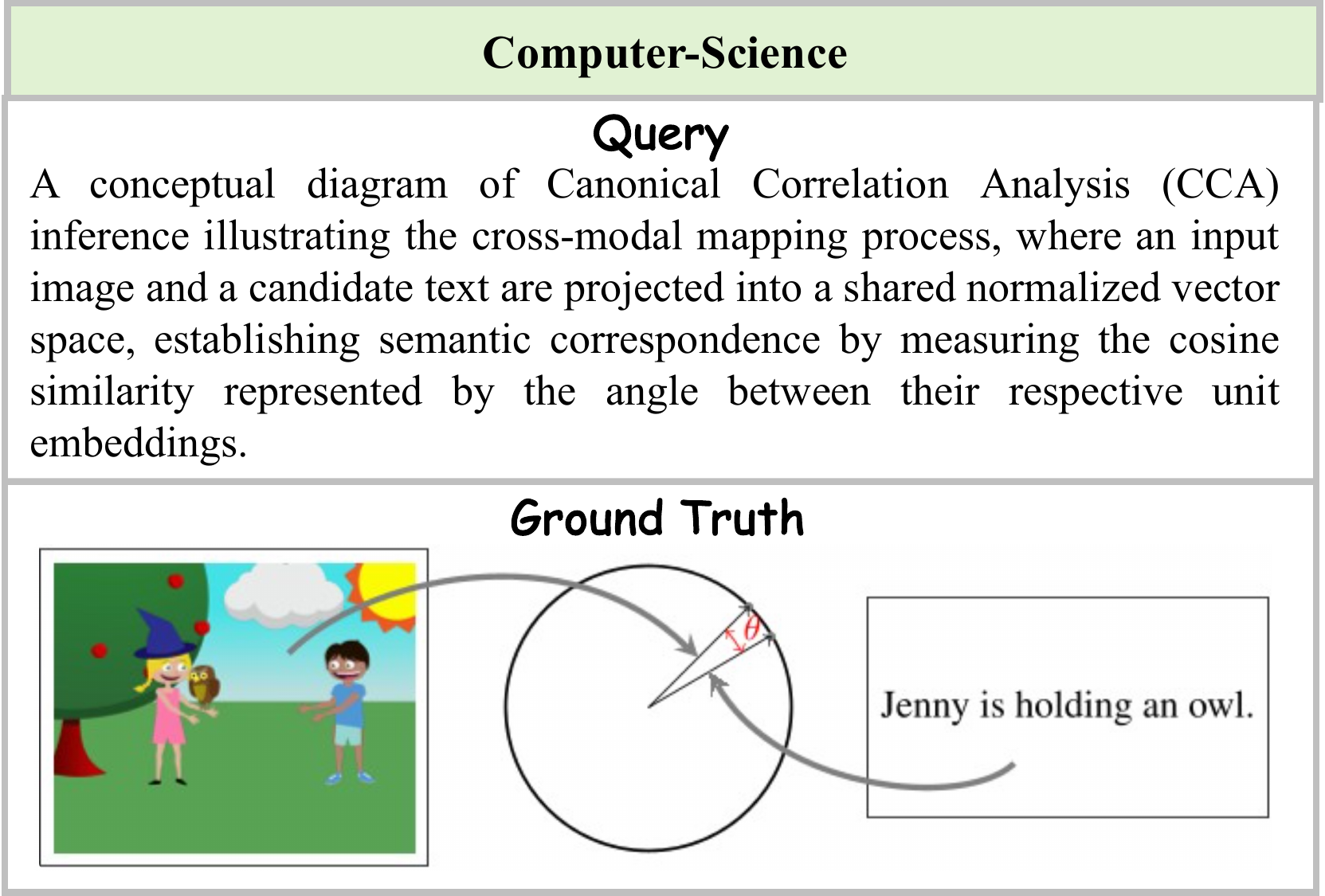}
  \caption{Visualized example of Computer-Science subtype.}
  \label{fig: case_Computer_Science}
\end{figure*}

\begin{figure*}[!th]
\centering
    \includegraphics[width=0.6\linewidth]{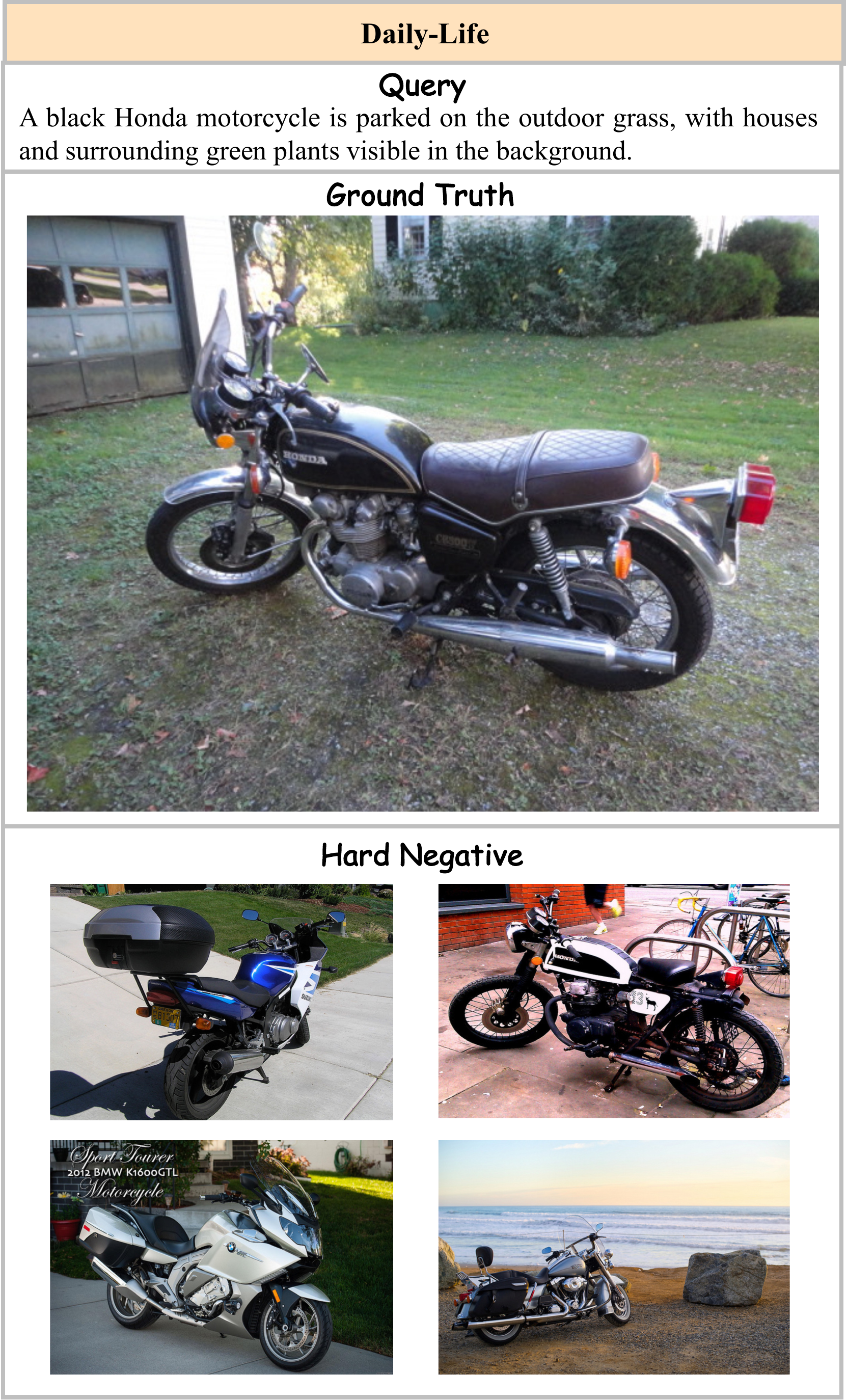}
  \caption{Visualized example of Daily-Life subtype.}
  \label{fig: case_Daily_Life}
\end{figure*}

\begin{figure*}[!th]
\centering
    \includegraphics[width=0.6\linewidth]{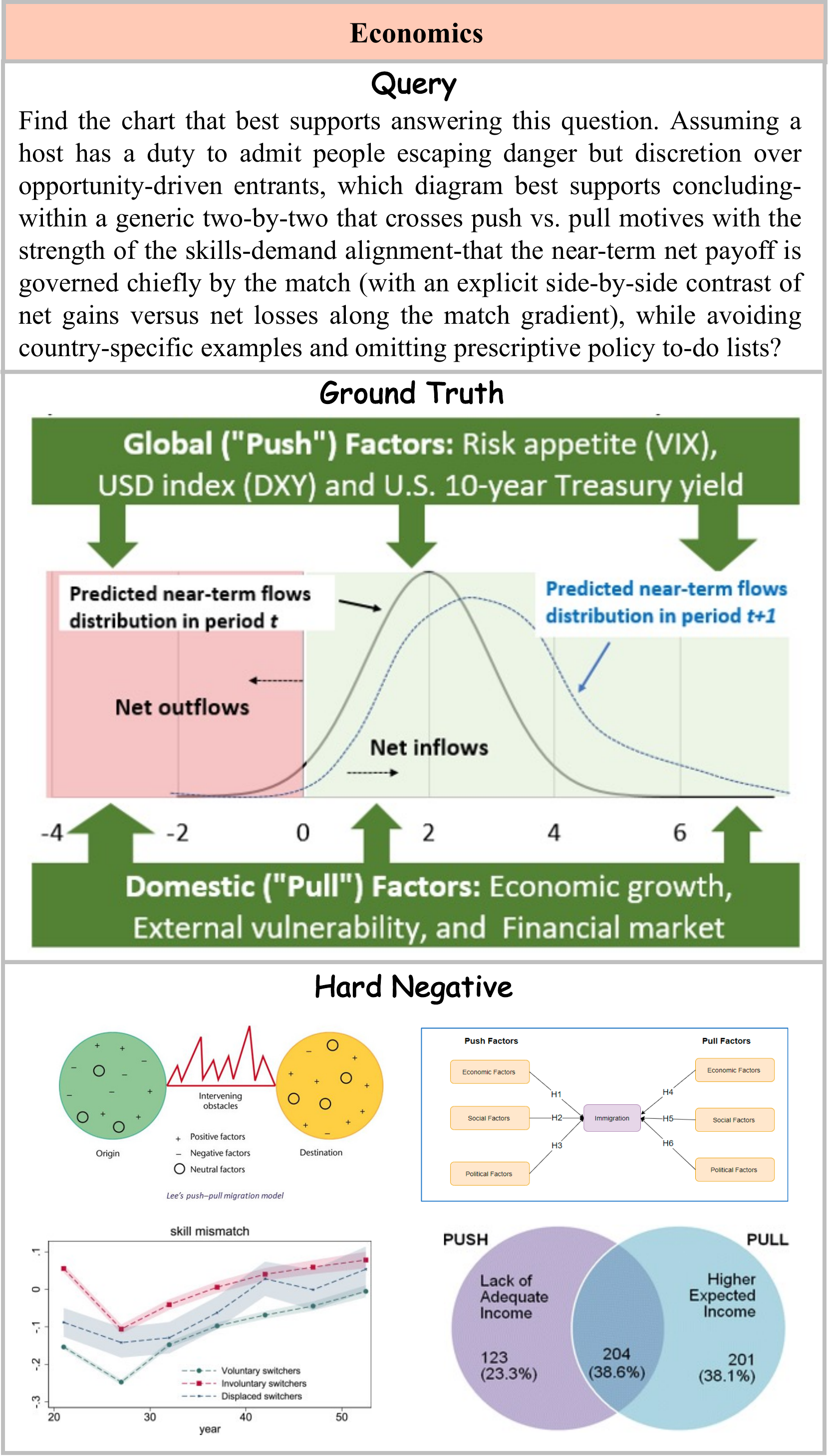}
  \caption{Visualized example of Economics subtype.}
  \label{fig: case_Economics}
\end{figure*}

\begin{figure*}[!th]
\centering
    \includegraphics[width=0.6\linewidth]{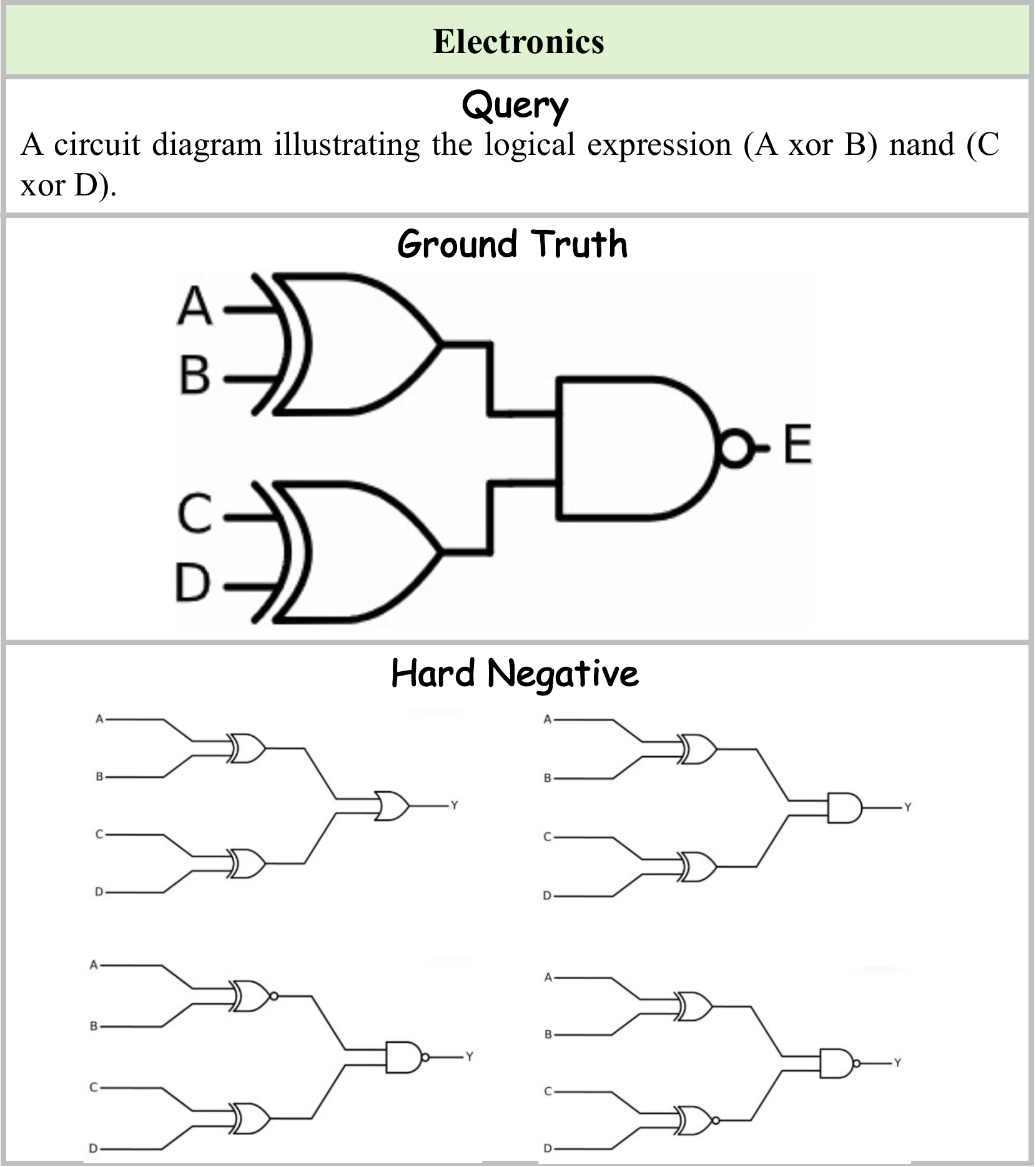}
  \caption{Visualized example of Electronics subtype.}
  \label{fig: case_Electronics}
\end{figure*}

\begin{figure*}[!th]
\centering
    \includegraphics[width=0.6\linewidth]{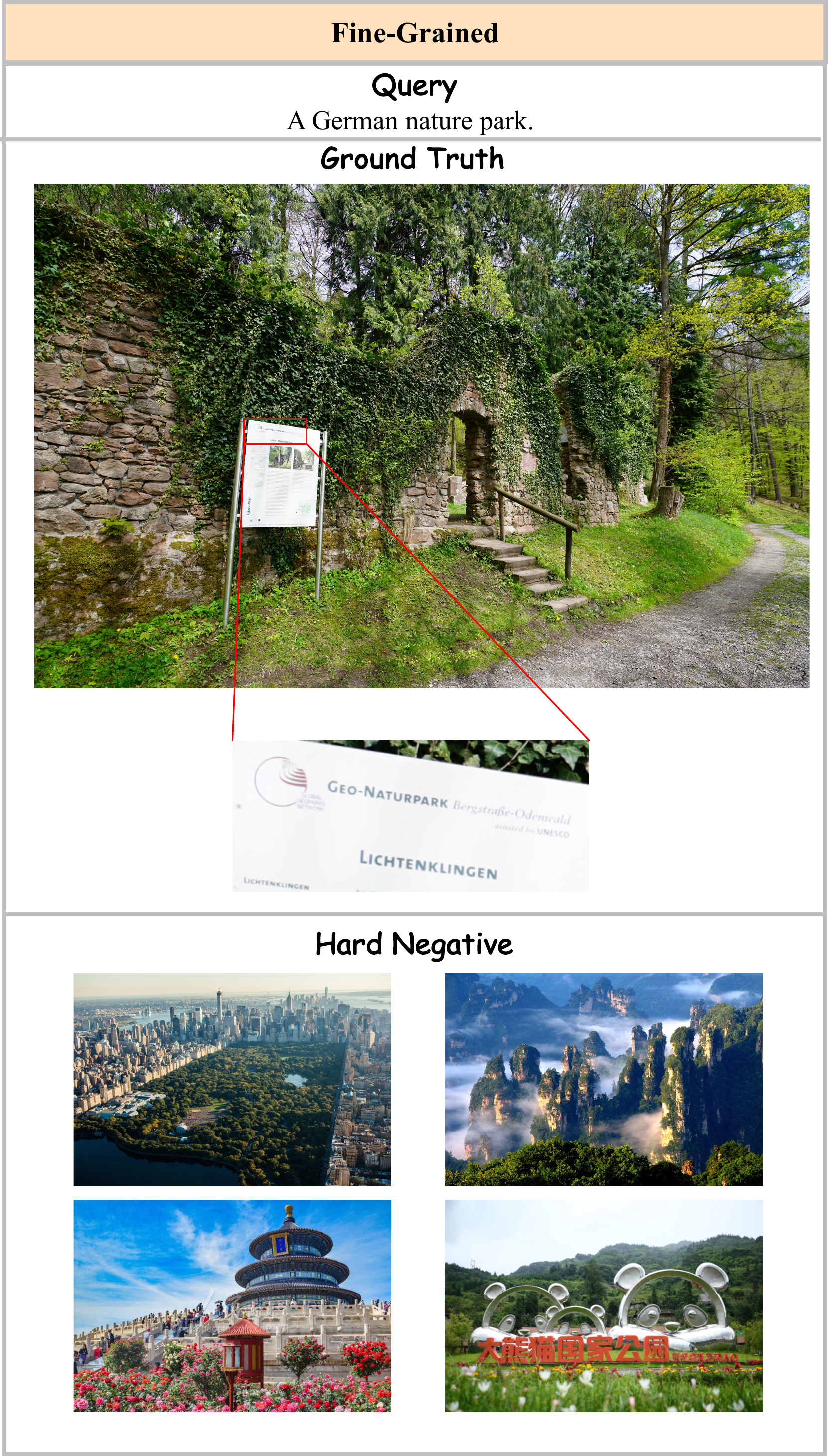}
  \caption{Visualized example of Fine-Grained subtype.}
  \label{fig: case_Fine_Grained}
\end{figure*}

\begin{figure*}[!th]
\centering
    \includegraphics[width=0.6\linewidth]{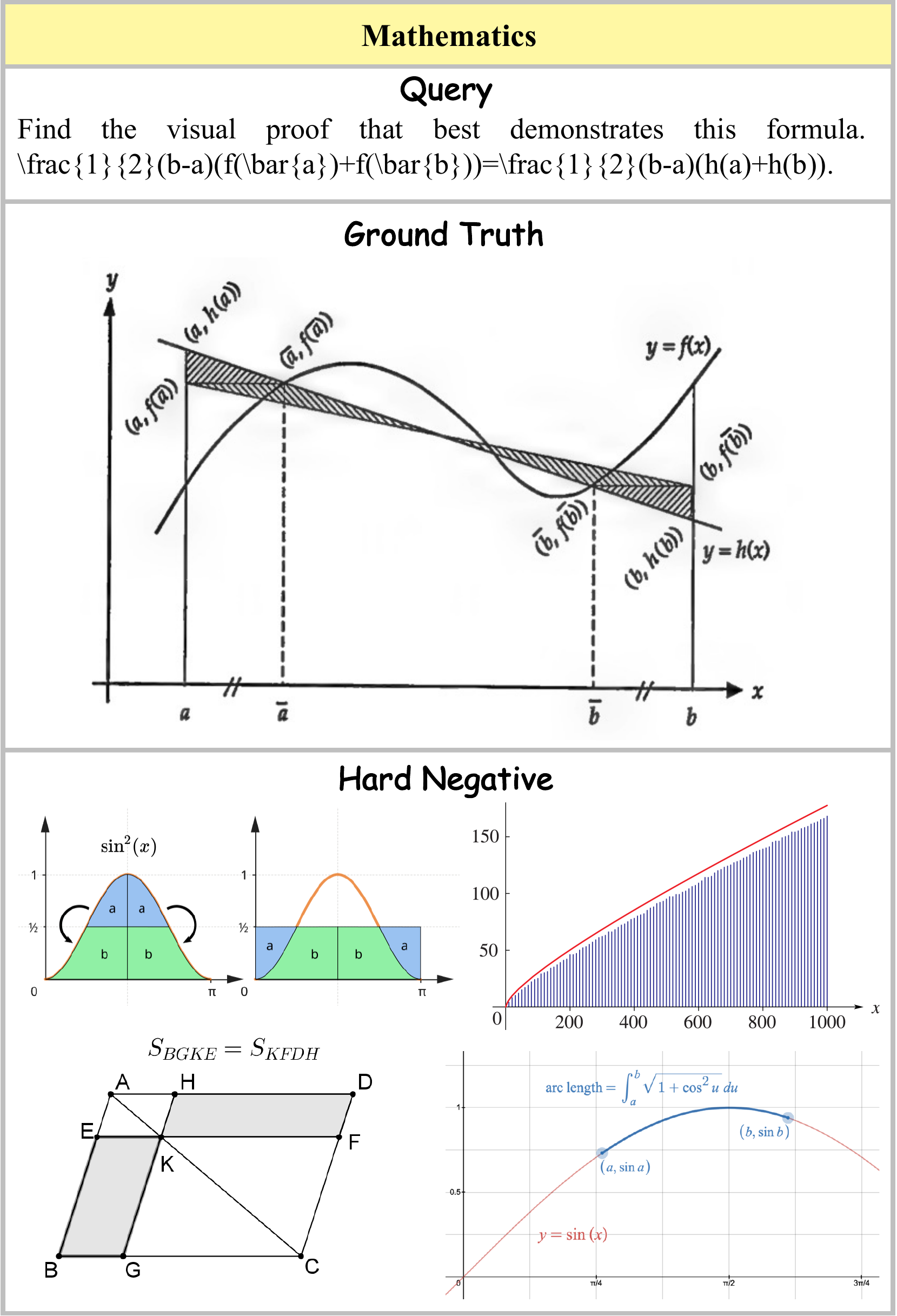}
  \caption{Visualized example of Mathematics subtype.}
  \label{fig: case_Mathematics}
\end{figure*}

\begin{figure*}[!th]
\centering
    \includegraphics[width=0.6\linewidth]{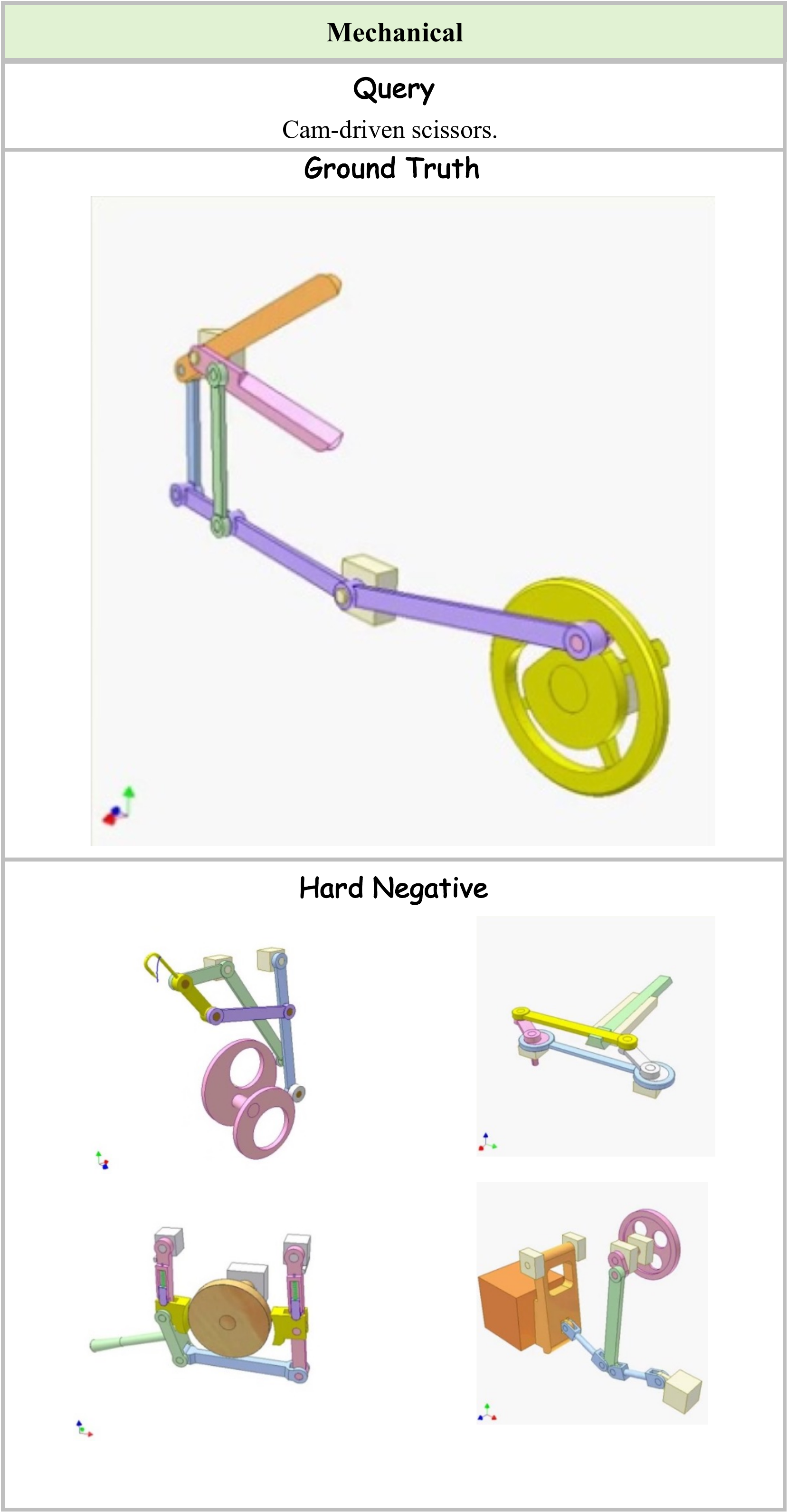}
  \caption{Visualized example of Mechanical subtype.}
  \label{fig: case_Mechanical}
\end{figure*}

\begin{figure*}[!th]
\centering
    \includegraphics[width=0.6\linewidth]{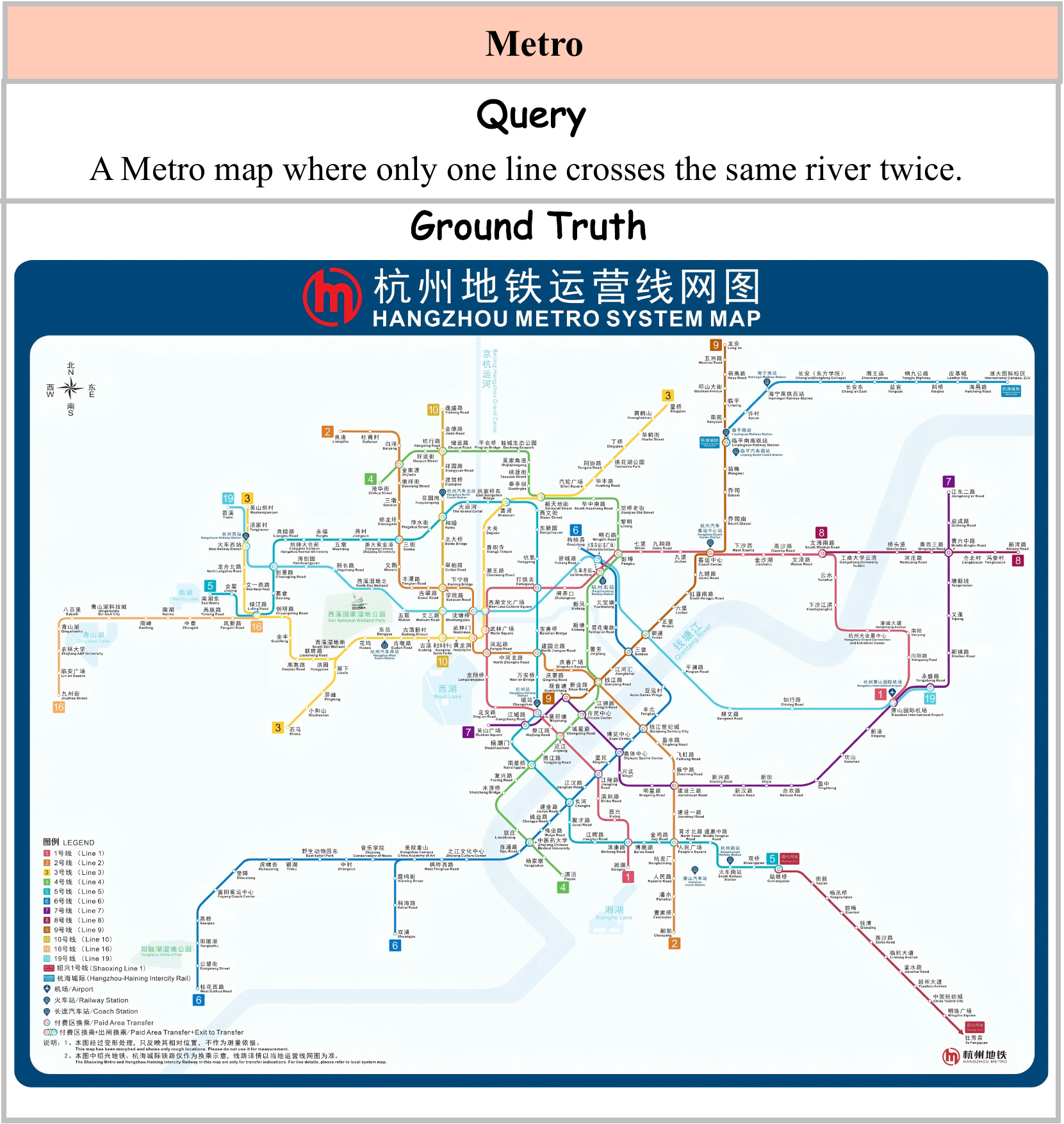}
  \caption{Visualized example of Metro subtype.}
  \label{fig: case_Metro}
\end{figure*}

\begin{figure*}[!th]
\centering
    \includegraphics[width=0.6\linewidth]{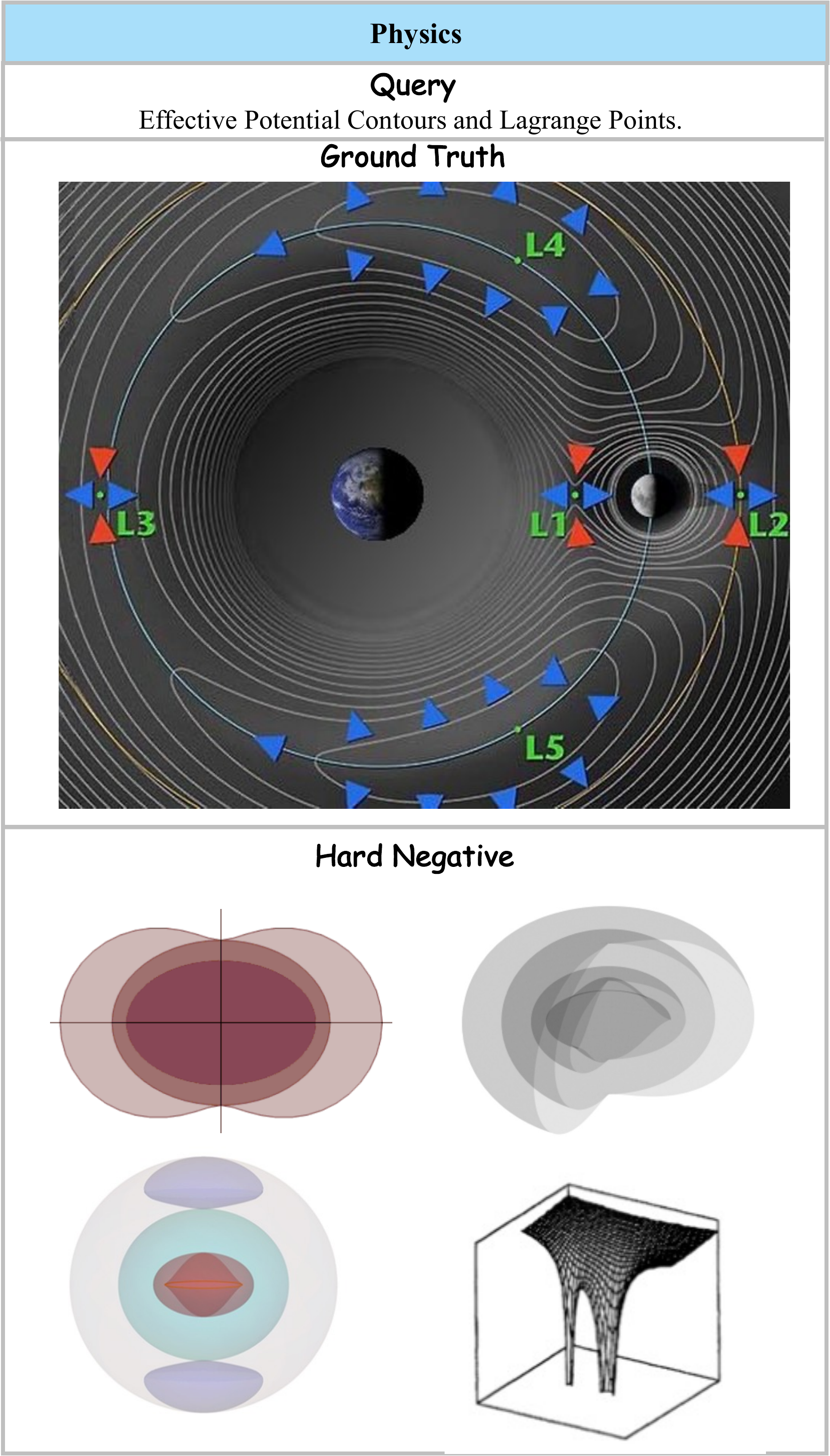}
  \caption{Visualized example of Physics subtype.}
  \label{fig: case_Physics}
\end{figure*}

\begin{figure*}[!th]
\centering
    \includegraphics[width=0.6\linewidth]{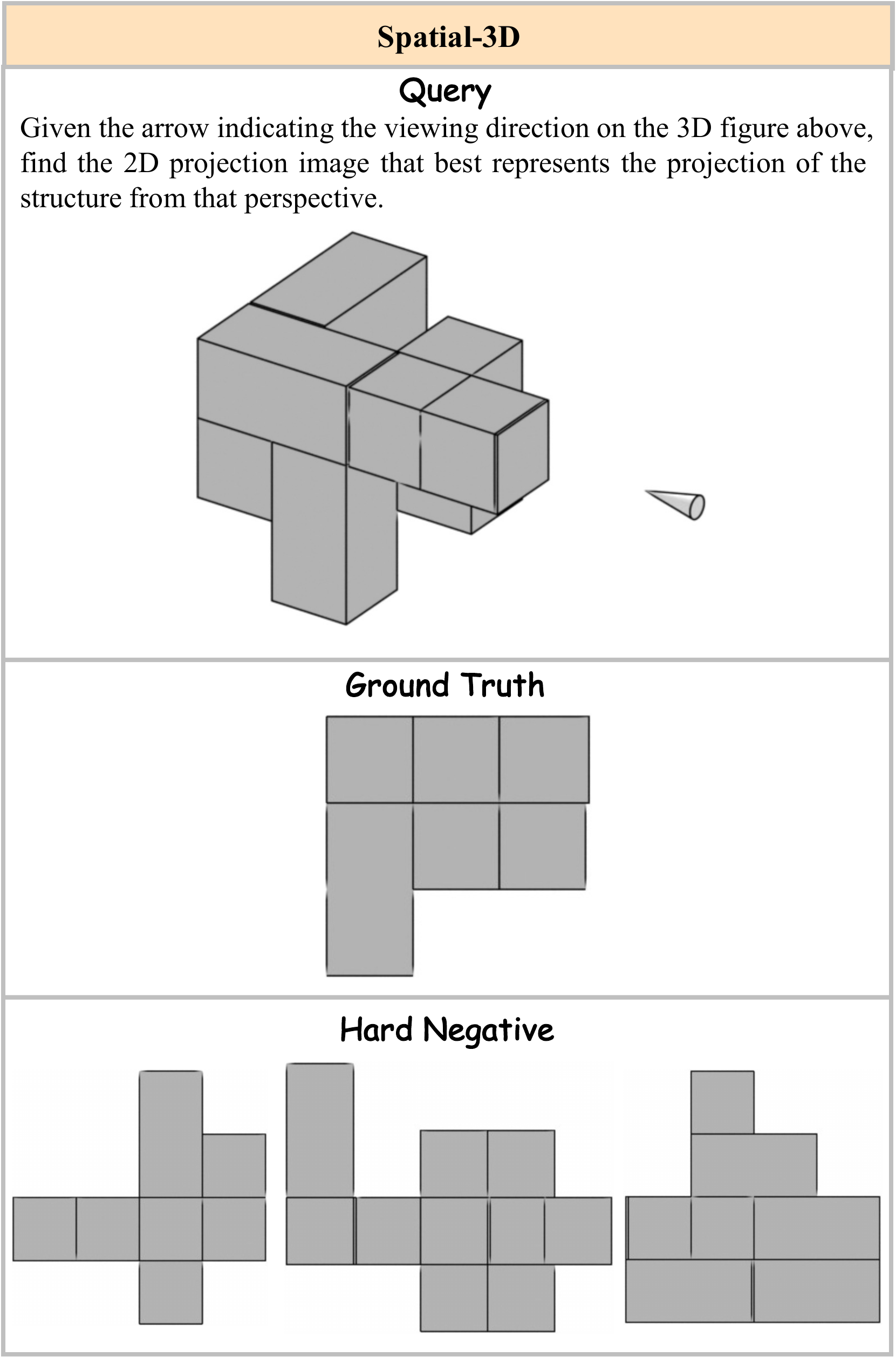}
  \caption{Visualized example of Spatial-3D subtype.}
  \label{fig: case_Spatial_3D}
\end{figure*}

\begin{figure*}[!th]
\centering
    \includegraphics[width=0.6\linewidth]{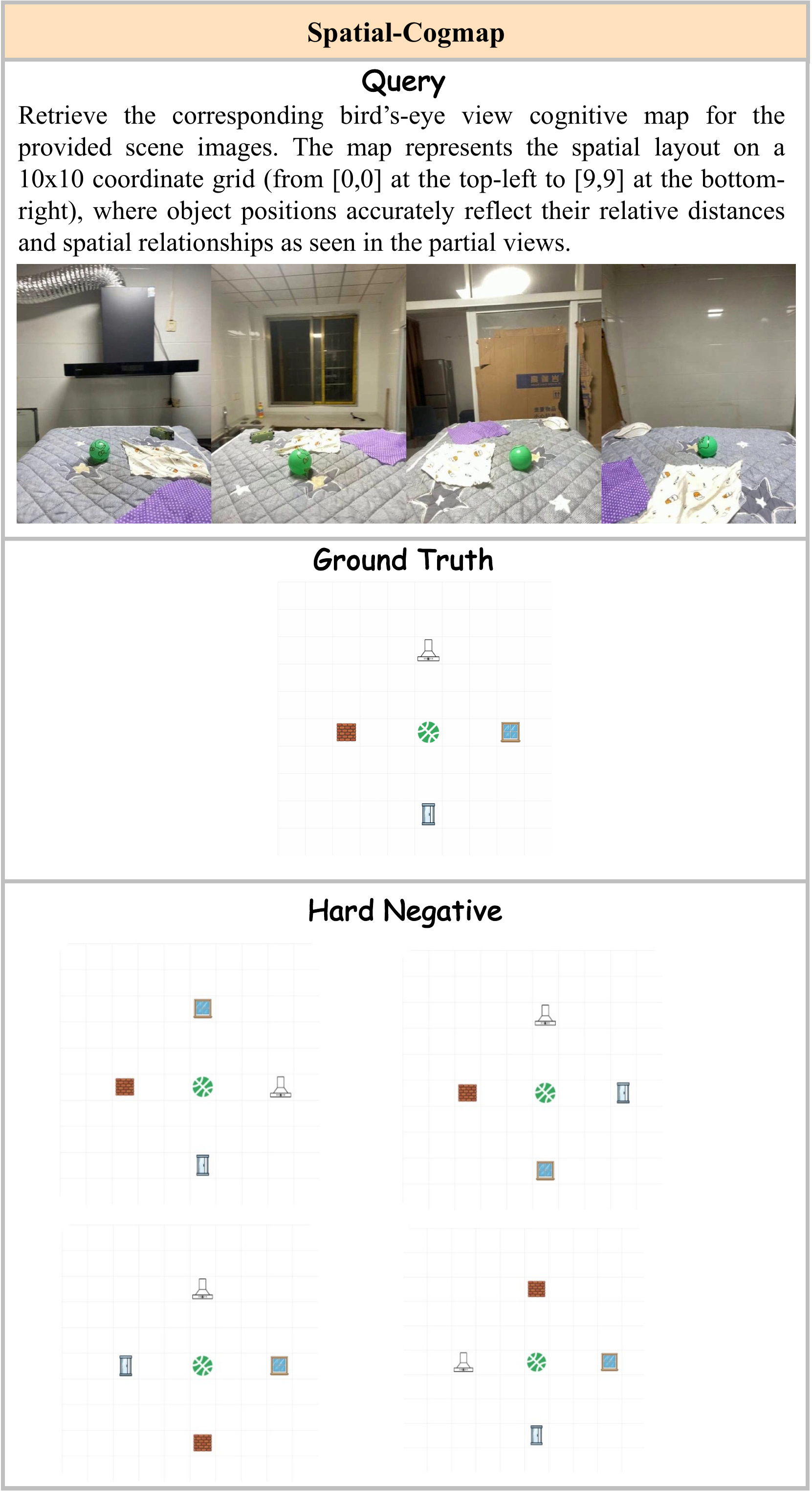}
  \caption{Visualized example of Spatial-Cogmap subtype.}
  \label{fig: case_Spatial_Cogmap}
\end{figure*}

In this section, we present representative examples for each task in ARK. Figures~\ref{fig: case_Art}--\ref{fig: case_Spatial_Cogmap} provide qualitative examples. These cases highlight the diversity of knowledge domains, visual data types, and reasoning skills covered by the benchmark.

\end{document}